\documentclass[preprint,12pt,authoryear]{elsarticle}

\usepackage[T1]{fontenc}
\usepackage[tablename=Table]{caption}
\usepackage{pdflscape}
\usepackage{longtable}
\usepackage[margin=1in]{geometry}
\usepackage{natbib}
\usepackage{subcaption}
\usepackage{graphicx}
\usepackage{tabularx}
\usepackage{array}
\usepackage{amssymb,amsmath}
\usepackage{amsthm}
\usepackage{amsfonts,mathtools}
\usepackage[ruled,vlined]{algorithm2e}
\usepackage{float}
\usepackage{booktabs}
\usepackage{multirow}
\usepackage{color}
\usepackage[table]{xcolor}
\usepackage{soul}
\usepackage{subcaption}
\usepackage{lscape}
\usepackage{placeins}
\usepackage{siunitx}
\usepackage{tikz}
\usetikzlibrary{arrows.meta,positioning,fit,calc}

\journal{Neural Networks}

\begin{document}

\begin{frontmatter}



\title{Structured Basis Function Networks: Loss-Centric Multi-Hypothesis Ensembles with Controllable Diversity} 


\author[addr1,addr2]{Alejandro Rodriguez Dominguez}
\ead{arodriguez@miraltabank.com}

\author[addr1]{Muhammad Shahzad\corref{mycorrespondingauthor}}
\ead{m.shahzad2@reading.ac.uk}

\author[addr1]{Xia Hong}
\ead{x.hong@reading.ac.uk}

\address[addr1]{University of Reading, Department of Computer Science, Whiteknights House, Reading, RG6 6UR, United Kingdom}
\address[addr2]{Miralta Finance Bank S.A.,28020, Madrid, Spain }

\begin{abstract}
Ensemble learning and multi-hypothesis prediction are two complementary paradigms for improving predictive performance and robustness. This work introduces the Structured Basis Function Network (s\mbox{-}BFN), a loss-aware ensemble framework that unifies these paradigms through a supervised aggregation mechanism motivated by the
loss-induced Bregman-centroid geometry of prediction space. The approach constructs a structured dataset of base predictions and trains a parametric combiner directly against the observed labels, with the centroidal formulation serving as a geometric reference rather than an exact centroid solver, while a \textcolor{black}{diversity-control parameter} regulates specialization among predictors. The proposed framework is evaluated across regression and classification tasks, including tabular data, image recognition, and large-scale temporally structured video classification. Empirical results show that s\mbox{-}BFN provides strong performance across the evaluated regimes and is competitive with the aggregation baselines considered in the study. The strongest gains are observed in higher-dimensional and temporally
structured settings, where aggregation capacity and controlled diversity are jointly beneficial, while simpler tabular regimes exhibit earlier saturation. These results highlight a capacity--complexity--diversity trade-off rather than a strictly monotonic relationship between diversity, ensemble size, and accuracy.
\end{abstract}



\end{frontmatter}





%

\section{Introduction}

Many real-world prediction problems involve inherent uncertainty, where multiple valid outputs may correspond to a single input. This ambiguity is especially pronounced in structured tasks such as 3D human pose estimation, trajectory prediction, or image-to-geometry mappings, where representing the conditional output distribution with a single deterministic prediction is often inadequate. To address this, multi-hypothesis prediction (MHP) frameworks aim to generate a set of plausible outputs, offering a more expressive and interpretable modeling of uncertainty.

Early work in Multiple Choice Learning (MCL)~\citep{NIPS2012_cfbce4c1,pmlr-v33-guzman-rivera14} formalized this idea through ensembles of predictors trained under a Winner-Takes-All (WTA) loss, encouraging hypothesis specialization. Diverse MCL further introduced regularization terms to enhance output diversity. However, these approaches typically require training disjoint networks with limited information sharing, which restricts scalability. The Multiple Hypotheses Prediction framework~\citep{rupprecht2017learning} advanced this line of work by introducing a Voronoi interpretation in loss space, enabling shared architectures and implicit output partitioning. Nevertheless, training instability and the absence of a principled mechanism for combining hypotheses into a final prediction remain open challenges. Subsequent WTA-based variants~\citep{DBLP:journals/corr/abs-2110-02858} improve hypothesis spread but do not resolve the problem of structured, loss-consistent aggregation.

In parallel, ensemble learning has long been recognized as an effective strategy for improving generalization and robustness. Classical methods such as Bagging~\citep{10.1023/A:1018054314350}, Boosting~\citep{inproceedingsBoosting}, and Stacking~\citep{WOLPERT1992241} leverage model diversity to reduce variance and overfitting. Foundational analyses, including ambiguity decomposition~\citep{NIPS1994_b8c37e33} and bias--variance--covariance trade-offs~\citep{Ueda1996GeneralizationEO}, highlight the importance of uncorrelated prediction errors. More recent work has connected ensemble aggregation to Bregman geometry~\citep{wood2023unified}, showing that loss-consistent ensemble predictors correspond to centroids induced by the task loss, thereby providing a principled geometric foundation for structured aggregation.

Despite their complementary objectives, multi-hypothesis prediction and ensemble learning have largely evolved independently. Existing MHP approaches emphasize output diversity but lack loss-aware integration mechanisms, while conventional ensembles focus on accuracy improvements without explicitly modeling structured ambiguity. Recent empirical evidence further indicates that the interaction between model capacity, diversity, and data complexity is nontrivial: indiscriminately increasing diversity in high-capacity networks can degrade performance~\citep{DBLP:journals/corr/abs-2302-00704}, whereas ensembles of smaller, heterogeneous models often generalize better in complex regimes~\citep{50123,10.1016/j.inffus.2022.12.010}. This complexity--capacity--diversity interaction motivates the need for ensemble formulations that adapt aggregation capacity and diversity to the structure of the prediction space.

To address these limitations, this work introduces the Structured Basis Function Network (s\mbox{-}BFN), a general-purpose framework that unifies multi-hypothesis prediction and ensemble learning under a loss-aware, geometry-consistent formulation. The approach constructs a structured representation of base learner predictions and learns a supervised aggregation function whose design is motivated by the loss-induced Bregman-centroid geometry of prediction space.
A closed-form least-squares solution for regression was previously introduced in~\citep{10.1007/978-3-031-77915-2_7}; this work extends the formulation through gradient-based optimization to accommodate arbitrary losses. This enables the selection of loss--geometry pairs that motivate the aggregation representation and output mapping. The Bregman centroid serves as a geometric reference for this construction rather than an explicit target of s\mbox{-}BFN training.
Predictor specialization during training is regulated through a tunable diversity mechanism, providing explicit control over the bias--variance--diversity trade-off~\citep{wood2023unified}.

The experimental evaluation is intentionally structured across datasets of increasing predictive and structural complexity, progressing from low-dimensional multivariate data to image classification and finally to temporally structured video recognition. This progression enables analysis of how aggregation capacity and controlled diversity interact as the dimensionality and geometry of the prediction space become richer, and supports the evaluation of the proposed framework under increasingly demanding predictive regimes.

Additionally, in classification settings, the formulation allows the number of basis functions to exceed the number of hypotheses, offering supplementary representational flexibility at the aggregation level. The empirical evaluation demonstrates overall improvements in accuracy, stability, and efficiency, particularly in heterogeneous and higher-complexity settings. In particular, probability-space aggregation with s\mbox{-}BFN is competitive with averaging, stacking, neural-ensemble, greedy-selection, and Mixture-of-Experts baselines, with relative performance depending on the dataset and ensemble configuration.

The contributions of this work are as follows.
\begin{enumerate}
    \item A loss-geometry-aware framework that unifies multi-hypothesis prediction and ensembling through a supervised structured combiner motivated by Bregman-centroid geometry,
    complemented by a parametric diversity mechanism.
    \item A gradient-based extension of the closed-form least-squares formulation that enables loss-task generalization beyond regression.
    \item A comprehensive empirical evaluation spanning multivariate, image, and temporal video data, characterizing how aggregation capacity and diversity jointly influence generalization across increasing levels of data and prediction-space complexity.
\end{enumerate}

The remainder of the paper is organized as follows. Section~\ref{litrev} reviews related work. Section~\ref{modprop} presents the framework. Section~\ref{secexp} reports experiments and analysis. Section~\ref{conc} concludes with implications and future directions.

\section{Literature Review}
\label{litrev}

\subsection{Multiple Hypothesis Prediction and Structured Output Modeling}

Many prediction problems exhibit inherent ambiguity, where a single input may correspond to multiple valid outputs. Multiple Hypotheses Prediction (MHP) methods address this by generating a discrete set of plausible outputs to approximate the conditional output distribution, offering a principled mechanism for modeling structured uncertainty. Early work in Multiple Choice Learning (MCL) introduced the Winner-Takes-All (WTA) loss, updating only the most accurate hypothesis per instance to encourage specialization among predictors~\citep{NIPS2012_cfbce4c1, pmlr-v33-guzman-rivera14}. However, this approach often led to unstable convergence and incomplete coverage of the output space.

To improve robustness, \citet{rupprecht2017learning} proposed a geometric reformulation of WTA using Voronoi tessellations in loss space, enabling hypotheses to specialize within well-defined regions while sharing parameters. Nevertheless, hard WTA updates remain prone to mode collapse, prompting the development of relaxed or evolving WTA losses that better preserve distributional structure and improve convergence stability~\citep{DBLP:journals/corr/abs-2110-02858}. Complementary approaches such as Mixture Density Networks (MDNs) represent uncertainty continuously via Gaussian mixtures but lack the explicit interpretability and hypothesis separation offered by discrete methods~\citep{li2019generating}. More recently, transformer-based architectures, such as the Multi-Hypothesis Transformer (MHFormer) and other multi-output models, have integrated multiple interacting hypotheses into a single backbone to capture temporal and structural ambiguities~\citep{li2023mhformer}.

Despite these advances, most MHP frameworks lack a structured, loss-aware
mechanism for hypothesis aggregation. The proposed s-BFN addresses this gap
by treating hypotheses as structured basis functions and learning a combiner
aligned with the underlying loss geometry.  \textcolor{black}{Relative to the MHP formulation of \citet{rupprecht2017learning}, which provides a mechanism for generating specialized hypotheses, s-BFN additionally represents their joint predictions as a structured input and learns a supervised nonlinear aggregation mapping to the observed labels. Thus, multiple-hypothesis specialization is inherited from the MHP setting, while s-BFN extends this setting to learned structured aggregation.} In doing so, it unifies MHP with ensemble-based reasoning, enabling
efficient and geometry-consistent integration of diverse predictors.

\subsection{Ensemble Diversity and the Complexity--Capacity--Diversity Trade-Off}

Ensemble learning improves generalization by aggregating diverse predictors. In regression with squared loss, the ambiguity decomposition ensures that ensemble error is never worse than the average individual error~\citep{NIPS1994_b8c37e33}. For classification, diversity has been examined through probabilistic~\citep{Tumer1996}, geometric~\citep{brown2004diversity}, and information-theoretic~\citep{Rosset2006} formulations, although no universal diversity metric consistently predicts gains~\citep{articleKuncheva}. More recent theory generalizes the bias--variance decomposition to arbitrary losses via the Bregman divergence framework~\citep{BREGMAN1967200}, showing that the optimal combiner depends on loss geometry~\citep{wood2023unified}. 

Deep ensembles approximate Bayesian model averaging and benefit from diversity generated by different initializations and optimization paths~\citep{lakshminarayanan2017simple, fort2019deep, wilson2020bayesian}. Recent extensions in regression further highlight the importance of ensemble construction for uncertainty estimation, including bootstrap-based mechanisms that account for finite-data uncertainty~\citep{SLUIJTERMAN2025131500}. However, empirical studies reveal that naive diversity promotion in high-capacity models can reduce accuracy~\citep{DBLP:journals/corr/abs-2302-00704}, whereas ensembles of smaller yet diverse predictors can rival or outperform large monolithic models with lower computational cost~\citep{50123}. Efficiency-oriented strategies, such as snapshot ensembles~\citep{Huang2017} and committee-based architectures~\citep{DBLP:conf/iclr/WangKCKME22}, leverage this balance between capacity and diversity. Moreover, the impact of diversity depends on task complexity: on datasets with high structural variation or label noise, ensembles tend to yield greater benefits~\citep{990132}, and difficulty-aware allocation of examples to specialized members can further boost performance~\citep{10.1016/j.inffus.2022.12.010}. Related regression ensemble approaches also optimize global ensemble error jointly with diversity among individual predictors, reinforcing the need to treat diversity as a system-level property rather than only an individual-learner attribute~\citep{DURANROSAL2025130520}.

Architectural advances have further expanded the design space for ensemble methods. Transformer-based models have demonstrated competitive performance across domains~\citep{dosovitskiy2021vit}, with efficiency-oriented variants such as EfficientViT reducing attention cost while preserving accuracy~\citep{liu2023efficientvit}. Multi-hypothesis transformers embed diversity within a single backbone, conceptually resonating with s-BFN's structured learning~\citep{li2023mhformer}. Beyond predictive accuracy, ensembles are central to robust uncertainty estimation, providing well-calibrated predictions under distributional shift and often outperforming Bayesian neural networks in practice~\citep{lakshminarayanan2017simple}. Loss-aware aggregation has been shown to further enhance calibration and robustness~\citep{10.1609/aaai.v37i7.26048,YUAN2025108016}, and pruning strategies that jointly optimize accuracy and diversity~\citep{GULDOGUS2025107544} or dynamic weighting schemes tailored to changing conditions~\citep{ZHOU2025107810} illustrate the continuing move toward structured and adaptive integration. Theoretical perspectives, such as minimax-optimal risk bounds for ensembles~\citep{ZINCHENKO2024364}, reinforce the need for principled combination rules that mirror the role of geometry in defining optimal aggregators.

\textcolor{black}{Recent semi-supervised ensemble approaches improve
pseudo-label reliability through complementary mechanisms, including
structural decorrelation of prediction heads~\citep{wu2025decorrelating},
channel-based ensembling to reduce bias and variance in
pseudo-labels~\citep{wu2024channel}, and joint use of predictive uncertainty
and confidence to assess pseudo-label utility~\citep{wu2026uncertainty}.
These approaches address a different setting from the diversity-control
mechanism considered here: s-BFN uses the task loss to identify the locally
preferred hypothesis and $\varepsilon$ to redistribute a fixed sample-wise
update weight across hypotheses, thereby regulating their specialization
during supervised multi-hypothesis learning rather than directly
decorrelating prediction heads, constructing pseudo-labels, or estimating
their utility. At a theoretical level, this mechanism is related to the
broader question of ensemble diversity studied by
\citet{wood2023unified}, who establish a loss-dependent
bias--variance--diversity framework. By coupling controlled hypothesis
specialization with learned structured aggregation, s-BFN provides an
empirical setting in which diversity control can be studied jointly with
base-learner capacity, aggregation capacity, and data complexity, motivating
the complexity--capacity--diversity trade-off examined in the following
sections.}

\section{Proposed Methodology}
\label{modprop}

This section first describes the multiple--hypotheses prediction framework with Voronoi tessellations, followed by the construction of the structured dataset and the training of a generic structured model. The role of loss geometry in providing a centroidal reference for ensemble aggregation is then outlined. The structured basis function network (s\mbox{-}BFN) and its optimization procedures are presented next, and finally, the incorporation of parametric diversity into the training process is detailed.
\textcolor{black}{For reference, Table~\ref{tab:notation} summarizes the main
notation used throughout the methodology. Additional symbols specific to
individual derivations or experiments are defined locally at first use.}

\begin{table}[t]
\centering
\caption{\textcolor{black}{Main notation used in the methodology.}}
\label{tab:notation}
\begin{tabular}{p{0.18\textwidth} p{0.72\textwidth}}
\hline
\textcolor{black}{\textbf{Symbol}} &
\textcolor{black}{\textbf{Meaning}} \\
\hline

\textcolor{black}{$N$} &
\textcolor{black}{Number of observations.} \\

\textcolor{black}{$M$} &
\textcolor{black}{Number of base predictors (hypotheses).} \\

\textcolor{black}{$C$} &
\textcolor{black}{Output dimension; number of classes in classification.} \\

\textcolor{black}{$f_{\boldsymbol{\theta}_j}$} &
\textcolor{black}{Base predictor $j$, parameterized by
$\boldsymbol{\theta}_j$.} \\

\textcolor{black}{$p_{ij}$} &
\textcolor{black}{Class-probability vector produced by base predictor $j$
for observation $i$.} \\

\textcolor{black}{$\boldsymbol{D}_i$} &
\textcolor{black}{Structured prediction vector obtained by concatenating
the $M$ base-predictor outputs for observation $i$.} \\

\textcolor{black}{$G$} &
\textcolor{black}{Supervised aggregation mapping applied to the structured
prediction vector.} \\

\textcolor{black}{$\varepsilon$} &
\textcolor{black}{Diversity-control parameter governing the redistribution
of predictor updates.} \\

\textcolor{black}{$\delta_i^{(j)}(\varepsilon)$} &
\textcolor{black}{Sample-wise modulation weight applied to the update of
predictor $j$.} \\

\textcolor{black}{$\phi$} &
\textcolor{black}{Strictly convex generator defining the Bregman divergence.} \\

\textcolor{black}{$B_\phi$} &
\textcolor{black}{Bregman divergence generated by $\phi$.} \\

\textcolor{black}{$K$} &
\textcolor{black}{Number of radial basis functions in the
s\mbox{-}BFN combiner.} \\

\textcolor{black}{$C_k$} &
\textcolor{black}{Centre of radial basis function $k$.} \\

\textcolor{black}{$\gamma_k$} &
\textcolor{black}{Positive scale parameter of radial basis function $k$.} \\

\textcolor{black}{$\Phi(\boldsymbol{D}_i;\vartheta)$} &
\textcolor{black}{Radial-basis representation of structured prediction
$\boldsymbol{D}_i$.} \\

\textcolor{black}{$\alpha$} &
\textcolor{black}{Parameters of the s\mbox{-}BFN linear readout.} \\

\textcolor{black}{$T$} &
\textcolor{black}{Softmax temperature in the cross-entropy specialization.} \\

\hline
\end{tabular}
\end{table}

\subsection{Multiple Hypotheses Preliminaries}

Given the training instances $\{\boldsymbol{x}_i\}_{i=1}^{N}$ and ground-truth labels $\{y_i\}_{i=1}^{N}$, the task in the supervised learning problem using a single hypothesis is to train a predictor $f_{\boldsymbol{\theta}}(\boldsymbol{x})$, parameterized by $\boldsymbol{\theta}$ such that the expected error for a particular loss $\mathcal{L}$ is minimized as follows:
\begin{equation}
\frac{1}{N}
\sum_{i=1}^{N}
\mathcal{L}
\left(
f_{\boldsymbol{\theta}}(\boldsymbol{x}_i),
y_i
\right)
\label{Eq10}
\end{equation}

Assuming that the training samples follow $p(\boldsymbol{x},y)$, the population-optimal prediction for a general loss $\mathcal{L}$ is the corresponding conditional-risk minimizer:
\begin{equation}
\textcolor{black}{
f^{\star}(\boldsymbol{x})
\in
\arg\min_{\hat{y}}
\mathbb{E}
\left[
\mathcal{L}(\hat{y},Y)
\mid X=\boldsymbol{x}
\right].
}
\label{eq:conditional_risk_minimizer}
\end{equation}

For squared-error regression, this reduces to the conditional mean,
\begin{equation}
\textcolor{black}{
f^{\star}(\boldsymbol{x})
=
\mathbb{E}[Y\mid X=\boldsymbol{x}]
=
\int_{\mathcal{Y}}
y\,p(y\mid\boldsymbol{x})\,dy,
}
\label{eq:conditional_mean_squared_loss}
\end{equation}
\textcolor{black}{
whereas for cross-entropy classification the population-optimal output is the conditional class-probability vector
$p(\,\cdot\mid\boldsymbol{x})$ \citep{Bishop1995}.}

In the case of multiple hypotheses, a set of prediction functions is considered, $\{f_{\boldsymbol{\theta}_j}(\boldsymbol{x})\}_{j=1}^M$, capable of providing $M$ predictions with the set of model parameters $\boldsymbol{\Theta}=\{\boldsymbol{\theta}_j\}_{j=1}^M$. The expected error for loss $\mathcal{L}$ in the MHP case is then formulated as follows
\citep{rupprecht2017learning}:

\begin{equation}
    \int_{\boldsymbol{X}}
    {
    \sum_{j=1}^{M}
    \int_{\mathcal{Y}_j(f_{\boldsymbol{\theta}_j}(\boldsymbol{x}))}
    {
    \mathcal{L}
    \left(
    f_{\boldsymbol{\theta}_j}(\boldsymbol{x}),
    y
    \right)
    }
    p(\boldsymbol{x},y)
    \,dy\,d\boldsymbol{x}
    }
    \label{eq14}
\end{equation}

During training, the Voronoi tessellation of the label space is induced by the losses computed from $M$ predictors and given as
\[
\mathcal{Y}
=
\bigcup_{j=1}^{M}
\mathcal{Y}_j
\left(
f_{\boldsymbol{\theta}_j}(\boldsymbol{x})
\right),
\]
where $\mathcal{Y}_j(f_{\boldsymbol{\theta}_j}(\boldsymbol{x}_i))$ represents the $j^{\mathrm{th}}$ cell, with $f_{\boldsymbol{\theta}_j}(\boldsymbol{x}_i)$ being the closest of the $M$ predictions to the label data for each training iteration \citep{rupprecht2017learning}:

\begin{align}
\mathcal{Y}_j
\left(
f_{\boldsymbol{\theta}_j}(\boldsymbol{x}_i)
\right)
&=
\left\{
y_i\in\mathcal{Y}_j :
\mathcal{L}
\left(
f_{\boldsymbol{\theta}_j}(\boldsymbol{x}_i),
y_i
\right)
<
\mathcal{L}
\left(
f_{\boldsymbol{\theta}_k}(\boldsymbol{x}_i),
y_i
\right)
\right.
\nonumber\\
&\qquad\qquad
\left.
\forall k\neq j
\right\}.
\label{Eq15}
\end{align}

While implementing (\ref{Eq15}), a typical approach adopted to avoid mode collapse is to relax the best-of-\textit{M} approach by updating all predictors in each iteration \citep{NIPS2012_cfbce4c1,pmlr-v33-guzman-rivera14}. Existing works either focus on multi-output prediction or do not provide an efficient way to combine the base learners or multiple hypotheses, often relying on numerical methods \citep{NIPS2012_cfbce4c1,pmlr-v33-guzman-rivera14,9893798}. To this end,
this work aims to efficiently combine, in a structured model, the set of hypotheses that form the centroidal Voronoi tessellations. Additionally, the experiments examine how manipulating the shape of the tessellations formed during predictor training affects generalization performance through the diversity induced among predictors.

\subsection{Structured Dataset Formation}
\label{datasetformation}

The structured dataset $\boldsymbol{D}$ aggregates the outputs of $M$ base predictors evaluated on $N$ inputs. Let
$f_{\boldsymbol{\theta}_j} : \mathcal{X} \to \mathbb{R}^{d_j}$ denote the $j$-th predictor with parameters $\boldsymbol{\theta}_j$. For input $\boldsymbol{x}_i$, the per-instance structured vector is defined as the concatenation
\[
\boldsymbol{D}_i :=
\big[
f_{\boldsymbol{\theta}_1}(\boldsymbol{x}_i)^\top,\,
\ldots,\,
f_{\boldsymbol{\theta}_M}(\boldsymbol{x}_i)^\top
\big]^\top
\in \mathbb{R}^{d_D},
\qquad
d_D := \sum_{j=1}^M d_j,
\]
and the full structured dataset is
\[
\boldsymbol{D} =
\begin{bmatrix}
\boldsymbol{D}_1^\top \\
\vdots \\
\boldsymbol{D}_N^\top
\end{bmatrix}
\in \mathbb{R}^{N \times d_D}.
\]

Two construction regimes are considered. In the plug-in regime, predictors $\{f_{\boldsymbol{\theta}_j}\}_{j=1}^M$ are trained independently and $\boldsymbol{D}$ is obtained by a forward pass on $\{\boldsymbol{x}_i\}_{i=1}^N$.
In the on-the-fly regime, $\boldsymbol{D}_i$ is formed at iteration $i$ and immediately consumed by a generic combiner $\mathcal{G}$ to produce $\hat{y}_i = \mathcal{G}(\boldsymbol{D}_i, \boldsymbol{\alpha})$. Parameters are then updated by first-order steps
\begin{align}
\boldsymbol{\alpha} &\leftarrow \boldsymbol{\alpha}
- \eta_\alpha \big(
\nabla_{\boldsymbol{\alpha}}\mathcal{L}(\hat{y}_i,y_i)
+ \nabla_{\boldsymbol{\alpha}}\mathcal{R}_\alpha(\boldsymbol{\alpha})
\big),
\label{eq:on-the-fly-alpha} \\
\boldsymbol{\theta}_j &\leftarrow \boldsymbol{\theta}_j
- \eta_\theta^{(j)} \big(
\nabla_{\boldsymbol{\theta}_j}\mathcal{L}(f_{\boldsymbol{\theta}_j}(\boldsymbol{x}_i),y_i)
+ \lambda_1^{\,j}\nabla_{\boldsymbol{\theta}_j}\mathcal{R}_{\theta_j}(\boldsymbol{\theta}_j)
\big),
\quad j=1,\ldots,M,
\label{eq:on-the-fly-theta}
\end{align}
where $\{\lambda_1^{\,j}\}_{j=1}^M$ are per-predictor regularization coefficients. The dimensionality of $f_{\boldsymbol{\theta}_j}(\boldsymbol{x}_i)$ (and hence $d_D$) together with the normalization of $\hat{y}_i$ are dictated by the loss-induced geometry introduced next.

\subsubsection{Loss Geometry and Centroid-based Ensemble Combiners}
\label{lossgeo}

The relationship between loss functions and ensemble aggregation rules is characterized by Bregman divergences \citep{BREGMAN1967200}. For admissible losses, the induced Bregman geometry defines a unique notion of central tendency in
prediction space, and Bregman-optimal aggregation of the base predictions corresponds to the associated centroidal rule \citep{banerjee2005clustering,wood2023unified}. In this work, squared loss is used for regression tasks and cross-entropy loss for classification.

Under Euclidean geometry (squared loss), $\mathcal{L}_i = \tfrac{1}{2}\|\hat{y}_i - y_i\|^2$, predictor outputs are scalar
($d_j = 1$), yielding $\boldsymbol{D} \in \mathbb{R}^{N \times M}$ with $\boldsymbol{D}_i \in \mathbb{R}^M$. The combiner produces an unnormalised scalar prediction
\[
\hat{y}_i = \mathcal{G}(\boldsymbol{D}_i, \boldsymbol{\alpha}).
\]

Under simplex geometry (cross-entropy),
$\mathcal{L}_i = -\log \hat{\boldsymbol{y}}_{i,y_i}$, each predictor returns a class-probability vector $\boldsymbol{p}_{ij} \in \mathbb{R}^C$, corresponding to $d_j = C$. For input $\boldsymbol{x}_i$,
\[
\boldsymbol{p}_{ij} =
\operatorname{softmax}\!\big(f_{\boldsymbol{\theta}_j}(\boldsymbol{x}_i)\big),
\qquad
\boldsymbol{D}_i =
\big[(\boldsymbol{p}_{i1})^\top,\ldots,(\boldsymbol{p}_{iM})^\top\big]^\top
\in \mathbb{R}^{MC},
\]
with $MC = \sum_{j=1}^M d_j = M \times C$. The combiner output is normalised as
\[
\hat{\boldsymbol{y}}_i =
\operatorname{softmax}\!\big(\mathcal{G}(\boldsymbol{D}_i,\boldsymbol{\alpha})\big)
\in \Delta^{C-1}.
\]

A summary of the corresponding geometry, loss functions, and prediction mappings is provided in Table~\ref{tab:geom_loss_shapes_pred}, and the overall workflow is illustrated in Figure~\ref{fig:ensemble_pipeline}. Taken together, this loss-induced characterization implies that ensemble aggregation can be viewed as a deterministic mapping from the structured prediction vector $\boldsymbol{D}_i$ to a unique point in prediction space.

\subsubsection{Centroid Mapping and Scope of Approximation}

For admissible losses, the induced Bregman divergence defines a unique
\textcolor{black}{centroidal aggregation of the base predictions under the
corresponding loss geometry} \citep{banerjee2005clustering,wood2023unified}.
For a fixed structured prediction $\boldsymbol{D}_i$, the centroid is given by
\begin{equation}
\textcolor{black}{
y_i^{\star}
=
g^{\star}(\boldsymbol{D}_i)
=
\arg\min_{y\in\mathcal{Y}}
\frac{1}{M}
\sum_{j=1}^{M}
B_{\phi}
\left(
y\,\|\,f_{\boldsymbol{\theta}_j}(\boldsymbol{x}_i)
\right),
\qquad
\nabla\phi(y_i^{\star})
=
\frac{1}{M}\sum_{j=1}^{M}
\nabla\phi\!\left(
f_{\boldsymbol{\theta}_j}(\boldsymbol{x}_i)
\right)
}
\label{eq:bregman-centroid}
\end{equation}
which is unique \textcolor{black}{under the regularity conditions associated
with the selected Bregman geometry}. Thus, $g^{\star}$ defines a deterministic
centroid mapping from $\boldsymbol{D}_i$ to $y_i^{\star}$, which
\textcolor{black}{minimizes the average Bregman divergence to the base
predictions and is continuous under the same regularity conditions}
\citep{wood2023unified}.

A direct implication is that, for a fixed admissible loss, the centroid $y_i^{\star}$ uniquely minimizes \textcolor{black}{the Bregman aggregation objective in (\ref{eq:bregman-centroid})}. \textcolor{black}{This statement concerns the geometric aggregation of the base predictions and does not imply that $y_i^{\star}$ is necessarily the Bayes-optimal predictor of the observed label under the supervised loss.} In particular, arithmetic averaging is loss-consistent under Euclidean geometry (squared loss), whereas other common heuristics (e.g., averaging in logit versus probability space) implicitly correspond to different aggregation geometries and may yield materially different behavior \citep{DBLP:conf/ijcai/TassiGFT22}.

\textcolor{black}{The structured basis function network introduced in this work is trained as a supervised parametric combiner on the structured predictions $\boldsymbol{D}_i$ and the observed labels $y_i$. It is therefore not trained to reproduce $g^{\star}$ and does not explicitly solve the centroid-defining optimization problem. Instead, the centroidal formulation provides the loss-geometric motivation for the structured representation and aggregation architecture.} No formal approximation error bounds, convergence rates, or loss-dependent distances to the exact centroid are claimed. \textcolor{black}{Accordingly, increasing the number of basis functions $K$ should be interpreted as increasing the capacity of the supervised combiner, rather than as guaranteeing convergence to the exact centroid.}

Finally, we emphasize that \textcolor{black}{estimation and optimization errors in the supervised aggregation stage are conceptually distinct from the error associated with representing the centroid reference.} \textcolor{black}{Appendix~C formalizes this distinction by separating the estimation, optimization, and supervised approximation terms from the centroid-reference representation error.} This distinction supports the use of \textcolor{black}{flexible supervised parametric combiners} in high-dimensional settings, where
\textcolor{black}{the loss-induced centroid serves as a geometric reference but need not coincide with the label-optimal supervised combination of the available predictions.
}

\subsection{s-BFN Optimization}
\label{sec:sbfnopt}

This section instantiates the generic combiner \(\mathcal{G}\) as a structured basis function network (s-BFN). Each structured vector \(\boldsymbol{D}_i\in\mathbb{R}^{d_D}\) is mapped by a radial–basis feature map with \(K\) units and then aggregated linearly in that feature space. If the radial-basis function is defined as
\[
\boldsymbol{\Phi}:\ \mathbb{R}^{d_D}\to\mathbb{R}^{K},\qquad
\big(\boldsymbol{\Phi}(\boldsymbol{u};\boldsymbol{\vartheta})\big)_k 
= \varphi\!\Big(\frac{\|\boldsymbol{u}-\boldsymbol{C}_k\|_2}{\gamma_k}\Big),\quad k=1,\ldots,K,
\]
where \(\boldsymbol{\vartheta}=\{(\boldsymbol{C}_k,\gamma_k)\}_{k=1}^K\) collects centres \(\boldsymbol{C}_k\in\mathbb{R}^{d_D}\) and scales \(\gamma_k>0\), and \(\varphi(\cdot)\) is any positive radial profile. The Gaussian specialisation used in the experiments sets \(\varphi(r)=\exp(-\tfrac{1}{2}r^2)\). Additional basis-function choices and their empirical effect on approximation accuracy are documented in~\ref{app:bf_compare}. Unless stated otherwise, the Euclidean instance uses \(K=M\); in the cross-entropy experiments, \(K\) is allowed to differ from \(M\) to increase representational capacity, following the relation \(K=\kappa M\), where \(\kappa \ge 1\) is an integer capacity multiplier controlling the number of basis functions in the aggregation module. Throughout the experiments, \(\kappa\) is restricted to integer values.

\medskip
\noindent\textit{Cross-entropy specialisation.}
Each base predictor’s logits are first mapped to class-probability vectors \(\boldsymbol{p}_{ij}\in\mathbb{R}^{C}\) via a softmax. Hence \(d_j=C\) and \(d_D=MC\), and the structured vector is
\(\boldsymbol{D}_i=[\boldsymbol{p}_{i1}^\top,\ldots,\boldsymbol{p}_{iM}^\top]^\top\in\mathbb{R}^{MC}\).
Accordingly, the RBF centres satisfy \(\boldsymbol{C}_k\in\mathbb{R}^{MC}\). In this setting, a temperature \(T>0\) scales the combiner logits prior to the final softmax, i.e., \(\hat{\boldsymbol{y}}_i=\operatorname{softmax}\big((\boldsymbol{\Phi}(\boldsymbol{D}_i)\,\boldsymbol{\alpha})/T\big)\). Inputs to \(\boldsymbol{\Phi}\) are the per-model probability vectors (not raw logits); the linear head produces logits. A per-sample normalisation of the \(K\) radial features (e.g., LayerNorm) may be applied before the linear aggregation to improve conditioning.

\begin{table}[t]
\centering
\footnotesize
\caption{Geometry, loss, generic-combiner prediction, and s-BFN prediction.}
\label{tab:geom_loss_shapes_pred}
\begin{tabularx}{\textwidth}{@{}l l X X@{}}
\toprule
\textbf{Geometry} & \textbf{Loss }$\mathcal{L}_i$ & \textbf{Prediction (generic)} & \textbf{Prediction (s-BFN)} \\
\midrule
Euclidean (squared loss) 
& $\tfrac{1}{2}\|\hat{y}_i - y_i\|^2$
& $\hat{y}_i=\mathcal{G}(\boldsymbol{D}_i,\boldsymbol{\alpha})$
& $\hat{y}_i=\boldsymbol{\Phi}(\boldsymbol{D}_i)\,\boldsymbol{\alpha}$ \\
Simplex (cross-entropy)
& $-\log \hat{\boldsymbol{y}}_{i,y_i}$
& $\hat{\boldsymbol{y}}_i=\operatorname{softmax}\!\big(\mathcal{G}(\boldsymbol{D}_i,\boldsymbol{\alpha})\big)$
& $\hat{\boldsymbol{y}}_i=\operatorname{softmax}\!\big(\boldsymbol{\Phi}(\boldsymbol{D}_i)\,\boldsymbol{\alpha}\big)$ \\
\bottomrule
\end{tabularx}
\end{table}

\textcolor{black}{
For high-dimensional classification problems, LayerNorm should not be interpreted as dimensionality reduction. The radial distances are computed from the complete structured vector $\boldsymbol{D}_i\in\mathbb{R}^{MC}$, with no projection applied before the RBF map. The subsequent transformation $\boldsymbol{\Phi}:\mathbb{R}^{MC}\rightarrow\mathbb{R}^{K}$ provides a $K$-dimensional nonlinear representation of these full-space distances. Thus, when $K<MC$, dimensionality reduction occurs only after the distances have been evaluated, while LayerNorm acts on the resulting radial features to improve conditioning. Behavior for substantially larger class spaces, including $C\geq1000$, is discussed in Appendix~B.1.
}

\subsubsection{Analytical ensemble training}
\label{sec:analytical}

Under the squared–loss (Euclidean) setting, each base output contributes one scalar coordinate; thus \(\boldsymbol{D}_i\in\mathbb{R}^{M}\). The feature map returns \(\boldsymbol{\Phi}(\boldsymbol{D}_i)\in\mathbb{R}^{K}\) with \(K=M\) in this instance, the combiner weights are \(\boldsymbol{\alpha}\in\mathbb{R}^{K}\), and the prediction is scalar \(\hat{y}_i\in\mathbb{R}\). With least squares and an \(\ell_2\) penalty \(\frac{\lambda_2}{2}\|\boldsymbol{\alpha}\|_2^2\) (\(\lambda_2>0\)), the unique closed-form solution is
\begin{equation}
\boldsymbol{\alpha}
=\big(\boldsymbol{\Phi}^\top \boldsymbol{\Phi}+\lambda_2\boldsymbol{I}\big)^{-1}\boldsymbol{\Phi}^\top \boldsymbol{y},
\label{weights}
\end{equation}
where now \(\boldsymbol{\Phi}\in\mathbb{R}^{N\times K}\) is obtained by stacking the feature vectors \(\boldsymbol{\Phi}(\boldsymbol{D}_i)\) row-wise.

\subsubsection{Iterative network training}
\label{iterativeNT}

Although a closed-form estimator exists for the squared-loss case with \(\ell_2\) regularization (Eq.~\eqref{weights}), non-quadratic objectives, such as cross-entropy, or practical requirements (mini-batch optimization, streaming data, non-quadratic regularizers) motivate a first-order training procedure. Let \(\mathcal{B}\) be a mini-batch of size \(b\) and let \(\boldsymbol{\Phi}_\mathcal{B}\in\mathbb{R}^{b\times K}\) stack the \(K\)-dimensional feature vectors \(\boldsymbol{\Phi}(\boldsymbol{D}_i)\) for \(i\in\mathcal{B}\). For a differentiable objective \(\mathcal{L}(\hat{y},y)+\mathcal{R}_\alpha(\boldsymbol{\alpha})\), the generic mini-batch gradient and update are
\[
\nabla_{\boldsymbol{\alpha}}\mathcal{L}_\mathcal{B}
=\boldsymbol{\Phi}_\mathcal{B}^{\!\top}\,\nabla_{\hat{y}_\mathcal{B}}\mathcal{L}_\mathcal{B}
+\nabla_{\boldsymbol{\alpha}}\mathcal{R}_\alpha(\boldsymbol{\alpha}),
\qquad
\boldsymbol{\alpha}\leftarrow \boldsymbol{\alpha}-\eta_\alpha\,\nabla_{\boldsymbol{\alpha}}\mathcal{L}_\mathcal{B}.
\]
In the squared-loss (Euclidean) instance, \(\boldsymbol{\alpha}\in\mathbb{R}^{K}\) with \(K=M\) and \(\hat{\boldsymbol{y}}_\mathcal{B}=\boldsymbol{\Phi}_\mathcal{B}\boldsymbol{\alpha}\in\mathbb{R}^{b}\). With targets \(\boldsymbol{y}_\mathcal{B}\in\mathbb{R}^{b}\),
\[
\nabla_{\boldsymbol{\alpha}}\mathcal{L}_\mathcal{B}
=\frac{1}{b}\,\boldsymbol{\Phi}_\mathcal{B}^{\!\top}
\big(\hat{\boldsymbol{y}}_\mathcal{B}-\boldsymbol{y}_\mathcal{B}\big)
+\nabla_{\boldsymbol{\alpha}}\mathcal{R}_\alpha(\boldsymbol{\alpha}),
\qquad
\boldsymbol{\alpha}\leftarrow \boldsymbol{\alpha}
-\eta_\alpha\!\left[
\frac{1}{b}\,\boldsymbol{\Phi}_\mathcal{B}^{\!\top}
\big(\hat{\boldsymbol{y}}_\mathcal{B}-\boldsymbol{y}_\mathcal{B}\big)
+\nabla_{\boldsymbol{\alpha}}\mathcal{R}_\alpha(\boldsymbol{\alpha})
\right].
\]
In the cross-entropy (simplex) instance, \(\boldsymbol{\alpha}\in\mathbb{R}^{K\times C}\) and \(\hat{\mathbf{Y}}_\mathcal{B}=\boldsymbol{\Phi}_\mathcal{B}\boldsymbol{\alpha}\in\mathbb{R}^{b\times C}\) are the logits. With row-wise class probabilities \(\hat{\mathbf{P}}_\mathcal{B}=\mathrm{softmax}(\hat{\mathbf{Y}}_\mathcal{B})\) and one-hot labels \(\mathbf{T}_\mathcal{B}\in\mathbb{R}^{b\times C}\),
\[
\nabla_{\boldsymbol{\alpha}}\mathcal{L}_\mathcal{B}
=\frac{1}{b}\,\boldsymbol{\Phi}_\mathcal{B}^{\!\top}
\big(\hat{\mathbf{P}}_\mathcal{B}-\mathbf{T}_\mathcal{B}\big)
+\nabla_{\boldsymbol{\alpha}}\mathcal{R}_\alpha(\boldsymbol{\alpha}),
\qquad
\boldsymbol{\alpha}\leftarrow \boldsymbol{\alpha}
-\eta_\alpha\!\left[
\frac{1}{b}\,\boldsymbol{\Phi}_\mathcal{B}^{\!\top}
\big(\hat{\mathbf{P}}_\mathcal{B}-\mathbf{T}_\mathcal{B}\big)
+\nabla_{\boldsymbol{\alpha}}\mathcal{R}_\alpha(\boldsymbol{\alpha})
\right].
\]
The per-sample (online) variant corresponds to the special case \(b{=}1\) and is exactly the format employed in Algorithm~\ref{alg:sbfn-training}. The cross-entropy, \(K\neq M\) case with temperature scaling and feature normalization is implemented in Algorithm~\ref{alg:srbfn-vector-compact}.

\begin{algorithm}[t!]
\caption{Iterative s-BFN Training for Regression with
$\varepsilon$-Diversity and MSE Loss}
\label{alg:sbfn-training}

\KwData{
Batch $b{=}1$ (online training);
samples $\{(\boldsymbol{x}_i,y_i)\}_{i=1}^{N}$;
base predictors $\{f_{\boldsymbol{\theta}_j}\}_{j=1}^{M}$;
s-BFN parameters $\boldsymbol{\alpha}\in\mathbb{R}^{M}$;
running moments $\{(\mu_j,\sigma_j)\}_{j=1}^{M}$ for per-predictor
univariate RBFs;
hyperparameters:
$\{\lambda_1^{(j)}\}_{j=1}^{M}$ ($\ell_1$ for base learners),
$\lambda_2$ (ridge for s-BFN),
$\varepsilon\in[0,1]$;
learning rates $\{\eta_\theta^{(j)}\}_{j=1}^{M}$,
$\eta_\alpha$.
}

\KwResult{
Trained $\{\boldsymbol{\theta}_j\}_{j=1}^{M}$ and
$\boldsymbol{\alpha}$.
}

\For{$i = 1$ \KwTo $N$}{

    \For{$j = 1$ \KwTo $M$}{

        $z_i^{(j)}
        \leftarrow
        f_{\boldsymbol{\theta}_j}(\boldsymbol{x}_i)$;

        $\ell_i^{(j)}
        \leftarrow
        (z_i^{(j)} - y_i)^2$;
    }

    \For{$j = 1$ \KwTo $M$}{

        \textcolor{black}{
        $\delta_i^{(j)}
        \leftarrow
        \mathbb{I}\!\left[
        j =
        \arg\min_{k\in\{1,\dots,M\}}
        \ell_i^{(k)}
        \right](1-\varepsilon)
        +
        \left(
        1-
        \mathbb{I}\!\left[
        j =
        \arg\min_{k\in\{1,\dots,M\}}
        \ell_i^{(k)}
        \right]
        \right)
        \frac{\varepsilon}{M-1}$;
        }

        $\boldsymbol{\theta}_j
        \leftarrow
        \boldsymbol{\theta}_j
        -
        \eta_\theta^{(j)}
        \left(
        \delta_i^{(j)}
        \nabla_{\boldsymbol{\theta}_j}
        (z_i^{(j)} - y_i)^2
        +
        \frac{\lambda_1^{(j)}}{N}
        \operatorname{sgn}(\boldsymbol{\theta}_j)
        \right)$;

        update $(\mu_j,\sigma_j)$ with $z_i^{(j)}$;

        $\phi_i^{(j)}
        \leftarrow
        \exp\!\left[
        -\frac{(z_i^{(j)}-\mu_j)^2}{2\sigma_j^2}
        \right]$;
    }

    $\hat{y}_i
    \leftarrow
    \sum_{j=1}^{M}
    \alpha_j\,\phi_i^{(j)}$;

    \For{$j = 1$ \KwTo $M$}{

        $\alpha_j
        \leftarrow
        \alpha_j
        -
        \eta_\alpha
        \left(
        (\hat{y}_i-y_i)\phi_i^{(j)}
        +
        \lambda_2\alpha_j
        \right)$;
    }
}

\end{algorithm}

\begin{algorithm}[t!]
\caption{Mini-batch s-BFN Training for Classification with Cross-Entropy}
\label{alg:srbfn-vector-compact}

\KwData{
Mini-batches
$(\mathbf{X}_{\mathcal{B}},\mathbf{y}_{\mathcal{B}})$
with
$\mathbf{X}_{\mathcal{B}}\in\mathbb{R}^{b\times d}$,
$\mathbf{y}_{\mathcal{B}}\in\{0,\dots,C{-}1\}^{b}$;
base models
$\{f_{\boldsymbol{\theta}^{(j)}}\}_{j=1}^{M}$;
s-BFN parameters:
centers $\mathbf{C}\in\mathbb{R}^{K\times(MC)}$,
\textcolor{black}{
scales
$\{\boldsymbol{\gamma}^{(e)}\}_{e=1}^{E}$,
with
$\boldsymbol{\gamma}^{(e)}\in\mathbb{R}_{>0}^{K}$,
}
combiner weights
$\boldsymbol{\alpha}\in\mathbb{R}^{K\times C}$,
temperature $T>0$;
hyperparameters:
$\varepsilon\in[0,1]$;
learning rates
$\{\eta_\theta^{(j)}\}_{j=1}^{M}$,
$\eta_\alpha$.
}

\KwResult{
Trained
$\{\boldsymbol{\theta}^{(j)}\}_{j=1}^{M}$,
$\boldsymbol{\alpha}$,
and (optionally) $\mathbf{C}$.
}

\ForEach{epoch \textcolor{black}{$e$}$=1,\dots,E$}{

    \textcolor{black}{
    $\boldsymbol{\gamma}
    \leftarrow
    \boldsymbol{\gamma}^{(e)}$;
    }

    \ForEach{mini-batch
    $(\mathbf{X}_{\mathcal{B}},\mathbf{y}_{\mathcal{B}})$}{

        \For{$j = 1$ \KwTo $M$}{

            $\mathbf{p}^{(j)}
            \leftarrow
            \mathrm{softmax}\!\left(
            f_{\boldsymbol{\theta}^{(j)}}
            (\mathbf{X}_{\mathcal{B}})
            \right)$;

            $\boldsymbol{\ell}^{(j)}
            \leftarrow
            \mathrm{CE}\!\left(
            \mathbf{p}^{(j)},
            \mathbf{y}_{\mathcal{B}}
            \right)$;
        }

        stack $\mathbf{P}$ and $\boldsymbol{\mathcal{L}}$;

        \For{$j = 1$ \KwTo $M$}{

            $\boldsymbol{\delta}^{(j)}
            \leftarrow
            \mathbb{I}\!\left[
            j =
            \arg\min_{k\in\{1,\dots,M\}}
            \boldsymbol{\ell}^{(k)}
            \right](1-\varepsilon)
            +
            \left(
            1-
            \mathbb{I}\!\left[
            j =
            \arg\min_{k\in\{1,\dots,M\}}
            \boldsymbol{\ell}^{(k)}
            \right]
            \right)
            \frac{\varepsilon}{M-1}$;

            $\mathcal{L}_{\theta^{(j)}}
            \leftarrow
            \frac{1}{b}
            \sum_{i=1}^{b}
            \delta_i^{(j)}\,
            \ell_i^{(j)}$;

            $\boldsymbol{\theta}^{(j)}
            \leftarrow
            \boldsymbol{\theta}^{(j)}
            -
            \eta_\theta^{(j)}
            \nabla
            \mathcal{L}_{\theta^{(j)}}$;
        }

        flatten $\mathbf{X}^{\mathrm{ens}}$ from $\mathbf{P}$ by
        concatenating the $M$ probability blocks per sample;

        $\mathbf{H}
        \leftarrow
        \exp\!\left(
        -\frac{
        \left\lVert
        \mathbf{X}^{\mathrm{ens}}-\mathbf{C}
        \right\rVert_2^2
        }{
        2\boldsymbol{\gamma}^2
        }
        \right)$;

        normalize rows of $\mathbf{H}$;

        $\hat{\mathbf{Y}}
        \leftarrow
        \mathbf{H}\boldsymbol{\alpha}/T$;

        $\hat{\mathbf{p}}
        \leftarrow
        \mathrm{softmax}(\hat{\mathbf{Y}})$;

        $\mathcal{L}_{\text{s-BFN}}
        \leftarrow
        \mathrm{CE}\!\left(
        \hat{\mathbf{p}},
        \mathbf{y}_{\mathcal{B}}
        \right)$;

        $\boldsymbol{\alpha}
        \leftarrow
        \boldsymbol{\alpha}
        -
        \eta_\alpha
        \nabla_{\boldsymbol{\alpha}}
        \mathcal{L}_{\text{s-BFN}}$;

        $\mathbf{C}
        \leftarrow
        \mathbf{C}
        -
        \eta_\alpha
        \nabla_{\mathbf{C}}
        \mathcal{L}_{\text{s-BFN}}$;
    }
}

\end{algorithm}

\textcolor{black}{
The positive scale sequence $\{\boldsymbol{\gamma}^{(e)}\}_{e=1}^{E}$ is specified before training. A fixed-scale configuration corresponds to
$\boldsymbol{\gamma}^{(e)}
=
\boldsymbol{\gamma}^{\star}
\in\mathbb{R}_{>0}^{K}$
for all $e$; otherwise, a prescribed positive annealing schedule may be used.
}

\begin{figure}[t!]
\centering
\begin{tikzpicture}[
scale=0.80, transform shape, 
>=Latex,
node distance = 0.9cm and 1.3cm, 
every node/.style={font=\small},
box/.style={draw, rounded corners, align=center, inner sep=4pt, minimum height=8mm, minimum width=33mm},
smallbox/.style={box, minimum width=29mm, minimum height=8mm},
widebox/.style={box, minimum width=42mm},
line/.style={-Latex, thick},
dashedframe/.style={draw, dashed, rounded corners, inner sep=4pt}
]

\node[smallbox] (p1) {Base predictor\\$f_{\boldsymbol{\theta}_1}(\boldsymbol{x})$};
\node[smallbox, right=of p1] (p2) {Base predictor\\$f_{\boldsymbol{\theta}_2}(\boldsymbol{x})$};
\node[right=of p2] (dots) {$\cdots$};
\node[smallbox, right=of dots] (p3) {Base predictor\\$f_{\boldsymbol{\theta}_M}(\boldsymbol{x})$};

\node[widebox, below=0.9cm of $(p1)!0.5!(p3)$] (D) {Structured dataset $\boldsymbol{D}$};

\draw[line] (p1) -- (D);
\draw[line] (p2) -- (D);
\draw[line] (p3) -- (D);

\node[widebox, below=0.85cm of D] (loss) {Select loss geometry:\\
Choose $\ell$ $\Longleftrightarrow$ Bregman divergence $\mathcal{B}_{\phi}$\\
\footnotesize (e.g., Euclidean / squared loss, KL / cross-entropy, Itakura--Saito)};

\draw[line] (D) -- (loss);

\node[widebox, below=0.85cm of loss] (breg) {\textcolor{black}{Bregman-centroid reference:}\\
\textcolor{black}{$\displaystyle \hat{\boldsymbol{z}}
= \arg\min_{\boldsymbol{z}}
\frac{1}{M}\sum_{j=1}^{M}
\mathcal{B}_{\phi}(\boldsymbol{z},\boldsymbol{z}_j)$}\\
\textcolor{black}{Provides the geometric reference for
$\mathcal{G}(\boldsymbol{D}_i,\boldsymbol{\alpha})$}};

\draw[line] (loss) -- (breg);

\node[widebox, below=0.85cm of breg] (sbfni) {s-BFN (radial-basis map, $K$ units):\\
$\mathcal{G}(\boldsymbol{D}_i,\boldsymbol{\alpha})
= \boldsymbol{\Phi}(\boldsymbol{D}_i;\boldsymbol{\vartheta})\boldsymbol{\alpha}$\\
\textcolor{black}{trained as a supervised combiner on the observed labels}};

\draw[line] (breg) -- (sbfni);

\node[smallbox, below left=0.6cm and 1.3cm of sbfni] (closed) {Closed-form solution\\
\footnotesize (Euclidean case)};
\node[smallbox, below right=0.6cm and 1.3cm of sbfni] (grad) {Iterative / gradient-based\\
\footnotesize (non-Euclidean)};

\draw[line] (sbfni.south) -| (closed.north);
\draw[line] (sbfni.south) -| (grad.north);

\node[dashedframe, fit=(p1) (p2) (dots) (p3) (D)] (frameA) {};
\node[dashedframe, fit=(loss) (breg) (sbfni) (closed) (grad)] (frameB) {};

\end{tikzpicture}

\caption{Workflow: base predictors produce a structured dataset $\boldsymbol{D}$. A suitable loss geometry is selected, inducing a Bregman divergence $\mathcal{B}_{\phi}$ (e.g., Euclidean/squared loss, KL/cross-entropy, or Itakura--Saito).
\textcolor{black}{The corresponding Bregman centroid \textcolor{black}{$\hat{\boldsymbol{z}}=\arg\min_{\boldsymbol{z}}
\frac{1}{M}\sum_{j=1}^{M} \mathcal{B}_{\phi}(\boldsymbol{z},\boldsymbol{z}_j)$} provides the geometric reference for the aggregation rule.} We instantiate $\mathcal{G}$ as an s-BFN with a radial map $\boldsymbol{\Phi}(\cdot;\boldsymbol{\vartheta})$ and $K$ units, \textcolor{black}{and train it as a supervised combiner on the observed labels rather than as an exact solver of the centroid optimization problem.} Optimization is closed-form in Euclidean settings and gradient-based for non-Euclidean losses.}
\label{fig:ensemble_pipeline}
\end{figure}

\subsubsection{Parametric diversity in learning}

Diversity across base predictors is induced by a sample-wise modulation factor $\delta_i^{(j)}(\varepsilon)\in\textcolor{black}{[0,1]}$ that reweights the parameter updates of each predictor. For input--label pair $(\boldsymbol{x}_i,y_i)$ and predictor $j$, write $D_{ij}=f_{\boldsymbol{\theta}_j}(\boldsymbol{x}_i)$. The $\varepsilon$-modulated update with generic per-predictor regularisation is

\begin{equation}
\boldsymbol{\theta}_{j}
\leftarrow
\boldsymbol{\theta}_{j}
-\eta_\theta^{(j)}
\left(
\delta_i^{(j)}
\nabla_{\boldsymbol{\theta}_j}
\mathcal{L}\!\big(D_{ij},y_i\big)
+
\lambda_1^{j}
\nabla_{\boldsymbol{\theta}_j}
\mathcal{R}_{\theta_j}(\boldsymbol{\theta}_j)
\right),
\label{predparam2}
\end{equation}

where the modulation term is

\begin{equation}
\delta_i^{(j)}
=
\begin{cases}
1-\varepsilon,
&
\text{if }
j=\arg\min\limits_{k=1,\dots,M}
\mathcal{L}\!\big(D_{ik},y_i\big),
\\[4pt]
\displaystyle \frac{\varepsilon}{M-1},
&
\text{otherwise},
\end{cases}
\label{equatruedelta}
\end{equation}

with $M$ the number of predictors and $\lambda_1^{j}>0$ a per-predictor regularisation coefficient.

\textcolor{black}{Equation~(\ref{equatruedelta}) applies for $M\geq2$; for the single-predictor case, we set $\delta_i^{(1)}=1$. The parameter $\varepsilon$ controls how the update weight is distributed between the predictor achieving the smallest instantaneous loss and the remaining predictors.} In this way, $\varepsilon$ controls the degree of specialisation and coupling among predictors, thereby discouraging collapse and encouraging complementary specialisation. {\color{black}For $M\geq2$, the modulation weights are non-negative and satisfy
\begin{equation}
\sum_{j=1}^{M}\delta_i^{(j)}=1.
\label{eq:epsilon_weight_sum}
\end{equation}
Thus, $\varepsilon$ redistributes a fixed total update weight among the predictors rather than adding stochastic noise to the training process. To characterize the effect of this redistribution on ensemble risk, let
\begin{equation}
\mathcal{R}_{G}(\boldsymbol{\Theta},\boldsymbol{\alpha})
=
\mathbb{E}
\left[
\mathcal{L}
\!\left(
G\!\left(\boldsymbol{D}(X),\boldsymbol{\alpha}\right),
Y
\right)
\right],
\label{eq:ensemble_risk}
\end{equation}
and let $w_i$ denote the predictor with the smallest loss for sample $i$.
Holding $\boldsymbol{\alpha}$ fixed and considering the data-loss component
of one sufficiently small update, define
\begin{equation}
A_i^{(j)}
=
\eta_\theta^{(j)}
\left\langle
\nabla_{\boldsymbol{\theta}_j}\mathcal{R}_{G},
\nabla_{\boldsymbol{\theta}_j}
\mathcal{L}\!\big(D_{ij},y_i\big)
\right\rangle .
\label{eq:epsilon_alignment}
\end{equation}

A first-order expansion gives
\begin{equation}
\mathcal{R}_{G}
\!\left(
\boldsymbol{\Theta}^{+},\boldsymbol{\alpha}
\right)
-
\mathcal{R}_{G}
\!\left(
\boldsymbol{\Theta},\boldsymbol{\alpha}
\right)
=
-
\left[
(1-\varepsilon)A_i^{(w_i)}
+
\frac{\varepsilon}{M-1}
\sum_{j\neq w_i}A_i^{(j)}
\right]
+
\mathcal{O}(\eta_\theta^2).
\label{eq:epsilon_risk_change}
\end{equation}

Accordingly, the modulated update locally decreases ensemble risk when the weighted quantity in brackets is positive. Relative to hard winner-takes-all training, a positive value of $\varepsilon$ provides a larger first-order risk reduction when
\begin{equation}
\frac{1}{M-1}
\sum_{j\neq w_i} A_i^{(j)}
>
A_i^{(w_i)}.
\label{eq:epsilon_alignment_condition}
\end{equation}
This condition means that sharing part of the update with the non-winning predictors is beneficial when their gradient directions are, on average, more useful for reducing the final ensemble risk than concentrating the entire update on the current winner.

The update rule does not guarantee monotonic improvement as $\varepsilon$
increases; rather, $\varepsilon$ provides continuous control over predictor
specialization. In the experiments, $\varepsilon$ is treated as a fixed
hyperparameter; when a single operating value is required, it is selected
using validation performance, whereas sensitivity analyses explicitly
evaluate multiple values of $\varepsilon$. A validation-driven adaptive
strategy could also be considered, but is not used in the present
implementation.
\textcolor{black}{Accordingly, $\varepsilon$ controls the degree of hypothesis
specialization during learning, and thereby the mechanism through which
diversity is induced; it is not itself a diversity measure. Classical
quantities such as ambiguity, disagreement, Bregman dispersion, and mutual
information instead measure properties of the resulting ensemble; establishing
a general mapping from $\varepsilon$ to these measures or to a
dataset-dependent optimal value is beyond the scope of this work.}
Their empirical relationships with the diversity-control mechanism
are examined in Section~\ref{secexp}.

After training with the diversity modulation, the ensemble prediction follows the loss-induced geometry. In the squared-loss (Euclidean) instance,

\[
\hat{y}_i
=
\boldsymbol{\Phi}\!\big(
\boldsymbol{D}_i(\varepsilon)
\big)\boldsymbol{\alpha},
\]
whereas in the cross-entropy (simplex) instance,
\[
\hat{\boldsymbol{y}}_i
=
\operatorname{softmax}\!\big(
\boldsymbol{\Phi}\!\big(
\boldsymbol{D}_i(\varepsilon)
\big)\boldsymbol{\alpha}
\big).
\]
Here, $\boldsymbol{D}_i(\varepsilon)$ denotes the structured vector obtained from the $\varepsilon$-modulated predictors, and $\boldsymbol{\alpha}$ are the s\mbox{-}BFN combiner parameters.

\section{Experiments}
\label{secexp}
The experimental framework adopts a progressive design, beginning with low-complexity tabular datasets and simple two-layer MLPs as base learners, which offer sufficient capacity for these tasks. As the dimensionality and heterogeneity of the data increase, the model complexity is scaled accordingly by incorporating deeper architectures suited to the problem. In later stages, image datasets are introduced, where deep learning models with higher capacity are employed. This allows us to evaluate how ensemble diversity, data complexity, and model capacity interact to shape generalization performance, and it also demonstrates the s-BFN framework’s flexibility in base-learner selection and task choice.

This progression is further extended to temporally structured video data, introducing a high-complexity regime with strong temporal dependencies and a substantially richer induced prediction space. In this setting, aggregation capacity and controlled diversity play a more prominent role, enabling us to assess the scalability and stability of \textcolor{black}{loss-aware aggregation} under high-dimensional and sequential prediction geometries.

Table~\ref{tab:complexity_progression} summarizes the experimental regimes considered in the evaluation. These regimes support the analysis of how aggregation capacity and controlled diversity behave as prediction spaces become richer and more structured. Across them, the proposed method is compared with task-appropriate baselines, including classical regression models, arithmetic multi-hypothesis aggregation, logit or probability averaging, linear fusion, and Mixture-of-Experts combiners.

\begin{table}[htbp]
\centering
\small
\caption{Experimental regimes considered in the evaluation.}
\label{tab:complexity_progression}
\begin{tabularx}{\textwidth}{%
>{\raggedright\arraybackslash}p{3.0cm}
>{\raggedright\arraybackslash}p{3.4cm}
>{\raggedright\arraybackslash}p{3.0cm}
>{\raggedright\arraybackslash}X}
\hline
\textbf{Regime} &
\textbf{Datasets} &
\textbf{Task type} &
\textbf{Prediction space} \\
\hline
Tabular regression &
Air Quality; Appliances Energy &
Multivariate regression &
Low-dimensional continuous output \\
\hline
Image classification &
MNIST; Fashion-MNIST; CIFAR-10 &
Multi-class classification &
10-class probability simplex \\
\hline
Video classification &
UCF101 &
101-class action recognition &
High-dimensional probability simplex with temporal structure \\
\hline
\end{tabularx}
\end{table}

\subsection{Multivariate Prediction}
\label{sec:multpred}
\subsubsection{Datasets}
The Air Quality dataset \citep{DEVITO2008750} contains the responses of a gas multisensor device deployed in an Italian city. Hourly averaged responses were recorded, along with gas concentration references from a certified analyzer. Precise estimation of pollutant distribution in cities is important for traffic management to help reduce respiratory illnesses, including cancer, in citizens \citep{DEVITO2008750}. The dataset consists of 9358 instances of hourly averaged responses from five metal oxide chemical sensors embedded in an air quality chemical multisensor device. The data was recorded from March 2004 to February 2005, and represents the longest freely available recordings of on-field responses from deployed air quality chemical sensor devices \citep{DEVITO2008750}. The goal in this work is to predict absolute humidity from the remaining variables in a multivariate regression problem.

The Appliances Energy Prediction dataset \citep{CANDANEDO201781} has been employed in this study. The energy consumption of appliances represents a significant portion of the aggregated electricity demand of the residential sector (up to 30\% \citep{CETIN2014716}), and is also important for power management of the grid \citep{ARGHIRA2012128}. 
The dataset spans approximately 4.5 months with 10-minute intervals. House temperature and humidity conditions were monitored using a ZigBee wireless sensor network, where each node transmitted data every 3.3 minutes. This data was then averaged into 10-minute intervals. Energy consumption data was recorded every 10 minutes using m-bus energy meters. Weather data from the nearest weather station (Chievres Airport, Belgium) was retrieved from a public dataset on Reliable Prognosis and combined with the experimental data using the date and time columns. Additionally, two random variables were added to the dataset for testing regression models and filtering out non-predictive attributes \citep{CANDANEDO201781}. \par

\subsubsection{s-BFN Variants}

The centres \(\boldsymbol{C}_j\) and scales \(\gamma_j\) of the radial units can be defined in multiple ways during iterative training. For the tabular regression experiments, three variants, denoted \(G_1\), \(G_2\), and \(G_3\), are summarized in Table~\ref{tab:sbfn_variants}. In \(G_1\), centres and scales are computed from all past predictions of each base predictor up to iteration \(i\). In \(G_2\), only the most recent \(k\) predictions are considered. Finally, \(G_3\) uses the same centre definition as \(G_2\), but sets the scale according to the maximum dispersion across predictors, yielding a radial response governed by cross-predictor separation.

\begin{table}[t]
\centering
\footnotesize
\captionsetup{justification=centering}
\caption{Iterative s-BFN variants with different centre (\(\boldsymbol{C}_j\)) and scale (\(\gamma_j\)) definitions. 
\(i=1,\dots,N\) denotes training iterations, \(j=1,\dots,M\) the predictors, and \(k\) a number of recent predictions.}
\label{tab:sbfn_variants}
\begin{tabular}{c|c|c|c}
\textbf{Variant} & \textbf{Centre \(\boldsymbol{C}_{j}\)} & \textbf{Scale \(\gamma_{j}\)} & \textbf{Basis function \(\varphi\)} \\
\hline
\(G_1\) & 
\(\frac{1}{i}\sum_{q=1}^{i} f_{\boldsymbol{\theta}_j}(\boldsymbol{x}_q)\) 
& 
\(\sqrt{\frac{1}{i-1}\sum_{q=1}^{i}\!\left(f_{\boldsymbol{\theta}_j}(\boldsymbol{x}_q)-\boldsymbol{C}_{j}\right)^2}\) 
& 
\(\exp\!\Big(-\frac{\|f_{\boldsymbol{\theta}_j}(\boldsymbol{x}_i)-\boldsymbol{C}_{j}\|^2}{2\gamma_{j}^2}\Big)\) \\

\(G_2\) & 
\(\frac{1}{k}\sum_{q=i-k+1}^{i} f_{\boldsymbol{\theta}_j}(\boldsymbol{x}_q)\) 
& 
\(\sqrt{\frac{1}{k-1}\sum_{q=i-k+1}^{i}\!\left(f_{\boldsymbol{\theta}_j}(\boldsymbol{x}_q)-\boldsymbol{C}_{j}\right)^2}\) 
& 
\(\exp\!\Big(-\frac{\|f_{\boldsymbol{\theta}_j}(\boldsymbol{x}_i)-\boldsymbol{C}_{j}\|^2}{2\gamma_{j}^2}\Big)\) \\

\(G_3\) & 
\(\frac{1}{k}\sum_{q=i-k+1}^{i} f_{\boldsymbol{\theta}_j}(\boldsymbol{x}_q)\) 
& 
\(\frac{\max_{p\neq j}\, d\!\left(f_{\boldsymbol{\theta}_j}(\boldsymbol{x}_i),f_{\boldsymbol{\theta}_p}(\boldsymbol{x}_i)\right)}{\sqrt{M}}\) 
& 
\(\frac{\|f_{\boldsymbol{\theta}_j}(\boldsymbol{x}_i)-\boldsymbol{C}_{j}\|^2}{\gamma_{j}}\) \\
\end{tabular}
\end{table}

In practice, these variants trade off stability and adaptability: $G_1$ smooths across the full training history, $G_2$ adapts more quickly to local dynamics, and $G_3$ enforces stronger diversity through cross-predictor dispersion. Complementary evidence on alternative centre-placement strategies (Gaussian vs.\ KMeans-derived) is provided in~\ref{app:bf_compare}.

\subsubsection{Models Performance $\&$ Comparisons}
\label{ModelsComparison}

For the individual predictors, for the first set of experiments based on multivariate regression on tabular datasets, low-capacity models are compared. One and two-layer feedforward networks were tested, and results are shown for the two-layer case with the number of neurons in each layer as $\upsilon$, learning rates $\boldsymbol{\eta_\theta}$ for the predictors, multiplicative factor of the initial parameters $\chi$, and regularization parameters $\lambda_1$. For the s-BFN model, the number of predictors or hypotheses is given by $M$, \textcolor{black}{diversity-control parameter} $\varepsilon$, and s-BFN regularization parameters $\lambda_2$. All values used for the hyperparameters are displayed in Table \ref{hyper}.

For the multivariate prediction experiments, the top-performing model versions from the original papers of the two datasets used~\citep{DEVITO2008750,WB:2014,CANDANEDO201781} are replicated for comparison (top competitors). These include Random Forest (RF), Gradient Boosting (Gboost), and Support Vector Machine with radial basis function kernel (SVM-RBF). For the proposed s-BFN, experiments are performed with 10 simulations per configuration, and the mean and standard deviation of the RMSE across the 10 folds are recorded as performance measures. In total, the experiments comprise 80 different hyperparameter configurations, including the single-hypothesis case \(M=1\). Additionally, benchmarking results using the baseline multiple-hypothesis prediction model with an arithmetic combiner~\citep{rupprecht2017learning} are included, where ensemble predictions are obtained by averaging the individual hypotheses induced by Voronoi tessellations.

In the tabular regression experiments, rather than performing an exhaustive search over the full combinatorial space listed in Table~\ref{hyper}, hyperparameters were explored using a structured grid strategy. Ensemble size \(M\), \textcolor{black}{diversity-control parameter} \(\varepsilon\), and aggregation capacity \(\kappa\) (implemented via the number of basis functions \(K = \kappa M\)) were treated as the primary axes of variation, while learning rates and regularization parameters were fixed after preliminary tuning. This design resulted in 80 evaluated configurations across datasets, enabling controlled comparisons while limiting confounding interactions among unrelated hyperparameters.

\begin{table}[t]
\centering
\captionsetup{justification=centering}
\caption{Hyperparameter search space used for tuning. $M$ is the number of hypotheses, $\upsilon$ the number of neurons per layer, $\boldsymbol{\eta_\theta}$ and $\eta_\alpha$ the learning rates for base learners and s-BFN respectively, $\chi$ a scaling factor for initializing predictor parameters $\boldsymbol{\Theta}$, $\varepsilon$ the diversity control parameter, and $\lambda_1$, $\lambda_2$ the base learners and ensemble regularization coefficients respectively.}
\label{tab:hyperparams}
\begin{tabular}{l|p{8cm}} 
\toprule
\textbf{Hyperparameter} & \textbf{ Values} \\
\midrule
Number of hypotheses ($M$) & 2, 5, 10, 20, 35 \\
Neurons per layer ($\upsilon$) & 20, 50, 200, 2000 \\
Learning rates ($\boldsymbol{\eta_\theta}$, $\eta_\alpha$) & 0.03, 0.3, 1.0 \\
Initialization scale ($\chi$) & 0.0001, 0.01, 0.1, 1.0 \\
\textcolor{black}{Diversity-control parameter} ($\varepsilon$) & 0, 0.1, 0.35, 0.5 \\
Regularization (predictors, $\lambda_1$) & 0, 0.0001, 0.01, 0.07 \\
Regularization (s-BFN, $\lambda_2$) & 0, 3, 5, 7 \\
\bottomrule
\end{tabular}
\label{hyper}
\end{table}

\paragraph{Absolute Humidity Prediction}
For the Air Quality dataset, results are reported in Table~\ref{tableauc_test}. The best-performing model in terms of generalization is the least-squares s-BFN when hyperparameters are tuned. The SVM-RBF is the second-best-performing model. The arithmetic combiner exhibits the lowest standard deviation and, consequently, the smallest variation of the mean RMSE across quartiles. The s-BFN attains mean RMSE values lower than all competing models except the SVM-RBF for approximately one quarter of the 80 evaluated hyperparameter configurations, reflecting its higher variability.

\paragraph{Energy Appliance Prediction}

For the energy appliance dataset, the same set of experiments was performed as for the air quality dataset, and results are shown in Table \ref{tableenergy_test}.  The arithmetic combiner is the model with the lowest standard deviation, and the SVM-RBF is the best-performing model among the competitors, consistent with the air absolute humidity prediction experiments.

Thus, for both tabular datasets, the results indicate that s-BFN can achieve strong generalization performance across the evaluated hyperparameter configurations, while also exhibiting higher variability than some classical baselines.

\subsubsection{Diversity and Generalization Performance}
\label{Divexpsec}

In this section, we examine how the generalization performance of s-BFN varies across different values of the \textcolor{black}{diversity-control parameter} \(\varepsilon\) and the number of hypotheses \(M\) on tabular datasets. Figures~\ref{fig:diversity_auc} and~\ref{fig:diversity_energy} report the performance on the air quality and energy appliances test sets, respectively, as a function of the \textcolor{black}{diversity-control parameter} $\varepsilon$ for fixed ensemble sizes $M$.
For each $(M,\varepsilon)$ pair, results are aggregated by averaging over all remaining hyperparameter configurations, yielding one representative value per pair and isolating the effect of $\varepsilon$ from other factors. The horizontal axis represents the \textcolor{black}{diversity-control parameter} $\varepsilon$, while separate curves correspond to different values of $M$.

For the air quality dataset (Figure~\ref{fig:diversity_auc}), the results show that the impact of diversity depends on the ensemble size.
For moderate values of $M$, increasing $\varepsilon$ can lead to improved generalization performance, while for larger ensemble sizes the effect saturates or becomes less pronounced.
The KMeans-SGD s-BFN ($G_3$ in Table \ref{tab:sbfn_variants}) exhibits smoother and more stable trends with respect to $\varepsilon$, whereas the closed-form variant shows increased sensitivity to the choice of $\varepsilon$, particularly for larger $M$.

For the energy appliances dataset (Figure~\ref{fig:diversity_energy}), a similar behavior is observed.
For moderate ensemble sizes, increasing the \textcolor{black}{diversity-control parameter} can improve generalization performance, while for very small ensembles ($M=2$) the effect of diversity is limited.
For larger values of $M$, performance tends to saturate, indicating that beyond a certain ensemble size additional diversity yields diminishing returns.
These results support the hypothesis that, for a given number of hypotheses, there exists an appropriate range of diversity values that balances complementarity among predictors and overall stability.

Taken together, these results indicate that on low-complexity tabular datasets, the benefits of increasing diversity are inherently limited by the expressive capacity required to exploit it. As ensemble size $M$ grows, performance gains saturate once the predictive variability introduced by $\varepsilon$ exceeds what the task complexity can effectively absorb. This behavior is consistent with a capacity--complexity--diversity trade-off: in simple settings, additional hypotheses and diversity do not translate into proportional generalization gains. In more complex prediction regimes, examined in the following sections, increased diversity requires correspondingly higher aggregation capacity to be beneficial, a pattern that becomes more pronounced for image and video data.

{\color{black}
From a practical perspective, this earlier saturation also indicates that the additional computational cost of s-BFN may not always be justified on simpler tabular problems, where lightweight baselines can already achieve competitive performance. Its additional complexity is therefore more relevant in regimes where complementary predictors provide measurable generalization gains.
}

\begin{table}[t!]
\centering
\captionsetup{justification=centering}
\caption{Absolute humidity prediction: statistics of the 10-fold validation test RMSE across all hyperparameter combinations. For each configuration, RMSE values are averaged across the 10 folds; the table reports the best configuration-level mean RMSE and dispersion statistics across configurations (Q: quartiles).}
\label{tableauc_test}
\begin{tabular}{lcccc}
\toprule
\textbf{Model} & \textbf{Best mean RMSE} & \textbf{Std. dev.} & \textbf{1st Q} & \textbf{3rd Q} \\
\midrule
\textbf{SVM-RBF}             & 29.83          & 1.99  & 34.80          & 37.65  \\
\textbf{Random Forest}       & 55.66          & 15.47 & 69.00          & 91.55  \\
\textbf{Gradient Boosting}   & 55.76          & 38.73 & 93.92          & 151.58 \\
\textbf{Arithmetic Combiner} & 39.19          & 0.15  & 41.75          & 43.93  \\
\textbf{s-BFN least-squares} & \textbf{22.46} & 9.14  & \textbf{38.98} & 54.71  \\
\bottomrule
\end{tabular}
\end{table}

\begin{table}[t]
\centering
\captionsetup{justification=centering}
\caption{Energy appliance prediction: statistics of the 10-fold validation test RMSE across all hyperparameter combinations. For each configuration, RMSE values are averaged across the 10 folds; the table reports the best configuration-level mean RMSE and dispersion statistics across configurations (Q: quartiles).}
\label{tableenergy_test}
\begin{tabular}{lcccc}
\toprule
\textbf{Model} & \textbf{Best mean RMSE} & \textbf{Std. dev.} & \textbf{1st Q} & \textbf{3rd Q} \\
\midrule
\textbf{SVM-RBF}             & 104.68           & 1.27  & 107.26          & 109.01 \\
\textbf{Random Forest}       & 298.46           & 29.35 & 328.48          & 373.81 \\
\textbf{Gradient Boosting}   & 292.08           & 67.26 & 389.10          & 476.89 \\
\textbf{Arithmetic Combiner} & 115.17           & 0.11  & 128.54          & 144.83 \\
\textbf{s-BFN least-squares} & \textbf{101.12}  & 8.26  & \textbf{102.25} & \textbf{103.64} \\
\bottomrule
\end{tabular}
\end{table}

\begin{figure}[t!]
\centering
\captionsetup{justification=centering}
\setlength\abovecaptionskip{0\baselineskip}

\begin{subfigure}{0.48\textwidth}
    \centering
    \includegraphics[width=\linewidth]{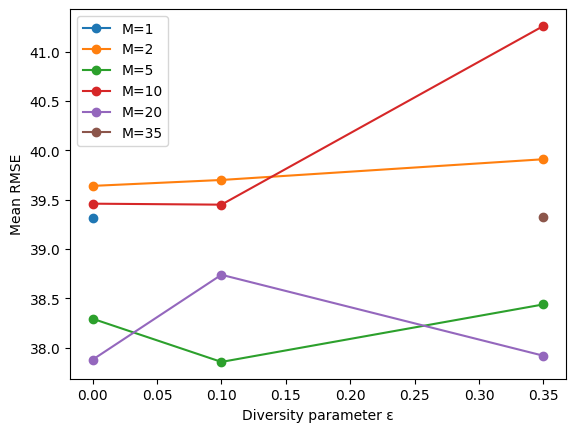}
    \caption{KMeans-SGD s-BFN ($G_3$).}
    \label{fig:auc_kmeans}
\end{subfigure}
\hfill
\begin{subfigure}{0.48\textwidth}
    \centering
    \includegraphics[width=\linewidth]{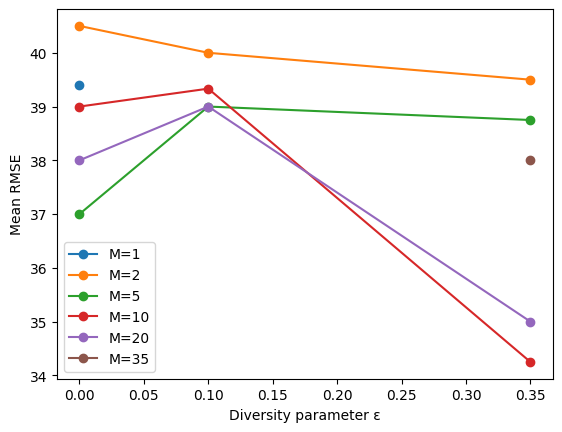}
    \caption{Closed-form s-BFN.}
    \label{fig:auc_closed}
\end{subfigure}

\caption{
Mean RMSE as a function of the \textcolor{black}{diversity-control parameter} $\varepsilon$ for different ensemble sizes $M$ on the air quality dataset.
For each $(M,\varepsilon)$ pair, results are aggregated by averaging over all remaining hyperparameter settings,
ensuring a clear and controlled assessment of the effect of $\varepsilon$.
}
\label{fig:diversity_auc}
\end{figure}

\begin{figure}[t!]
\centering
\captionsetup{justification=centering}
\setlength\abovecaptionskip{0\baselineskip}

\begin{subfigure}{0.48\textwidth}
    \centering
    \includegraphics[width=\linewidth]{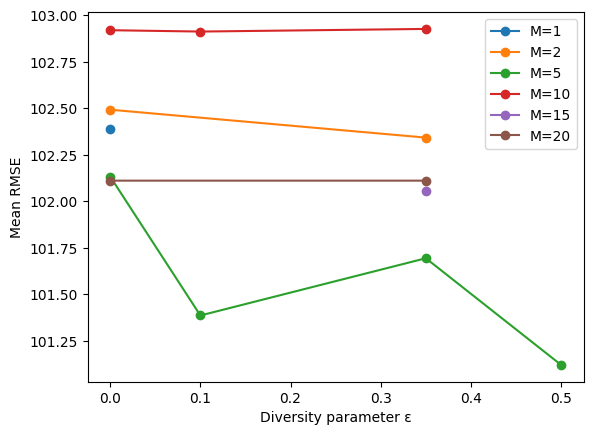}
    \caption{KMeans-SGD s-BFN ($G_3$).}
    \label{fig:energy_kmeans}
\end{subfigure}
\hfill
\begin{subfigure}{0.48\textwidth}
    \centering
    \includegraphics[width=\linewidth]{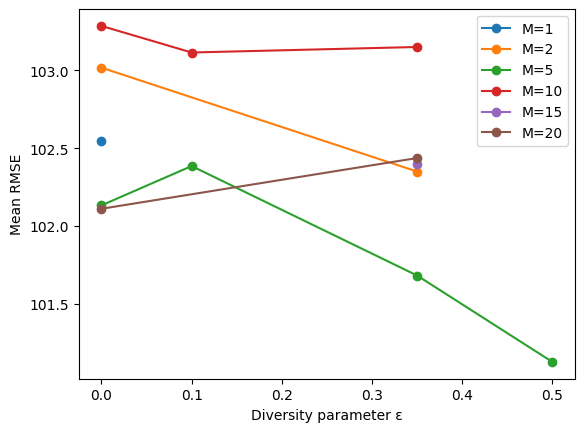}
    \caption{Closed-form s-BFN.}
    \label{fig:energy_closed}
\end{subfigure}

\caption{
Mean RMSE as a function of the \textcolor{black}{diversity-control parameter} $\varepsilon$ for different ensemble sizes $M$ on the energy
appliances dataset.
For each $(M,\varepsilon)$ pair, results are aggregated by averaging over all remaining hyperparameter settings,
ensuring a clear and controlled assessment of the effect of $\varepsilon$.}
\label{fig:diversity_energy}
\end{figure}

\subsection{Multiple Hypothesis Deep Ensembles}
\label{NumericDeeplearning}

\subsubsection{Data}
To examine the relationship between ensemble diversity, model capacity, and data complexity, three benchmark datasets of increasing difficulty were used: MNIST, Fashion-MNIST, and CIFAR-10. MNIST \citep{lecun1998mnist} consists of 28×28 grayscale images of handwritten digits with relatively low intra-class variability. Fashion-MNIST \citep{xiao2017fashion}, which has the same format as MNIST, contains grayscale images of clothing items and is considered more complex due to higher variability in shapes and textures. CIFAR-10 \citep{krizhevsky2009learning} is significantly more challenging, comprising 32×32 color images from ten diverse object classes with substantial variation in background, pose, and lighting. These datasets enable a controlled increase in complexity while keeping class structure consistent across experiments.

\subsection{Deep Base Learners and s-BFN Variants}
The base learners used in the experiments comprise three neural network architectures designed to vary in depth and representational capacity. The first, SimpleCNN, is a compact convolutional neural network with two convolutional layers followed by two fully connected layers, totaling approximately 0.15 million trainable parameters. The second, ResNetTiny1, is a lightweight residual network with three residual stages, block configurations of [1,1,1], and channel widths [16,32,64], resulting in about 0.28 million parameters. The third, ResNetTiny2, increases both depth and capacity with [1,2,2] residual blocks and channels [24,48,96], yielding roughly 0.56 million parameters. Ensembles are constructed either homogeneously, where all base learners share the same architecture, or heterogeneously, where the three architectures are alternated cyclically to promote architectural diversity.\par

For comparison, a Mixture-of-Experts (MoE) combiner is also implemented. While both s-BFN and Mixture-of-Experts employ sample-dependent aggregation, their objectives and inductive biases differ fundamentally. MoE learns unconstrained
gating weights directly in expert-index space to optimize predictive performance. \textcolor{black}{In contrast, s\mbox{-}BFN learns a supervised nonlinear mapping from the structured base-prediction representation to the final prediction, with its architecture motivated by the loss-induced Bregman-centroid geometry of prediction space rather than being trained to reproduce the corresponding centroid \citep{wood2023unified}.} Evaluation is performed across \(n=5\) holdout test splits. For each configuration, we report classification accuracy of individual base learners, the arithmetic
mean ensemble, s-BFN, and MoE. Mean accuracy and 95\% confidence intervals are computed from the empirical variance across splits.

Throughout the experimental analysis, we distinguish between the composition of the ensemble and the aggregation mechanism applied to it. Labels such as ``Homo.\ CNN'', ``Homo.\ ResNet'', and ``Het.\ Mixed'' refer solely to the architectural composition of the base learners (homogeneous or heterogeneous ensembles), while ``s-BFN'', ``MoE'', and ``Mean Aggregation'' denote aggregation rules applied to a given ensemble. The ``Mean Aggregation'' combiner corresponds to logit averaging (i.e., averaging pre-softmax scores followed by a single softmax), as detailed in Section~\ref{sec:ablation-averaging}. The ``Arithmetic Mean'' baseline, in contrast, corresponds to the arithmetic mean of the individual base-learner accuracies and is reported for reference only, rather than as a prediction-space aggregation mechanism.
Architectural composition and aggregation strategy are therefore treated as orthogonal design choices in all reported results.

\subsubsection{\textcolor{black}{Aggregation baselines}}
\label{sec:ablation-averaging}

To extend the original aggregation comparison with stronger directly relevant baselines, the controlled benchmark additionally includes a validation-selected single expert (SingleBest), probability averaging (MeanProb) \citep{lakshminarayanan2017simple}, linear stacking (LinearStack) and a capacity-matched nonlinear stacker (MLPStack-Matched) \citep{WOLPERT1992241}, the stacking and model-averaging variants of Regularized Neural Ensemblers (NE-Stack and NE-MA) \citep{pmlr-v293-arango25a}, and greedy ensemble selection (GES) \citep{Caruana2004Ensemble}. The original ``Mean'' combiner corresponds to LogitAverage in this controlled benchmark.

\textcolor{black}{MLPStack-Matched is a one-hidden-layer nonlinear stacker
mapping the same $MC$-dimensional concatenated expert predictions used by
s-BFN through a ReLU hidden layer of width $h$ to a $C$-dimensional linear
output. For each configuration, $h\geq 4$ is chosen to match the trainable
parameter count of the corresponding s-BFN combiner as closely as an integer
hidden width permits, giving $h(MC+1)+C(h+1)$ parameters. Training uses Adam
with learning rate $10^{-3}$, batch size 512, at most 150 epochs, and
early-stopping patience 20; validation negative log-likelihood is used for
early stopping and model selection.}

To contextualise results under the same loss geometry, a standard averaging baseline is considered alongside the proposed s-BFN and the MoE combiner. The baseline, denoted ``Mean'' in the results, corresponds to logit averaging: pre-softmax scores from the base models are averaged, followed by a single softmax. This scheme is widely used in ensembling and distillation \citep{Hinton2014,DBLP:conf/ijcai/TassiGFT22}. A temperature-scaled variant is optionally tuned on a validation split. \textcolor{black}{In the controlled aggregation benchmark, LogitAverage is evaluated without
additional temperature fitting.} In contrast, s-BFN does not average logits. Each base model's logits are first mapped to class probabilities via softmax, the resulting probability vectors are concatenated into a structured input, passed through a radial-basis feature map, and linearly combined to produce ensemble logits, which are then normalised by a temperature-scaled softmax. Thus, s-BFN implements a learned, loss-aware \textcolor{black}{combiner over probability inputs whose architecture is motivated by the corresponding centroidal geometry}, rather than a fixed averaging rule.

\textcolor{black}{In the controlled aggregation benchmark, the MoE gate is
trained post-hoc on the frozen expert probability outputs; the expert
parameters are not jointly updated during gate training. The gate uses a
hidden ReLU layer followed by a softmax over the $M$ experts to obtain
sample-dependent mixture weights and is trained using the same
combiner-training and validation protocol as the other learned aggregation
methods.}

MoE combines the base probabilities via these learned per-sample mixture
weights, yielding a gated weighted sum of the base probability vectors, with
temperature \(T\) applied where relevant. For reference, the
accuracy-averaging baseline, denoted "Arithmetic Mean" in the results, is
also reported, corresponding to the arithmetic mean of the individual base
learners' accuracies, as in \citet{rupprecht2017learning}. A concise summary
of the ensemble variants, aligned with the names used in the figures and
tables, is provided in Table~\ref{tab:ensemble_variants}.

\begin{table}[t!]
\centering
\caption{Ensemble variants used in the architecture, diversity, and computational efficiency analyses. s-BFN: probability inputs $\to$ RBF features $\to$ linear logits $\to$ softmax with temperature \(T\) (learned, loss-aware combiner). MoE: gated weighted sum of base probabilities. Mean Aggregation: logit averaging (average logits, then a single softmax). Arithmetic Mean: arithmetic mean of base accuracies \citep{rupprecht2017learning}.}
\label{tab:ensemble_variants}
\begin{tabular}{lll}
\toprule
Name in figs/tables & Base learners & Combiners \\
\midrule
"Het.\ Mixed"   & SimpleCNN, ResNetTiny1, ResNetTiny2 & s-BFN, MoE, Mean Agg. \\
"Homo.\ ResNet" & ResNetTiny2 (or ResNetTiny1)        & s-BFN, MoE, Mean Agg. \\
"Homo.\ CNN"    & SimpleCNN                            & s-BFN, MoE, Mean Agg. \\
"Arithmetic Mean"     & Same as corresponding ensemble       & Accuracy avg. (baseline) \\
\bottomrule
\end{tabular}
\end{table}

For the controlled aggregation benchmark, the
\textcolor{black}{diversity-control parameter} $\varepsilon$ is applied only
during upstream expert training. For each fixed
$(\mathrm{dataset},M,\varepsilon,\mathrm{seed})$ configuration, all
aggregation methods subsequently operate on the same frozen expert predictions
and use the same combiner-training, validation, and test partitions. Thus,
the comparison isolates the aggregation mechanism conditional on a common
diversity-controlled expert ensemble; $\varepsilon$ is not supplied to any
aggregation method. The complete configuration-level results are reported
in~\ref{app:full-aggregation-grid}.

For inferential analysis, accuracy is first averaged over the five common seeds within each $(M,\varepsilon)$ configuration, yielding one paired observation per configuration ($n=16$ for MNIST and Fashion-MNIST and $n=12$ for CIFAR-10). A Friedman test across the ten aggregation methods indicates significant differences on MNIST ($\chi_F^2(9)=120.79$, $p=9.20\times10^{-22}$), Fashion-MNIST ($\chi_F^2(9)=125.26$, $p=1.11\times10^{-22}$), and CIFAR-10 ($\chi_F^2(9)=94.02$, $p=2.53\times10^{-16}$). Post-hoc paired two-sided Wilcoxon signed-rank tests compare s\mbox{-}BFN with each competing aggregation rule, with Holm correction over the nine comparisons within each dataset.

\begin{table*}[t]
\color{black}
\centering
\caption{
\color{black}{Configuration-level comparison of s\mbox{-}BFN against competing aggregation rules over the complete controlled grid. The competing baselines are a validation-selected single expert (SingleBest), probability averaging (MeanProb)~\citep{lakshminarayanan2017simple}, logit averaging (LogitAverage)~\citep{DBLP:conf/ijcai/TassiGFT22}, linear stacking (LinearStack) and a capacity-matched nonlinear stacker (MLPStack-Matched)~\citep{WOLPERT1992241}, Mixture-of-Experts (MoE)~\citep{Jacobs1991Adaptive}, the stacking and model-averaging variants of Regularized Neural Ensemblers (NE-Stack and NE-MA)~\citep{pmlr-v293-arango25a}, and greedy ensemble selection (GES)~\citep{Caruana2004Ensemble}. $\Delta$ denotes the mean accuracy difference (s\mbox{-}BFN minus baseline), in percentage points, across the evaluated $(M,\varepsilon)$ configurations. W/T/L denotes the number of configurations in which s\mbox{-}BFN obtains higher/equal/lower mean accuracy. $p_{\mathrm{H}}$ is the paired two-sided Wilcoxon signed-rank $p$-value after Holm correction over the nine comparisons within each dataset. The complete configuration-level accuracies and 95\% confidence intervals over the five seeds are reported in~\ref{app:full-aggregation-grid}. Bold values indicate $p_{\mathrm{H}}<0.05$.}
}
\label{tab:strong-aggregation-baselines}

\resizebox{\textwidth}{!}{%
\begin{tabular}{lccc|ccc|ccc}
\toprule
& \multicolumn{3}{c|}{MNIST ($n=16$)}
& \multicolumn{3}{c|}{Fashion-MNIST ($n=16$)}
& \multicolumn{3}{c}{CIFAR-10 ($n=12$)} \\
Baseline
& $\Delta$ & W/T/L & $p_{\mathrm{H}}$
& $\Delta$ & W/T/L & $p_{\mathrm{H}}$
& $\Delta$ & W/T/L & $p_{\mathrm{H}}$ \\
\midrule

SingleBest
& +13.72 & 16/0/0 & \textbf{$2.75\times10^{-4}$}
& +19.26 & 16/0/0 & \textbf{$2.75\times10^{-4}$}
& +11.72 & 12/0/0 & \textbf{0.0044} \\

MeanProb
& +10.36 & 16/0/0 & \textbf{$2.75\times10^{-4}$}
& +15.76 & 16/0/0 & \textbf{$2.75\times10^{-4}$}
& +7.59 & 12/0/0 & \textbf{0.0044} \\

LogitAverage
& +9.56 & 16/0/0 & \textbf{$2.75\times10^{-4}$}
& +14.59 & 16/0/0 & \textbf{$2.75\times10^{-4}$}
& +7.09 & 12/0/0 & \textbf{0.0044} \\

LinearStack
& -1.33 & 1/1/14 & \textbf{0.0228}
& -0.74 & 2/0/14 & \textbf{0.0126}
& -1.30 & 3/0/9 & 0.3296 \\

MLPStack-Matched
& -0.26 & 7/0/9 & 0.8603
& +0.64 & 11/0/5 & 0.1928
& -0.59 & 5/0/7 & 0.4072 \\

MoE
& +0.16 & 13/0/3 & 0.3252
& +1.80 & 13/0/3 & \textbf{0.0064}
& +0.38 & 8/0/4 & 0.4072 \\

NE-Stack
& +7.89 & 16/0/0 & \textbf{0.0017}
& +11.46 & 16/0/0 & \textbf{$2.75\times10^{-4}$}
& +5.13 & 12/0/0 & \textbf{0.0044} \\

NE-MA
& +9.56 & 16/0/0 & \textbf{$2.75\times10^{-4}$}
& +13.91 & 16/0/0 & \textbf{$2.75\times10^{-4}$}
& +6.41 & 12/0/0 & \textbf{0.0044} \\

GES
& +9.39 & 16/0/0 & \textbf{$2.75\times10^{-4}$}
& +14.50 & 16/0/0 & \textbf{$2.75\times10^{-4}$}
& +5.69 & 12/0/0 & \textbf{0.0044} \\

\bottomrule
\end{tabular}%
}
\end{table*}

Table~\ref{tab:strong-aggregation-baselines} shows that s\mbox{-}BFN achieves consistently higher mean accuracy than SingleBest, MeanProb, LogitAverage, NE-Stack, NE-MA, and GES across every evaluated configuration of all three datasets, with all corresponding Holm-adjusted comparisons significant at the 0.05 level. Relative to MoE, the advantage is significant on Fashion-MNIST but not on MNIST or CIFAR-10. No significant difference is observed between s\mbox{-}BFN and MLPStack-Matched on any dataset. LinearStack is the strongest comparator, with a significant advantage on MNIST and Fashion-MNIST but not on CIFAR-10.

\subsection{Analysis of Ensemble Diversity, Capacity, and Dataset Complexity}
\label{complexity}

This section examines how ensemble diversity, model capacity, and dataset complexity interact to shape classification accuracy and generalization. The objective is analytical rather than competitive; achieving state-of-the-art results is outside the scope. Instead, the focus is on interpreting performance trade-offs introduced by different ensemble configurations under increasing task complexity.

Two families of diagnostics are reported as the
\textcolor{black}{diversity-control} parameter \(\varepsilon\) is varied.
The first follows the bias--variance--diversity perspective grounded in
Bregman geometry \citep{wood2023unified}. Here, the Bregman centroid of the
base predictions is used as a geometric reference in the geometry induced
by the training loss, Euclidean for squared loss and negative entropy on the
probability simplex for cross-entropy.
\textcolor{black}{For the classification diagnostics, the corresponding
left-sided negative-entropy/KL centroid is implemented as the normalized
geometric mean of the base probability predictions, with a positive numerical
stabilizer applied inside the logarithm.}
Two quantities are reported: the error at this centroid, interpreted as bias,
and the dispersion of base predictions around it, interpreted as diversity.
The second adopts the PAC-Bayesian disagreement framework for majority votes
\citep{10.5555/2789272.2831140}, reporting three metrics: Gibbs risk, expected
error of a randomly drawn base learner; expected disagreement between
independently drawn learners; and the ensemble's majority-vote error. Where
applicable, notably under low disagreement, the associated
\(\mathcal{C}\)-bound is plotted as a reference. All metrics are computed on
test sets using the same partitions as the accuracy plots and averaged across
random seeds.

Experimental results on MNIST, Fashion-MNIST, and CIFAR-10 (Figures~\ref{fig:hete_MIST}, \ref{fig:hete_MIST_FASHION}, and \ref{fig:hete_CIFARV2}) show a strong dependence of ensemble performance on prediction diversity. Across the evaluated datasets, increasing \(\varepsilon\) tends to improve accuracy and reduce generalization error, especially in heterogeneous ensembles. The effect strengthens with task complexity (Figures~\ref{diversityMNIST}, \ref{diversityfashionMNIST}, \ref{hetecompCIFAR_diver2}). Consistent with the disagreement analysis, expected disagreement tends to
decrease as $\varepsilon$ increases, while Gibbs risk varies more modestly
and majority-vote error typically exhibits a U-shaped profile with a minimum
at intermediate values.

The optimal ensemble size \(M\) depends on both dataset complexity and diversity. On MNIST, accuracy peaks at moderate \(M\) (Figure~\ref{M_MNIST}); on CIFAR-10, the relation between accuracy and \(M\) is less regular (Figure~\ref{M_CIFAR10}), with capacity gains materializing only when diversity is sufficiently high. Architectural heterogeneity becomes increasingly critical as complexity rises: the performance gap between homogeneous and heterogeneous ensembles widens with task difficulty (Figures~\ref{accuracyMNIST}, \ref{accuracyFashionMNIST}, \ref{accuracyCIFAR10}), and the strongest improvements occur in the intermediate \(\varepsilon\) regime identified above.

Combiner comparison shows that learned aggregation is generally advantageous, although the ranking among s-BFN, logit averaging, and MoE is not identical for every dataset and ensemble composition. The proposed s-BFN combiner achieves the strongest overall performance in most heterogeneous settings and provides clear gains over the arithmetic base-learner average, while MoE remains competitive in some homogeneous configurations. This behavior is consistent with the interpretation that the benefit of learned probability-space aggregation depends on both architectural diversity and the complementarity of base predictions. On CIFAR-10, the distribution of accuracies is less skewed toward the maximum, reflecting reduced confidence under increased complexity. Optimization parameters also influence stability: a learning rate of 0.1 yields stable results (Figure~\ref{hetecompfashionMNIST_lr}), whereas higher rates degrade accuracy and increase variance, particularly in diverse ensembles.

\begin{figure}[t!]
    \centering

    \begin{subfigure}[t]{0.49\textwidth}
        \includegraphics[width=\linewidth]{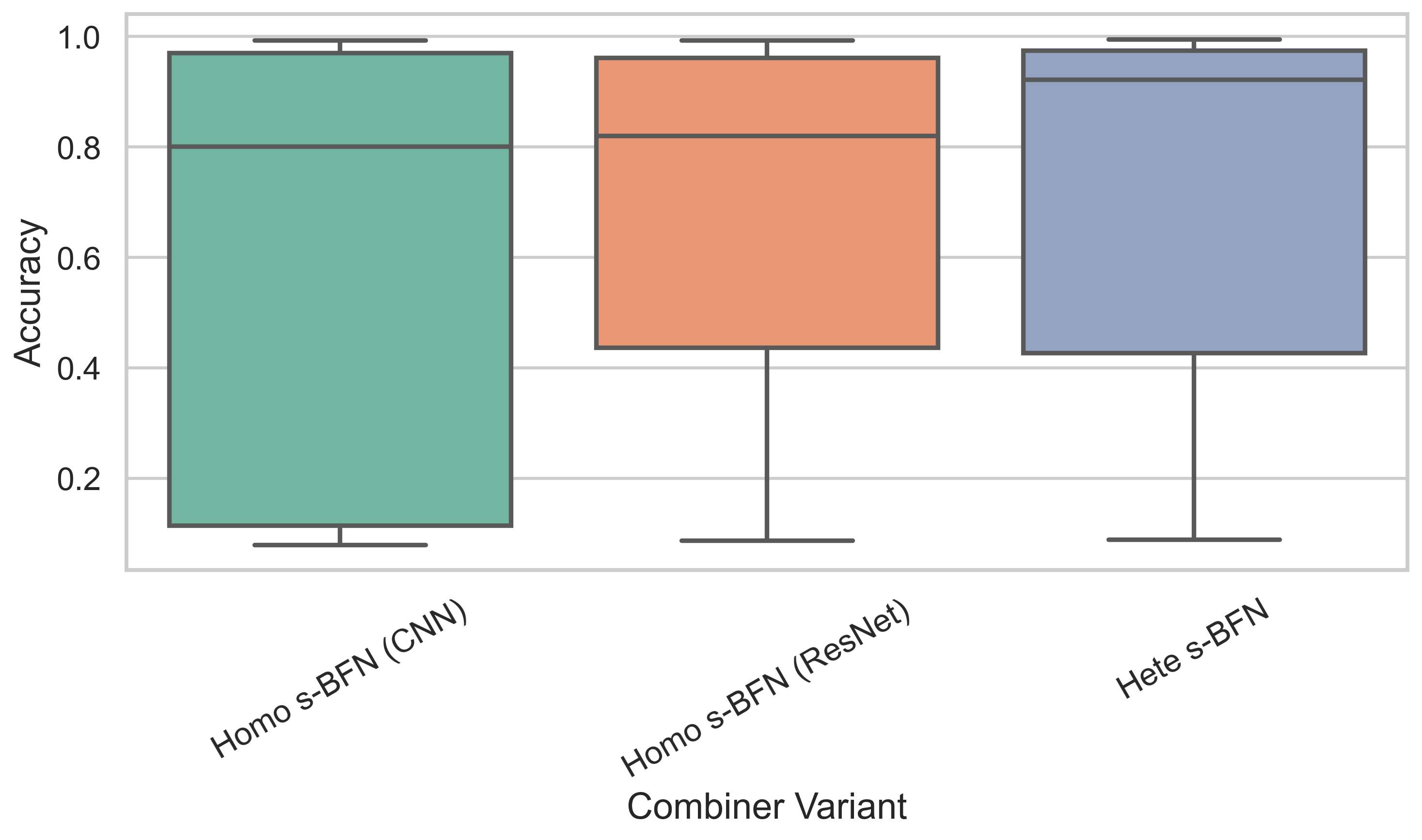}
        \caption{Accuracies of homogeneous variants}
        \label{accuracyMNIST}
    \end{subfigure}
    \hfill
    \begin{subfigure}[t]{0.49\textwidth}
        \includegraphics[width=\linewidth]{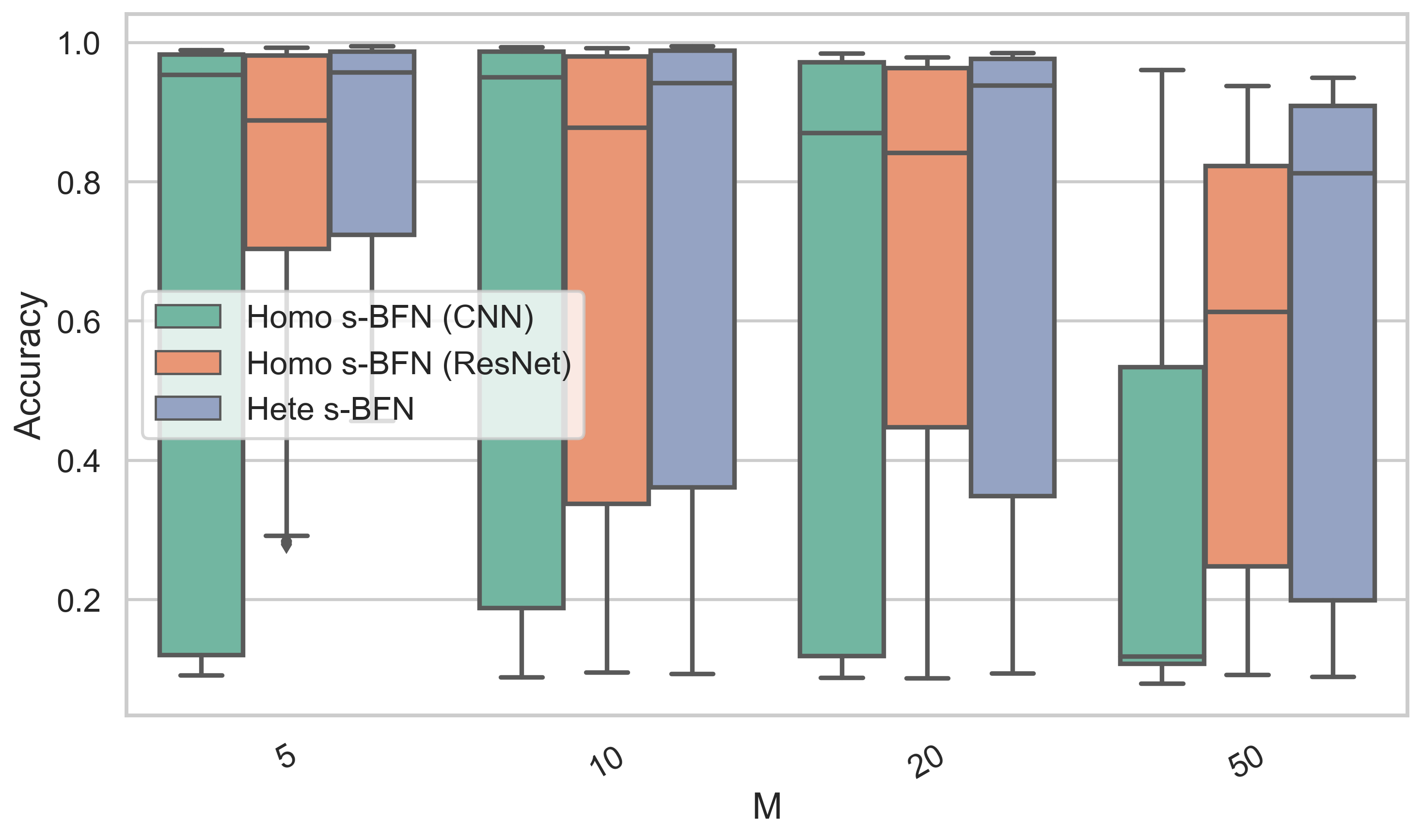}
        \caption{Accuracies by number of learners (M)}
        \label{M_MNIST}
    \end{subfigure}

    \vspace{0.8em}

    \begin{subfigure}[t]{0.49\textwidth}
        \includegraphics[width=\linewidth]{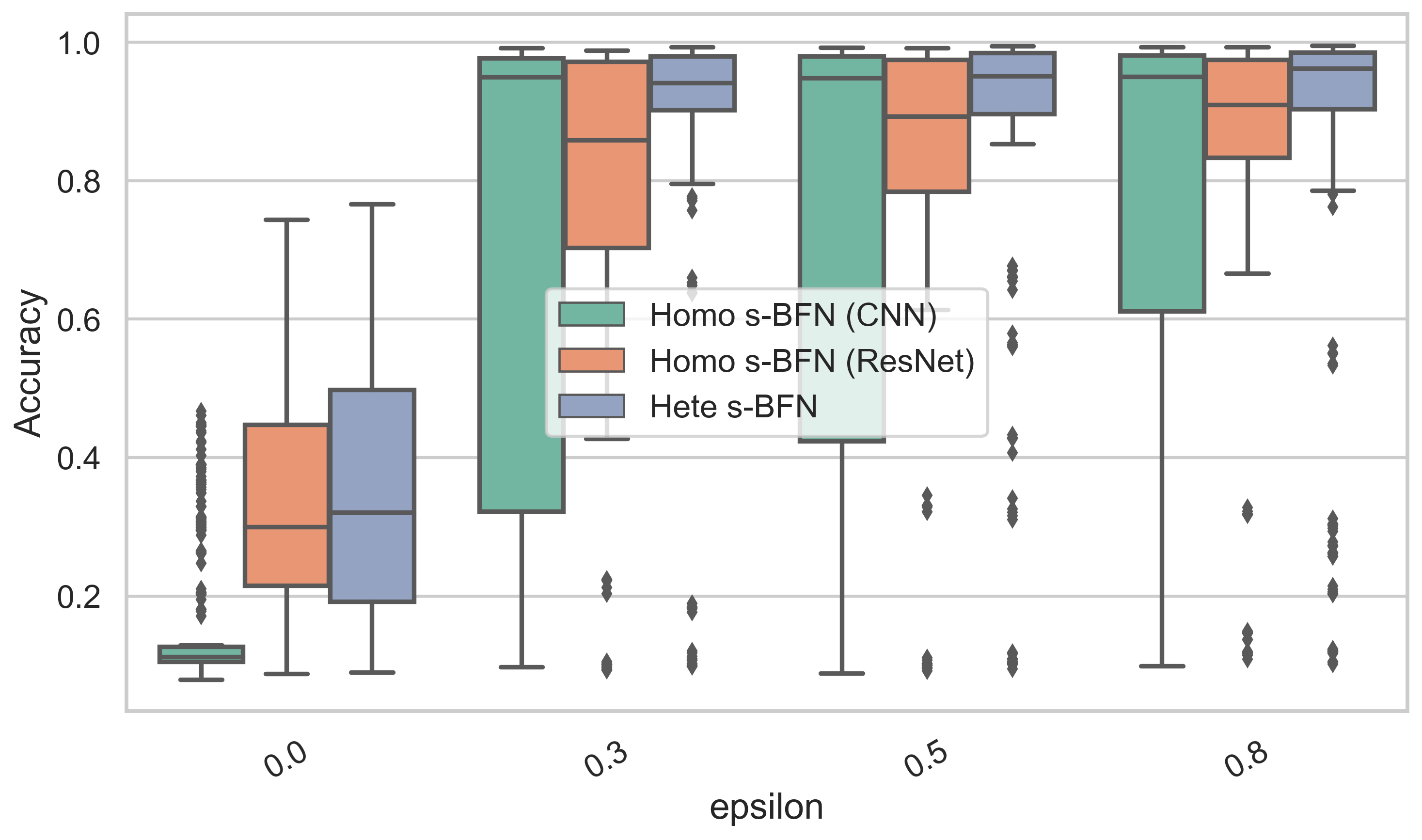}
        \caption{Homogeneous vs.\ heterogeneous by diversity $(\varepsilon)$}
        \label{diversityMNIST}
    \end{subfigure}
    \hfill
    \begin{subfigure}[t]{0.49\textwidth}
        \includegraphics[width=\linewidth]{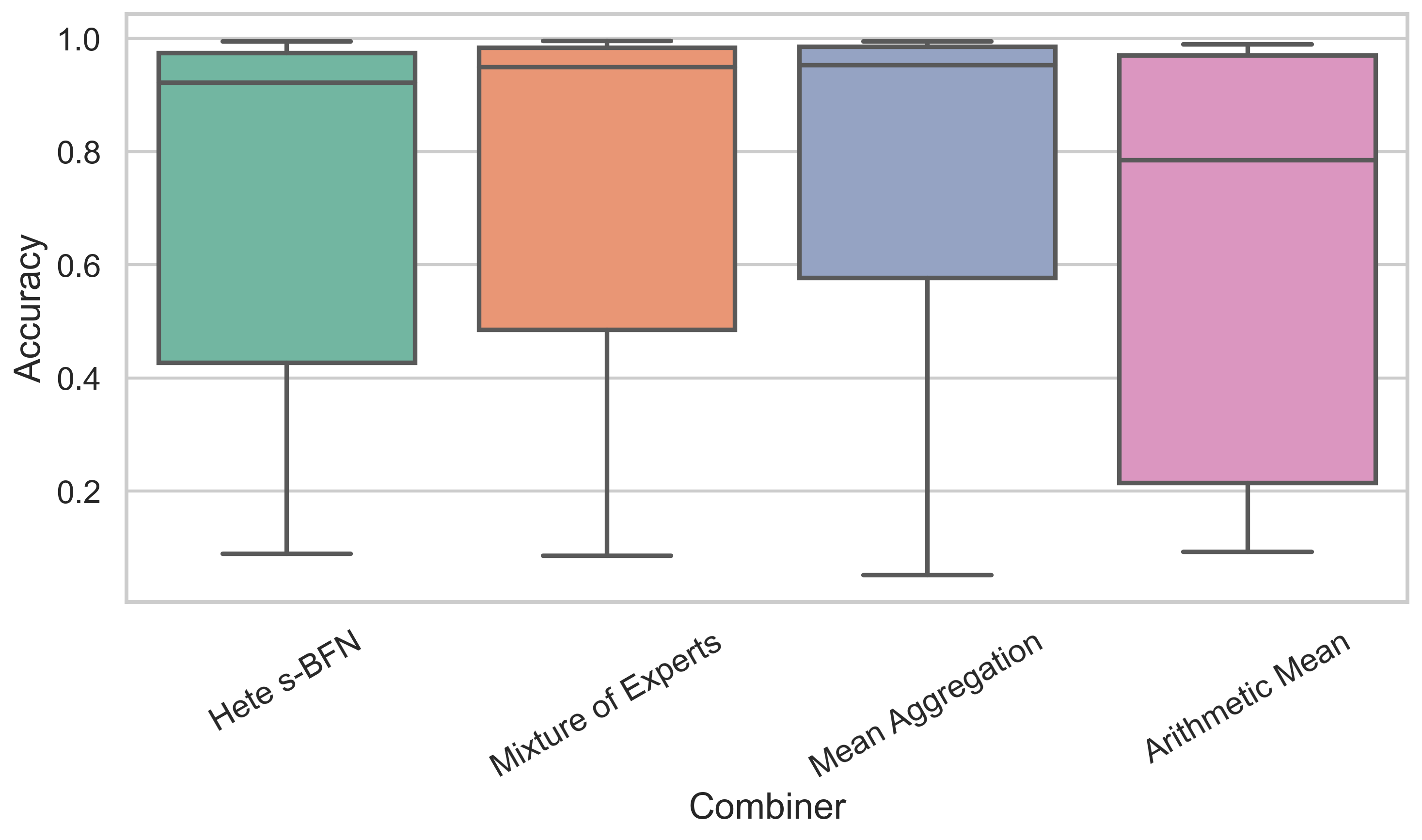}
        \caption{Accuracies of heterogeneous variants}
        \label{hetecompMNIST_M}
    \end{subfigure}

    \vspace{0.8em}

    \begin{subfigure}[t]{0.49\textwidth}
        \includegraphics[width=\linewidth]{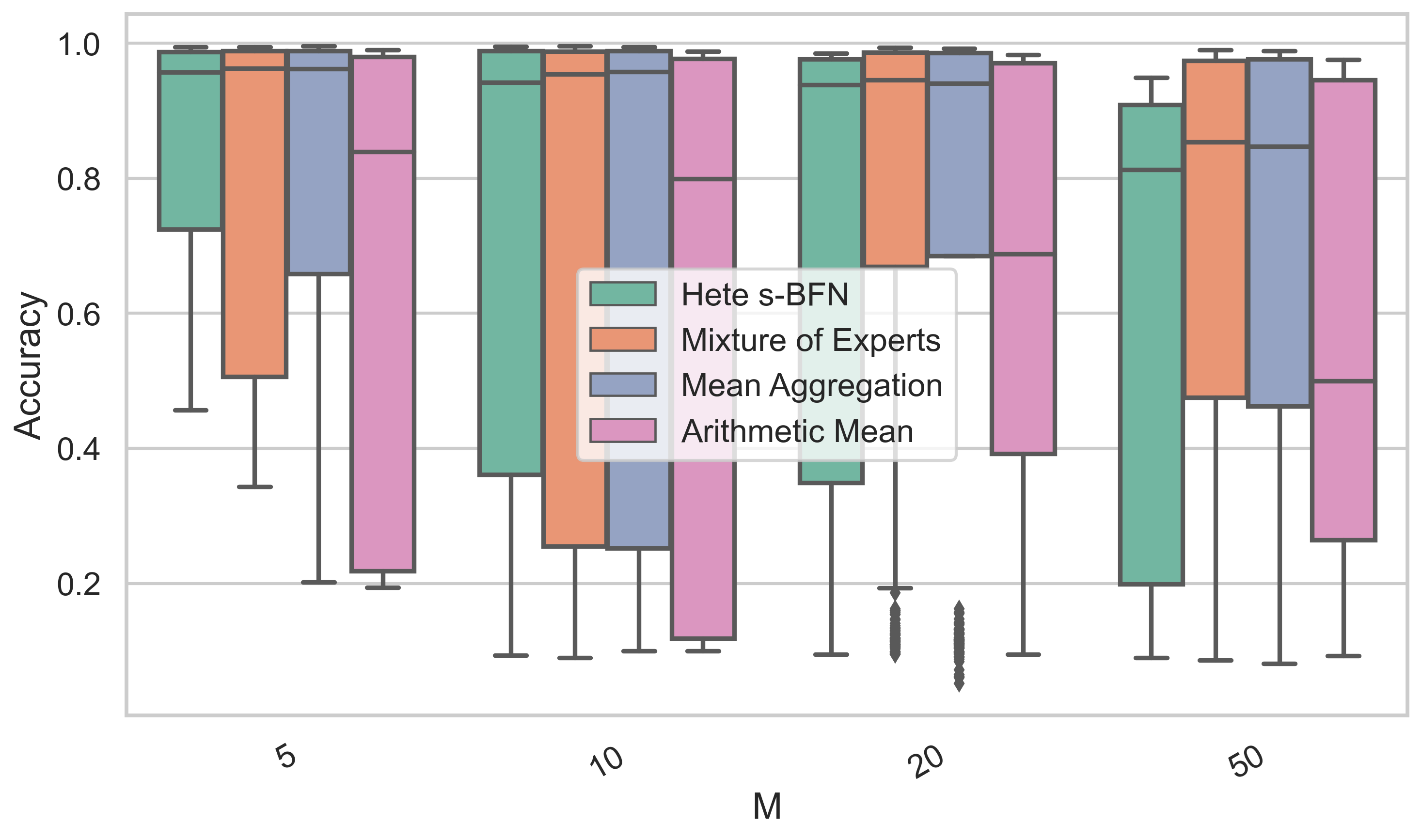}
        \caption{Accuracies of heterogeneous variants by M}
        \label{hetecompMNIST_M_by}
    \end{subfigure}
    \hfill
    \begin{subfigure}[t]{0.49\textwidth}
        \includegraphics[width=\linewidth]{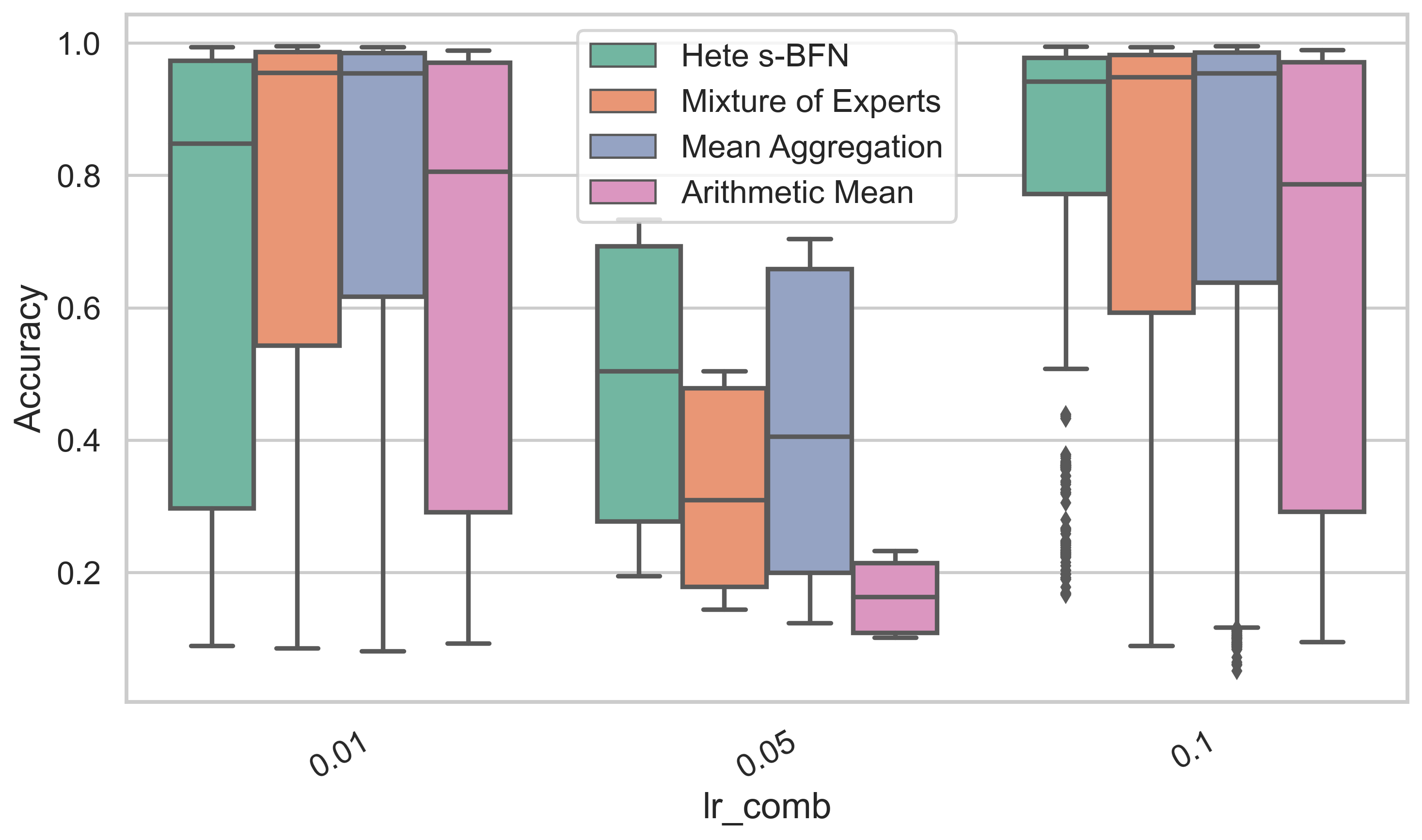}
        \caption{Accuracies of heterogeneous variants by learning rate ($\eta_{\alpha}$)}
        \label{hetecompMNIST_diver2}
    \end{subfigure}

    \caption{Classification accuracy for different ensemble configurations on MNIST.
Ensemble labels indicate base-learner composition (homogeneous or heterogeneous),
while aggregation rules (s-BFN, MoE, Mean Aggregation, Arithmetic Mean) are reported separately.}
    \label{fig:hete_MIST}
\end{figure}

\begin{figure}[t!]
    \centering

    \begin{subfigure}[t]{0.49\textwidth}
        \includegraphics[width=\linewidth]{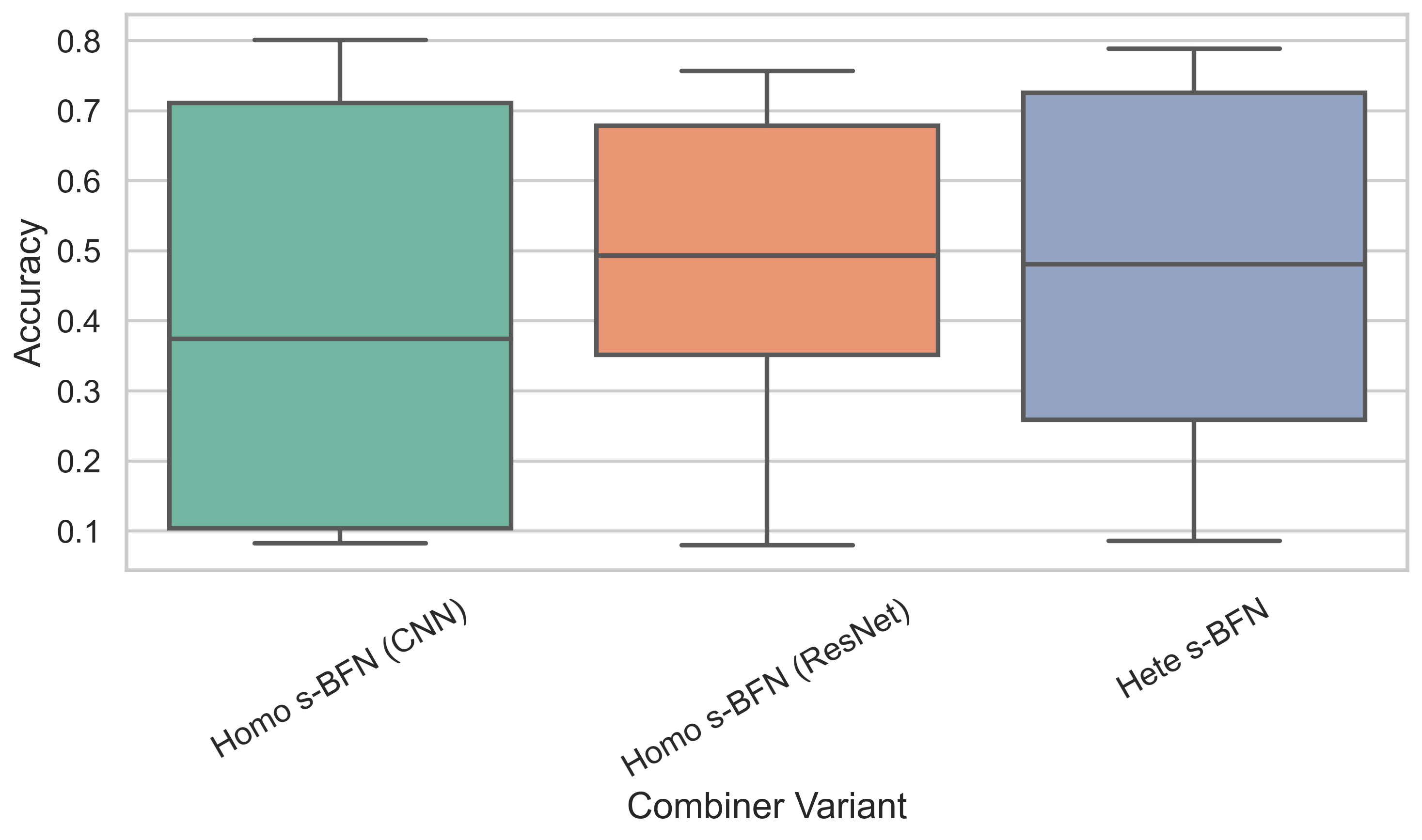}
        \caption{Accuracies of homogeneous variants}
        \label{accuracyFashionMNIST}
    \end{subfigure}
    \hfill
    \begin{subfigure}[t]{0.49\textwidth}
        \includegraphics[width=\linewidth]{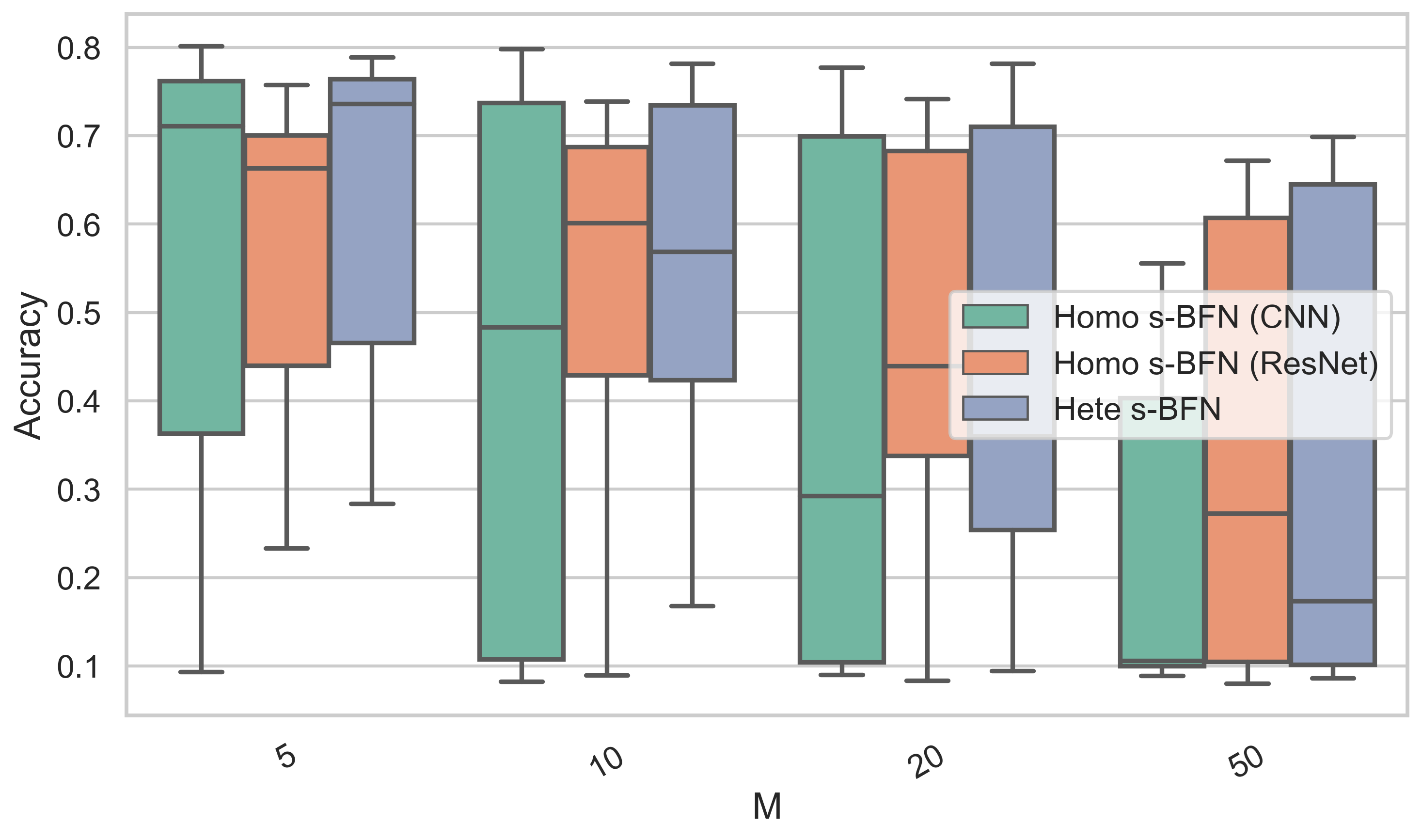}
        \caption{Accuracies by number of learners (M)}
        \label{M_FashionMNIST}
    \end{subfigure}

    \vspace{0.8em}

    \begin{subfigure}[t]{0.49\textwidth}
        \includegraphics[width=\linewidth]{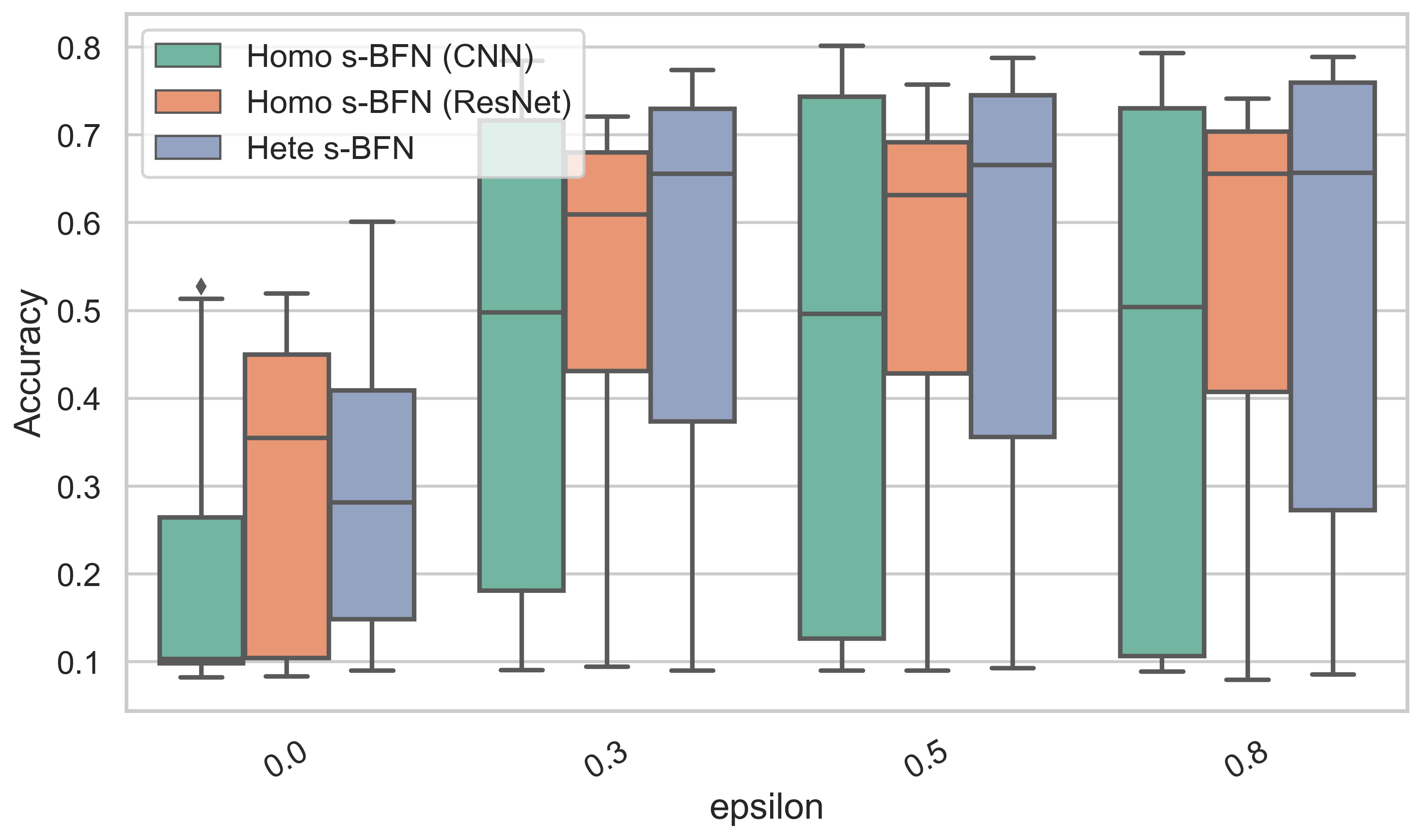}
        \caption{Homogeneous vs.\ heterogeneous by diversity $(\varepsilon)$}
        \label{diversityfashionMNIST}
    \end{subfigure}
    \hfill
    \begin{subfigure}[t]{0.49\textwidth}
        \includegraphics[width=\linewidth]{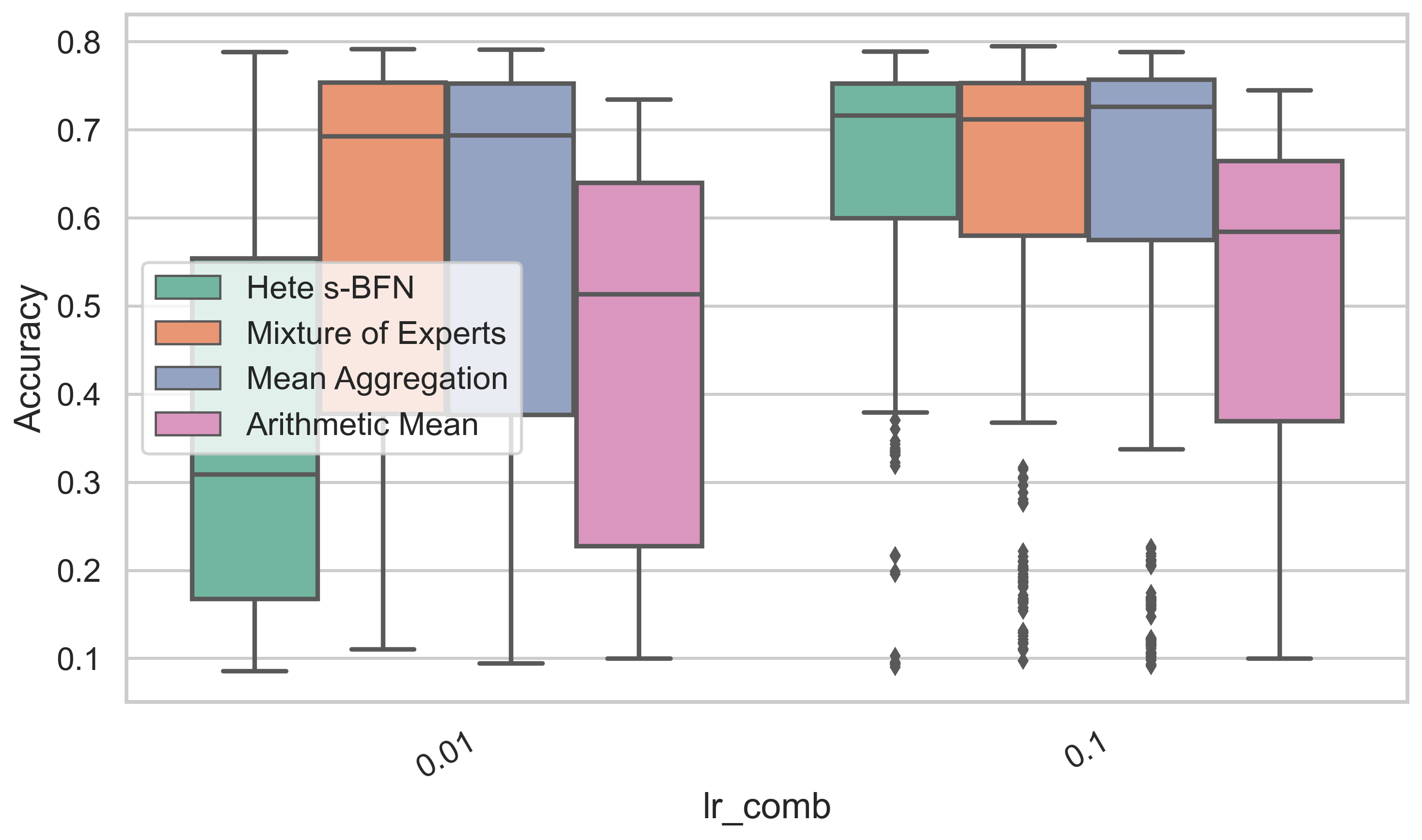}
        \caption{Accuracies of heterogeneous variants by learning rate}
        \label{hetecompfashionMNIST_lr}
    \end{subfigure}

    \vspace{0.8em}

    \begin{subfigure}[t]{0.49\textwidth}
        \includegraphics[width=\linewidth]{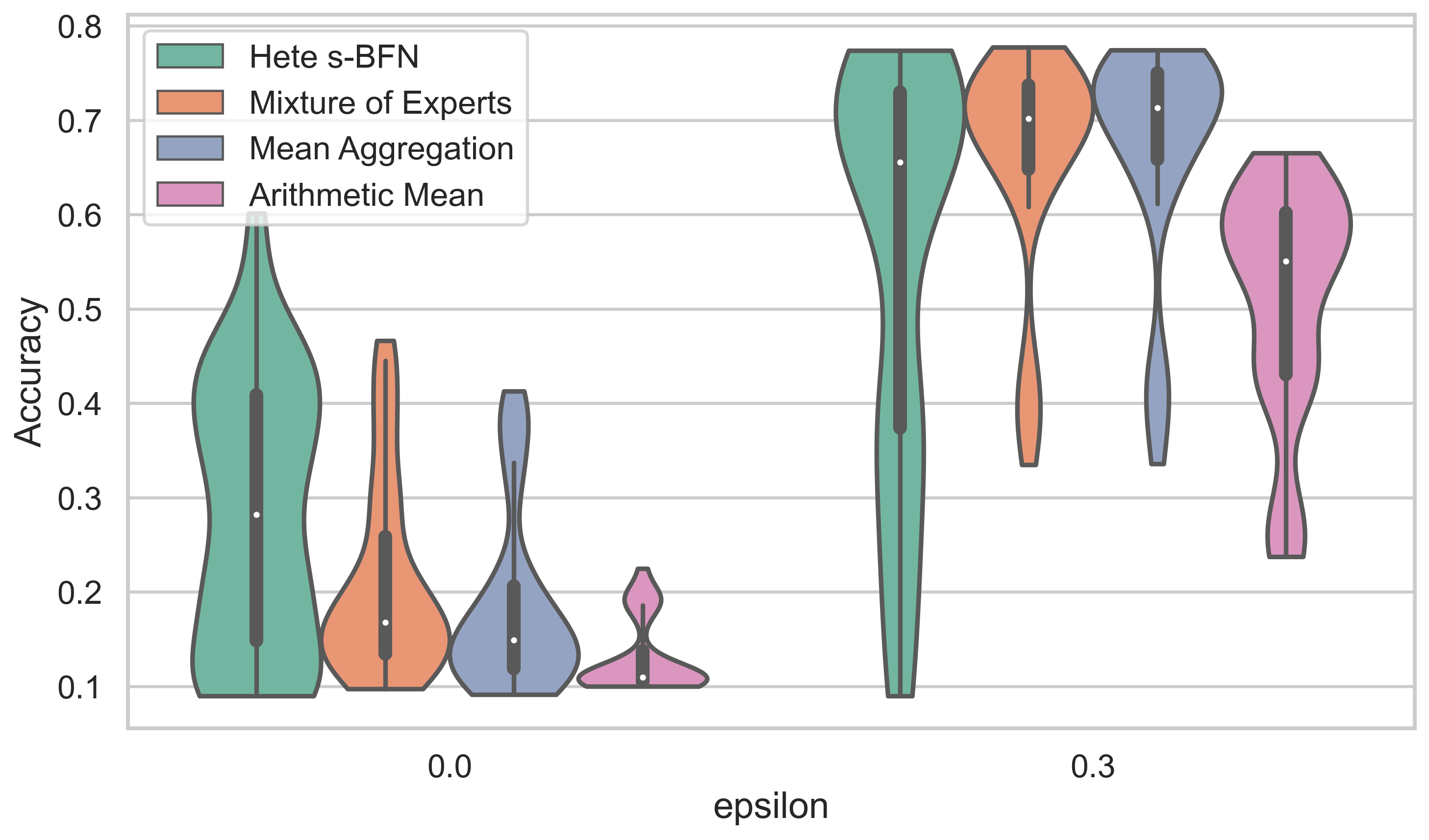}
        \caption{Heterogeneous variants by diversity $(\varepsilon\in[0,0.3])$}
        \label{hetecompfashMNIST_diver1}
    \end{subfigure}
    \hfill
    \begin{subfigure}[t]{0.49\textwidth}
        \includegraphics[width=\linewidth]{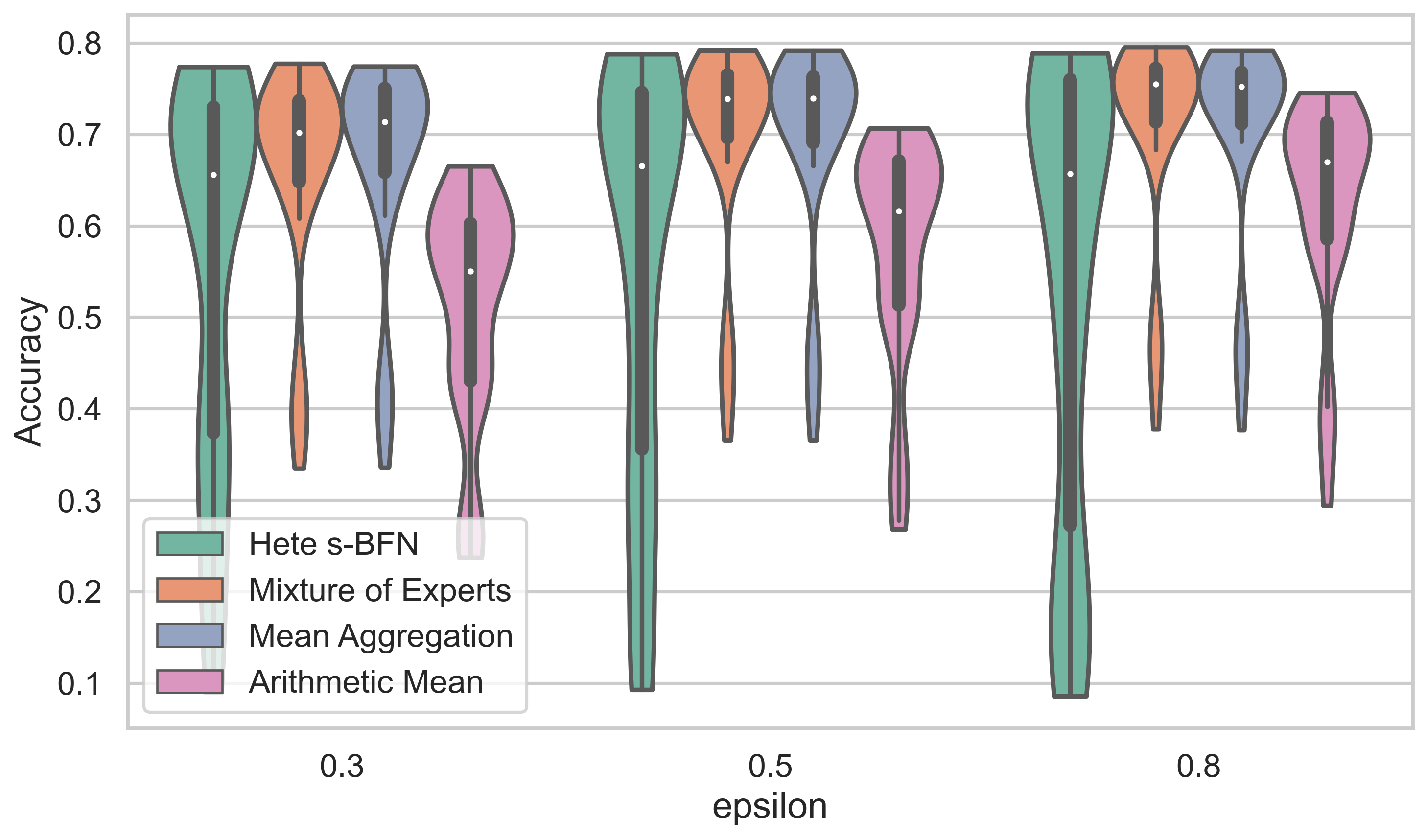}
        \caption{Heterogeneous variants by diversity $(\varepsilon\in[0.3,0.8])$}
        \label{hetecompfashMNIST_diver2}
    \end{subfigure}

    \caption{Classification accuracy for different ensemble configurations on Fashion-MNIST.}
    \label{fig:hete_MIST_FASHION}
\end{figure}

\begin{figure}[t!]
    \centering

    \begin{subfigure}[t]{0.49\textwidth}
        \centering
        \includegraphics[width=\linewidth]{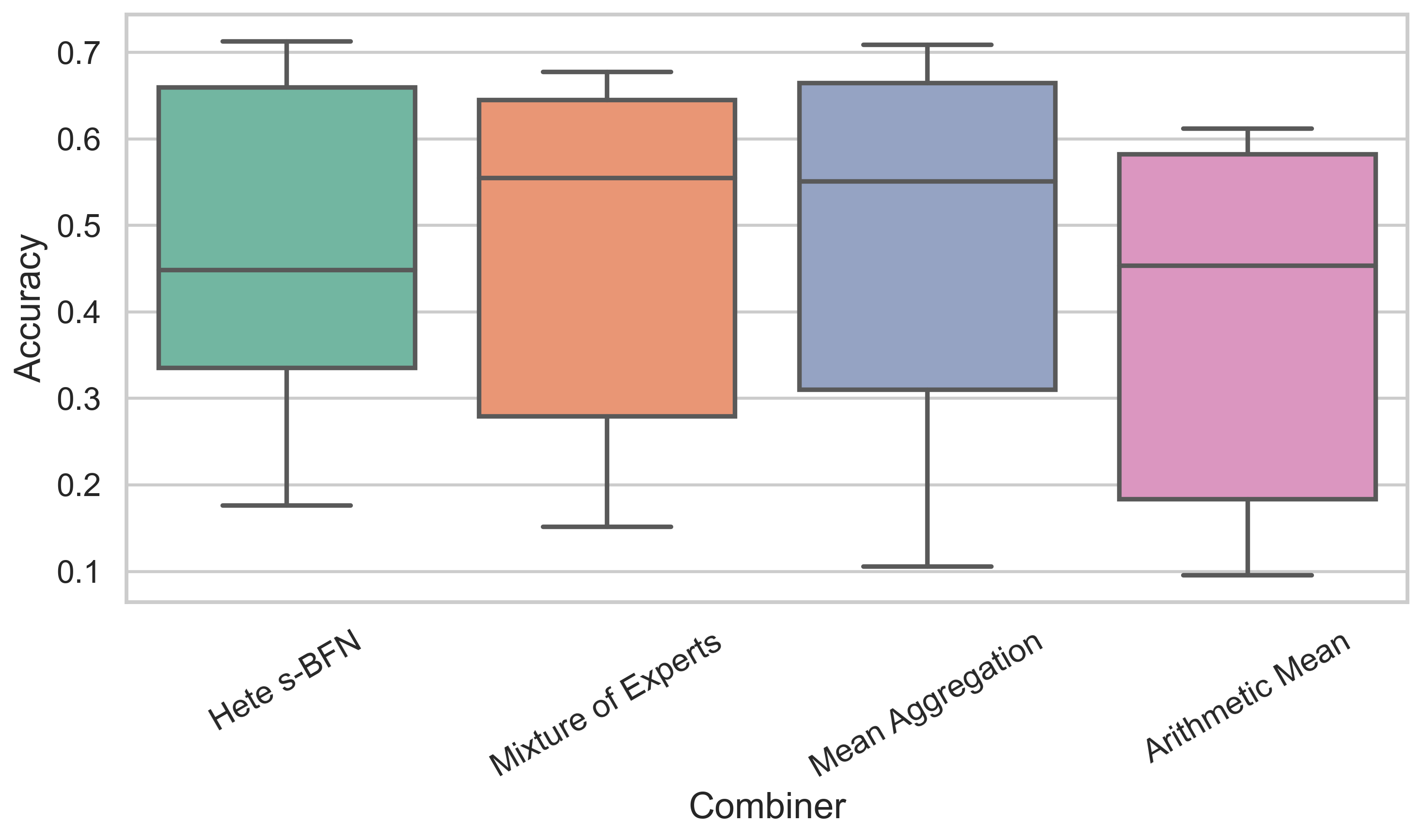}
        \caption{Accuracies of heterogeneous variants vs.\ others}
        \label{accuracyCIFAR10}
    \end{subfigure}
    \hfill
    \begin{subfigure}[t]{0.49\textwidth}
        \centering
        \includegraphics[width=\linewidth]{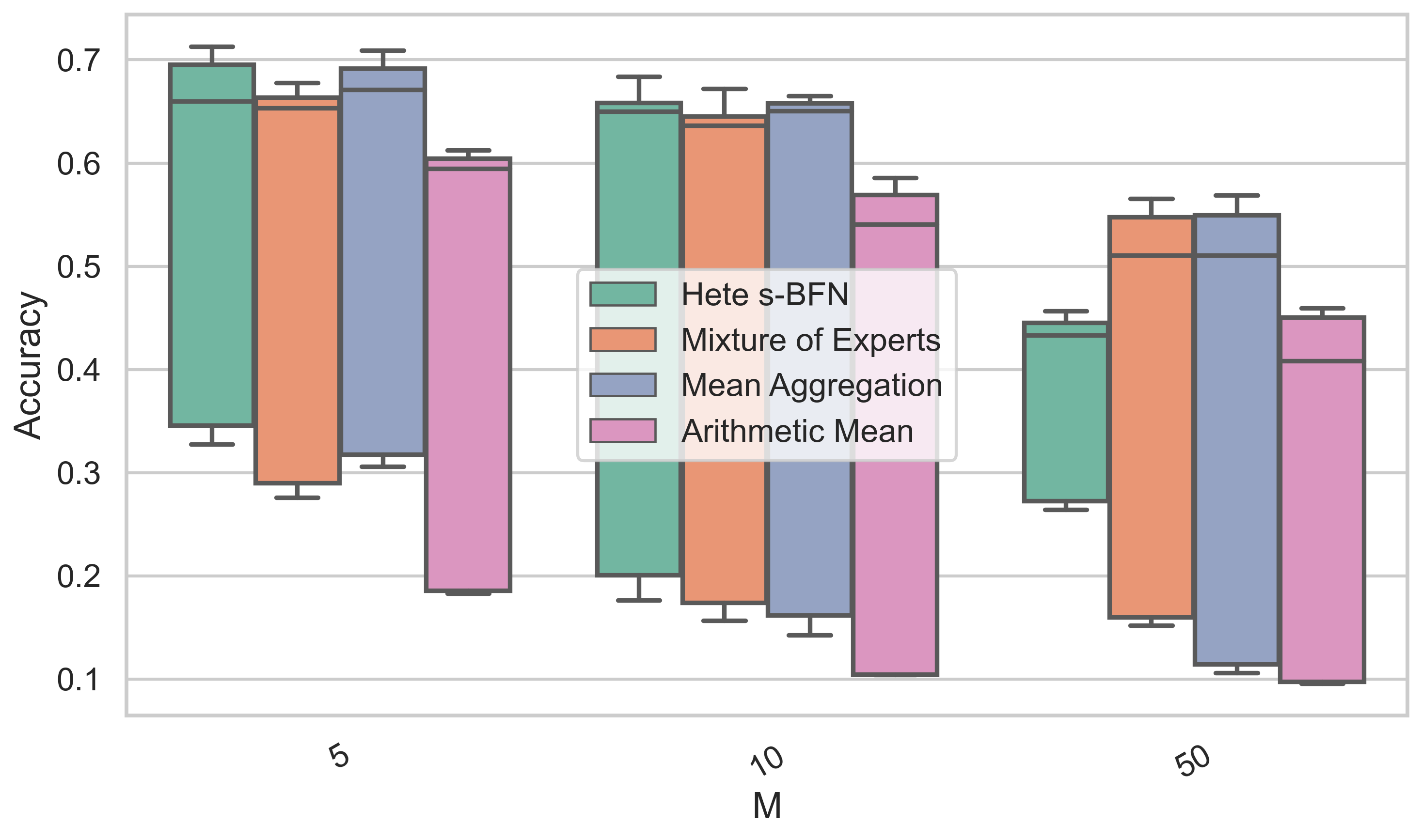}
        \caption{Heterogeneous variants by number of learners (M)}
        \label{M_CIFAR10}
    \end{subfigure}

    \vspace{1em}

    \begin{subfigure}[t]{0.49\textwidth}
        \centering
        \includegraphics[width=\linewidth]{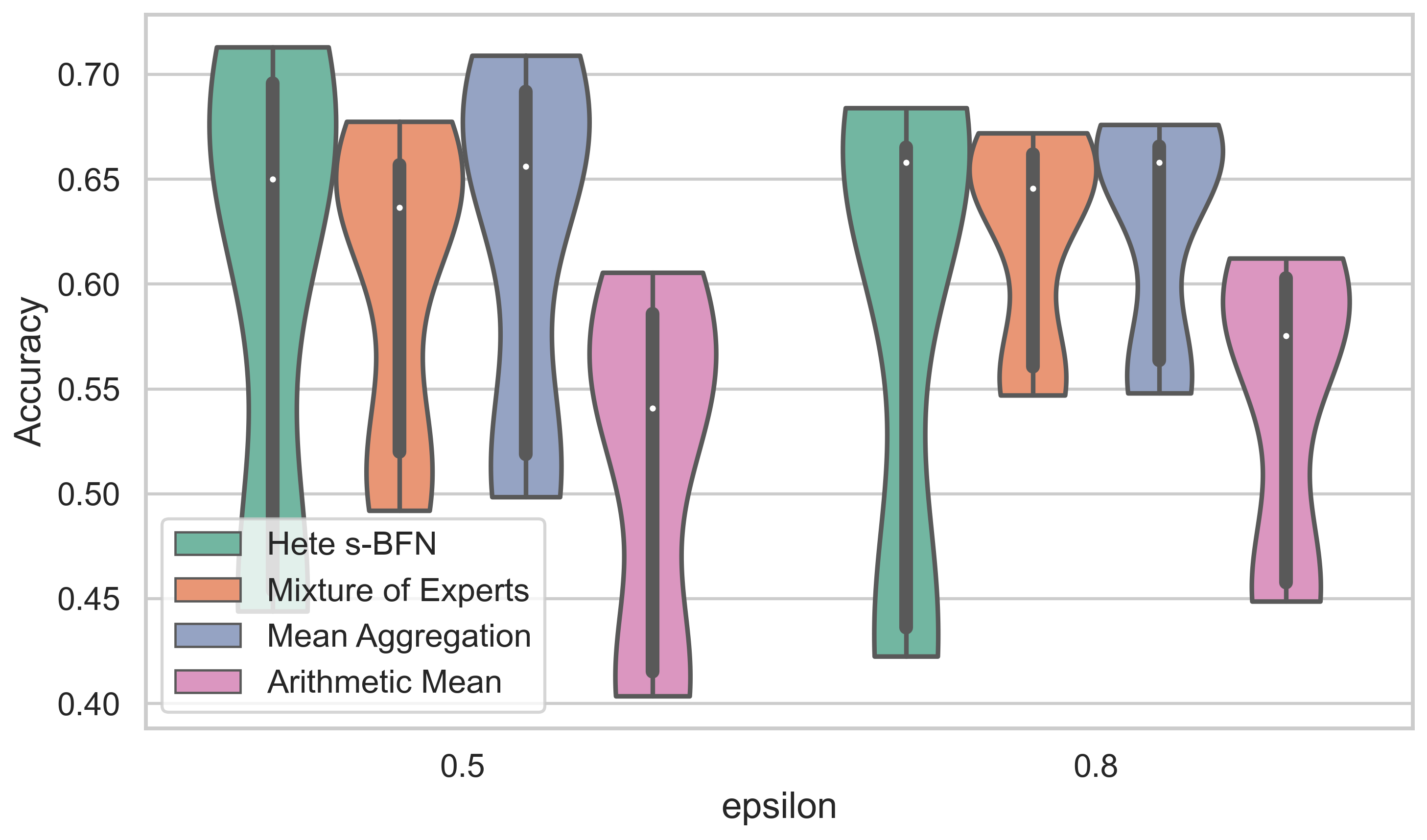}
        \caption{Heterogeneous variants by diversity $(\varepsilon\in[0.3,0.8])$}
        \label{hetecompCIFAR_diver1}
    \end{subfigure}
    \hfill
    \begin{subfigure}[t]{0.49\textwidth}
        \centering
        \includegraphics[width=\linewidth]{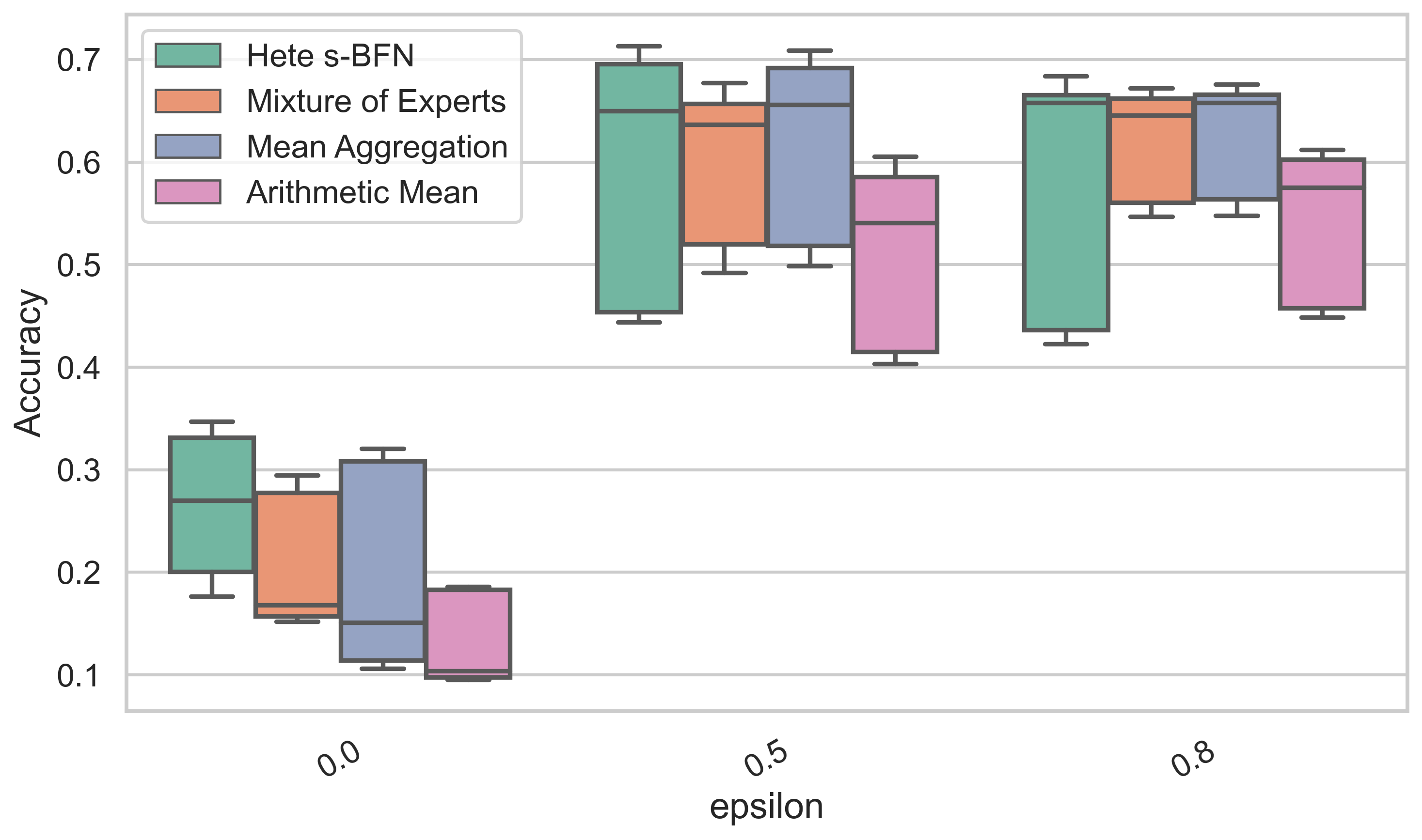}
        \caption{Heterogeneous vs.\ others by diversity $(\varepsilon\in[0,0.8])$}
        \label{hetecompCIFAR_diver2}
    \end{subfigure}

    \caption{Classification accuracy for different ensemble configurations on CIFAR-10.}
    \label{fig:hete_CIFARV2}
\end{figure}

\begin{figure}[t!]
    \centering
    \includegraphics[width=0.7\textwidth]{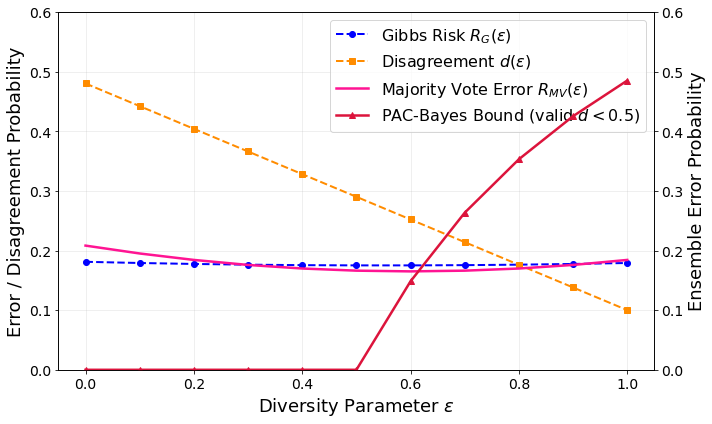}
    \caption{
    Gibbs risk \(R_G(\varepsilon)\) and expected disagreement \(d(\varepsilon)\) (left axis), together with majority-vote error \(R_{MV}(\varepsilon)\) and its PAC-Bayes upper bound (right axis), for varying \textcolor{black}{diversity-control parameter} \(\varepsilon\). The curves illustrate an intermediate \(\varepsilon\) where the bound and empirical error are minimized.
    }
    \label{fig:disagreement_analysis}
\end{figure}

In summary, ensemble performance reflects a multidimensional trade-off among diversity, capacity, and data complexity. The strongest results are typically associated with moderate-to-high diversity (\(\varepsilon\in[0.3,0.8]\)), heterogeneous learners, robust combination rules such as the s-BFN combiner, and conservative learning rates. The observed trends are consistent with both the Bregman bias--variance--diversity framework \citep{wood2023unified} and the PAC-Bayesian disagreement analysis \citep{10.5555/2789272.2831140}, jointly explaining the emergence of an intermediate \(\varepsilon\) that minimizes error while reducing base-level disagreement and Gibbs risk.

The preceding image-classification experiments establish how ensemble composition, the \textcolor{black}{diversity-control parameter}, and combiner choice interact in static 10-class prediction spaces. These results show that the benefits of s\mbox{-}BFN are most pronounced when architectural diversity and sufficient aggregation capacity are present. We now extend the evaluation to temporally structured video classification, where the prediction space is higher-dimensional, the inputs exhibit strong sequential variation, and the aggregation problem becomes correspondingly richer.

\subsection{Video Classification on UCF101}
\label{subsec:ucf101}

\subsubsection{Dataset, Architectures, and Training Setup}

We evaluate the proposed structured Basis Function Network (s\mbox{-}BFN) on the UCF101 action recognition benchmark~\citep{soomro2012ucf101}, which contains 13{,}320 video clips across 101 action categories. The dataset exhibits substantial variability in motion patterns, camera viewpoints, scene context, and temporal structure, making it a suitable benchmark for assessing ensemble aggregation methods on sequential data. Performance is reported using top-1 classification accuracy on standard validation and test splits. In this setting, the UCF101 experiments provide a large-scale evaluation of the proposed framework under temporal and high-dimensional prediction conditions. Consequently, the architectural and aggregation design choices established in the lower-complexity experiments are held fixed, allowing the UCF101 analysis to focus on scalability and robustness rather than additional mechanism isolation.

Each ensemble member is instantiated from a Kinetics-400 pretrained 3D convolutional architecture: R3D-18, MC3-18, or R(2+1)D-18 \citep{hara2018can,tran2018closer,kay2017kinetics}. For all models, the final classification layer is replaced with a 101-way linear head. To balance scalability and adaptation, only the final residual block and classifier are fine-tuned, while earlier layers remain frozen. Ensembles are constructed heterogeneously by cycling architectures across members, ensuring architectural diversity without increasing per-model complexity.

Training proceeds in two stages. First, all experts are warm-started using standard cross-entropy optimization. Second, we apply the proposed $\varepsilon$-structured training rule, which encourages controlled specialization by modulating the update assigned to the lowest-loss expert per sample while distributing non-zero update weight across the remaining experts. This mechanism provides a direct and stable means of regulating diversity during training.

\subsubsection{Aggregation Methods and Experimental Design}

At inference time, expert predictions are combined using several aggregation strategies. In addition to the proposed s-BFN, we consider probability averaging (\emph{Mean}), a linear fusion baseline (\emph{Fusion}), and a Mixture-of-Experts (\emph{MoE}) model with a learned gating network. All aggregation methods operate on expert probability outputs, are trained using identical data splits, and follow the same evaluation protocol. This ensures that observed performance differences are attributable to the aggregation mechanism rather than architectural or training confounders.

For s-BFN, the aggregation capacity is allowed to scale independently of
ensemble size by increasing the number of basis functions proportionally
with the number of experts. This design reflects the hypothesis that
aggregation quality depends not only on the number of hypotheses, but also
on the representational capacity used to combine them.
\textcolor{black}{The aggregation-capacity setting is part of the experimental
design rather than a model-selection variable: $K$ is scaled proportionally
with $M$ and is not selected according to validation or test performance.}

\textcolor{black}{
For $M=20$ and $C=101$, the structured prediction vector has dimension $d_D=MC=2020$. The RBF distances are evaluated in this full prediction space, with no dimensionality reduction applied before the radial mapping; the resulting responses are represented by $K=400$ radial features before the final 101-class linear readout. LayerNorm, when used, acts on these
radial features to improve conditioning rather than reducing the dimensionality of the structured input.
}

\subsubsection{Results and Sensitivity to \textcolor{black}{Diversity-Control Parameter}}

Figure~\ref{fig:ucf101-val-eps} reports validation accuracy as a function of the \textcolor{black}{diversity-control parameter} $\varepsilon$ for ensemble sizes $M\in\{5,10,20\}$. Across most settings, increasing $\varepsilon$ improves performance, but the gains are not strictly monotonic with ensemble size $M$. This behavior is expected because $M$ controls the number of hypotheses, whereas $K$ controls aggregation capacity, and the benefit of additional hypotheses depends on whether the learned combiner can exploit their complementary predictions. In Table~\ref{tab:ucf101-main}, for example, $M=10$ and $\varepsilon=0.6$ does not improve over $M=10$ and $\varepsilon=0.3$, whereas $M=20$ and $\varepsilon=0.6$ gives the strongest test accuracy. This pattern should not be interpreted as a violation of ensemble theory, but as evidence that ensemble size, \textcolor{black}{the diversity-control parameter}, and aggregation capacity interact jointly.

\textcolor{black}{For each ensemble size $M$, the value of $\varepsilon$ used in the subsequent aggregation comparison is selected according to validation accuracy in Table~\ref{tab:ucf101-main}; the test set is not used for this selection.} A similar pattern is observed on the test set, as shown in Figure~\ref{fig:ucf101-test-eps}. The best-performing configuration ($M=20$, $\varepsilon=0.6$) achieves a test accuracy of 86.4\%, outperforming probability averaging, linear fusion, and Mixture-of-Experts baselines under identical training conditions.

Figure~\ref{fig:ucf101-heatmap} provides a global view of validation performance over ensemble size and \textcolor{black}{the diversity-control parameter}, showing structured performance variation rather than a single isolated optimum. The apparent non-monotonicity across $M$ and $\varepsilon$ is therefore an empirical manifestation of the proposed capacity--complexity--diversity trade-off. Classical ensemble theory predicts that additional diverse predictors may improve generalization when their errors are complementary, but it does not require uniform gains for every increase in $M$ or $\varepsilon$. In high-dimensional temporal classification, additional experts can be beneficial only when their predictions add exploitable complementary structure and when the aggregation model has sufficient capacity to combine them.

Figure~\ref{fig:ucf101-temperature} further illustrates how the aggregation mechanism adapts as diversity increases through the learned temperature parameter. A detailed analysis of aggregation capacity, predictive entropy, and temperature behavior is provided in~\ref{app:ucf101}. 

Taken together, these results are consistent with a capacity--complexity--diversity trade-off. While limited aggregation capacity is sufficient for simpler prediction spaces, higher-dimensional and temporally structured settings induce richer prediction geometries in which increased aggregation capacity and controlled diversity can yield tangible improvements in generalization performance.

\begin{table}[t!]
\centering
\caption{UCF101 performance of s-BFN under controlled variation of ensemble size $M$ and \textcolor{black}{diversity-control parameter} $\varepsilon$.}
\label{tab:ucf101-main}
\begin{tabular}{c c c c c}
\toprule
$M$ & $\varepsilon$ & $K$ & Val Acc. & Test Acc. \\
\midrule
5  & 0.1 & 100 & 0.795 & 0.792 \\
5  & 0.3 & 100 & 0.827 & 0.822 \\
5  & 0.6 & 100 & 0.843 & 0.843 \\
10 & 0.1 & 200 & 0.821 & 0.823 \\
10 & 0.3 & 200 & 0.855 & 0.843 \\
10 & 0.6 & 200 & 0.841 & 0.843 \\
20 & 0.1 & 400 & 0.815 & 0.822 \\
20 & 0.3 & 400 & 0.844 & 0.843 \\
20 & 0.6 & 400 & \textbf{0.858} & \textbf{0.864} \\
\bottomrule
\end{tabular}
\end{table}

\begin{table}[t]
\centering
\caption{Test accuracy comparison of aggregation strategies on UCF101, using the \textcolor{black}{validation-selected} $\varepsilon$ per ensemble size $M$ for s-BFN.}
\label{tab:ucf101-combiners}
\begin{tabular}{c c c c c}
\toprule
$M$ & Mean & MoE & Fusion & s-BFN \\
\midrule
5  & 0.803 & 0.774 & 0.364 & \textbf{0.843} \\
10 & 0.786 & 0.732 & 0.277 & \textbf{0.843} \\
20 & 0.766 & 0.692 & 0.326 & \textbf{0.864} \\
\bottomrule
\end{tabular}
\end{table}

\begin{figure}[ht]
\centering
\includegraphics[width=0.55\linewidth]{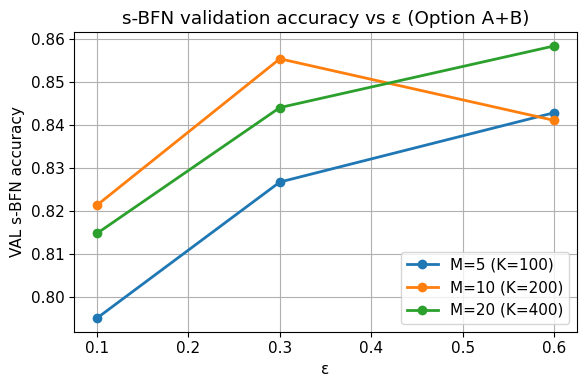}
\caption{Validation accuracy of s-BFN on UCF101 as a function of the \textcolor{black}{diversity-control parameter} $\varepsilon$ for different ensemble sizes $M$.}
\label{fig:ucf101-val-eps}
\end{figure}

\begin{figure}[t]
\centering
\includegraphics[width=0.55\linewidth]{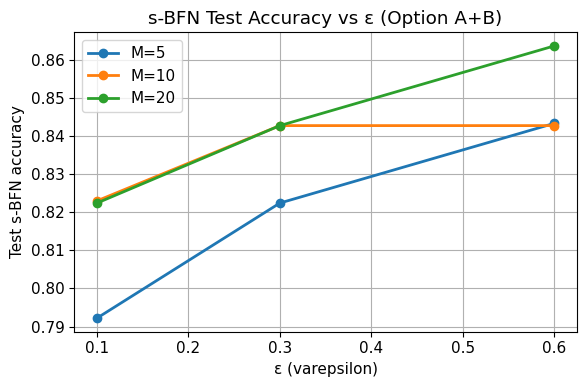}
\caption{Test accuracy of s-BFN on UCF101 as a function of the \textcolor{black}{diversity-control parameter} $\varepsilon$ for different ensemble sizes $M$.}
\label{fig:ucf101-test-eps}
\end{figure}

\begin{figure}[t]
\centering
\includegraphics[width=0.55\linewidth]{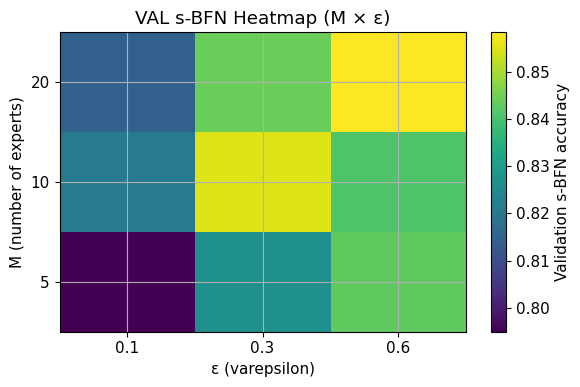}
\caption{Heatmap of validation accuracy over ensemble size $M$ and the \textcolor{black}{diversity-control parameter} $\varepsilon$.}
\label{fig:ucf101-heatmap}
\end{figure}

\begin{figure}[t]
\centering
\includegraphics[width=0.55\linewidth]{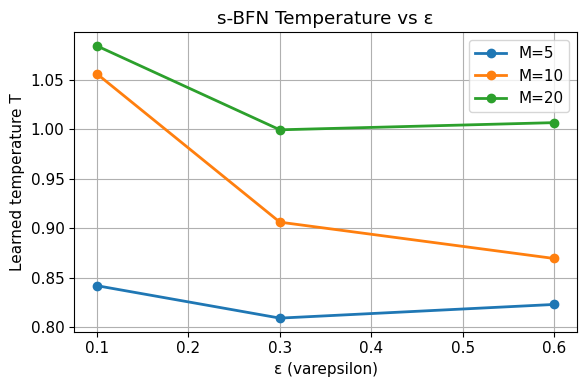}
\caption{Learned aggregation temperature as a function of the \textcolor{black}{diversity-control parameter} $\varepsilon$ for different ensemble sizes $M$.}
\label{fig:ucf101-temperature}
\end{figure}

\subsection{Computational Efficiency and Stability Analysis}

This section returns to the image-classification experiments and evaluates the computational efficiency of structured basis function network (s\mbox{-}BFN) ensembles, focusing on training cost, inference cost, stability, and performance gain over individual base learners. Results on CIFAR-10 and Fashion-MNIST (Figures~\ref{fig:ComputationalEfficiency_CIFAR1} and \ref{fig:ComputationalEfficiency_Fashion_MNIST}) reveal consistent patterns across homogeneous and heterogeneous ensembles, as well as different combiners, including s-BFN, logit averaging, arithmetic mean, and MoE. All measurements were obtained on a Lenovo ThinkPad P17 Gen~2 mobile workstation equipped with an Intel Core i7-11800H CPU, 16\,GB RAM, a 512\,GB NVMe SSD, and an NVIDIA RTX A2000 GPU; operating system: Windows~10 Pro. Experiments used a CUDA-enabled deep-learning framework.

\textcolor{black}{
Conditional on the same already-computed expert predictions, the aggregation
cost can be decomposed into three components. First, constructing the
structured input concatenates the $M$ expert probability vectors into
$\boldsymbol{D}_i\in\mathbb{R}^{MC}$, requiring $O(MC)$ operations. Second,
s-BFN evaluates $K$ radial basis functions, each requiring a distance in the
$MC$-dimensional structured space, giving $O(KMC)$ cost. Third, the resulting
$K$ features are mapped to $C$ output logits by the linear readout, requiring
$O(KC)$ operations. Simple probability averaging requires only the first-order
$O(MC)$ aggregation over expert probabilities. Thus, conditional on common
expert inference, the incremental s-BFN overhead relative to averaging is
dominated by the $O(KMC)$ RBF evaluation, followed by the $O(KC)$ readout.
Table~\ref{tab:efficiency_all} complements this complexity analysis with empirical efficiency
metrics for the evaluated Fashion-MNIST and CIFAR-10 configurations.
}

Figures~\ref{fig:eff_cifar10_1} and \ref{fig:eff_fashionmnist_1_4} show that heterogeneous ensembles achieve higher accuracy with comparable or lower training time than homogeneous CNN ensembles. This effect is most evident in CIFAR-10, where CNN-only ensembles are inefficient relative to mixed-architecture ones. In Fashion-MNIST, training time grows roughly linearly with ensemble size, and the additional overhead of heterogeneous architectures remains moderate (Figure~\ref{fig:eff_fashionmnist_2_3}).

Inference costs increase slightly with accuracy, as seen in CIFAR-10 (Figure~\ref{fig:eff_cifar10_2}), but the s-BFN combiner maintains a favorable trade-off. In Fashion-MNIST, inference time scales predictably with ensemble size (Figure~\ref{fig:eff_fashionmnist_2_1}), with heterogeneous ensembles incurring only marginal extra cost.

Stability across training runs, reported in Figures~\ref{fig:eff_cifar10_2_1} and \ref{fig:eff_fashionmnist_2_1}, is generally higher for s-BFN than for logit averaging, MoE, or arithmetic mean. Variance is particularly reduced when using heterogeneous learners, underscoring robustness to random initialization and stochastic optimization. Accuracy gains over the arithmetic base-learner average accuracies are shown in Figure~\ref{fig:eff_cifar10_4}. The s-BFN combiner delivers the largest improvements in most reported settings, especially in heterogeneous ensembles where diversity is highest. Similar trends are observed in Fashion-MNIST, where performance improvements come at modest additional training or inference cost.

Overall, the results demonstrate that the s-BFN combiner achieves an effective balance between accuracy and efficiency. Heterogeneous ensembles with s-BFN not only yield higher accuracy but also remain computationally practical, with stable generalization and scalability across datasets. The modest cost of added diversity supports the practical feasibility of s-BFN for performance-sensitive classification tasks.

\begin{figure}[t!]
    \centering

    \begin{subfigure}[t]{0.49\textwidth}
        \includegraphics[width=\linewidth]{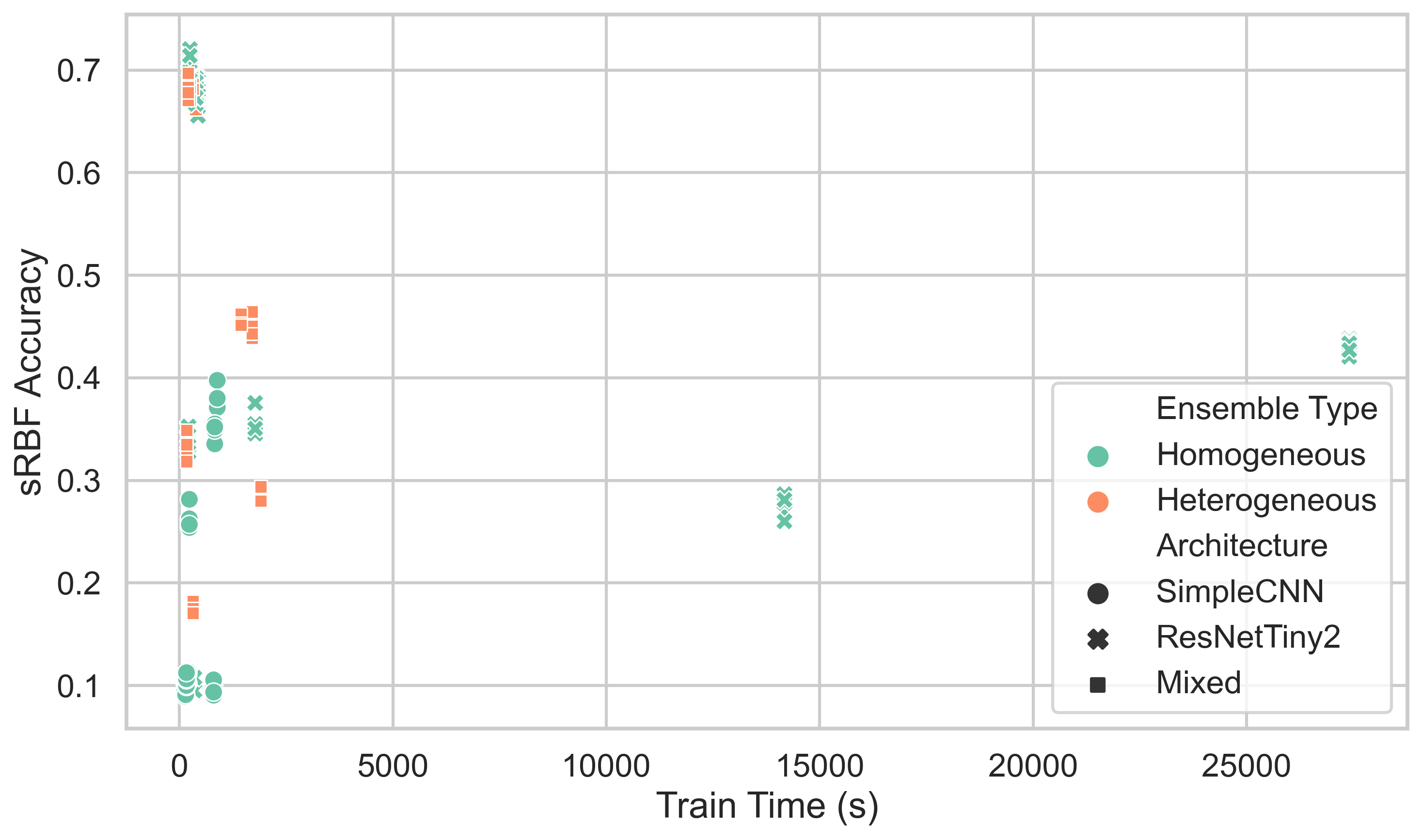}
        \caption{Training time vs.\ accuracy (CIFAR-10)}
        \label{fig:eff_cifar10_1}
    \end{subfigure}
    \hfill
    \begin{subfigure}[t]{0.49\textwidth}
        \includegraphics[width=\linewidth]{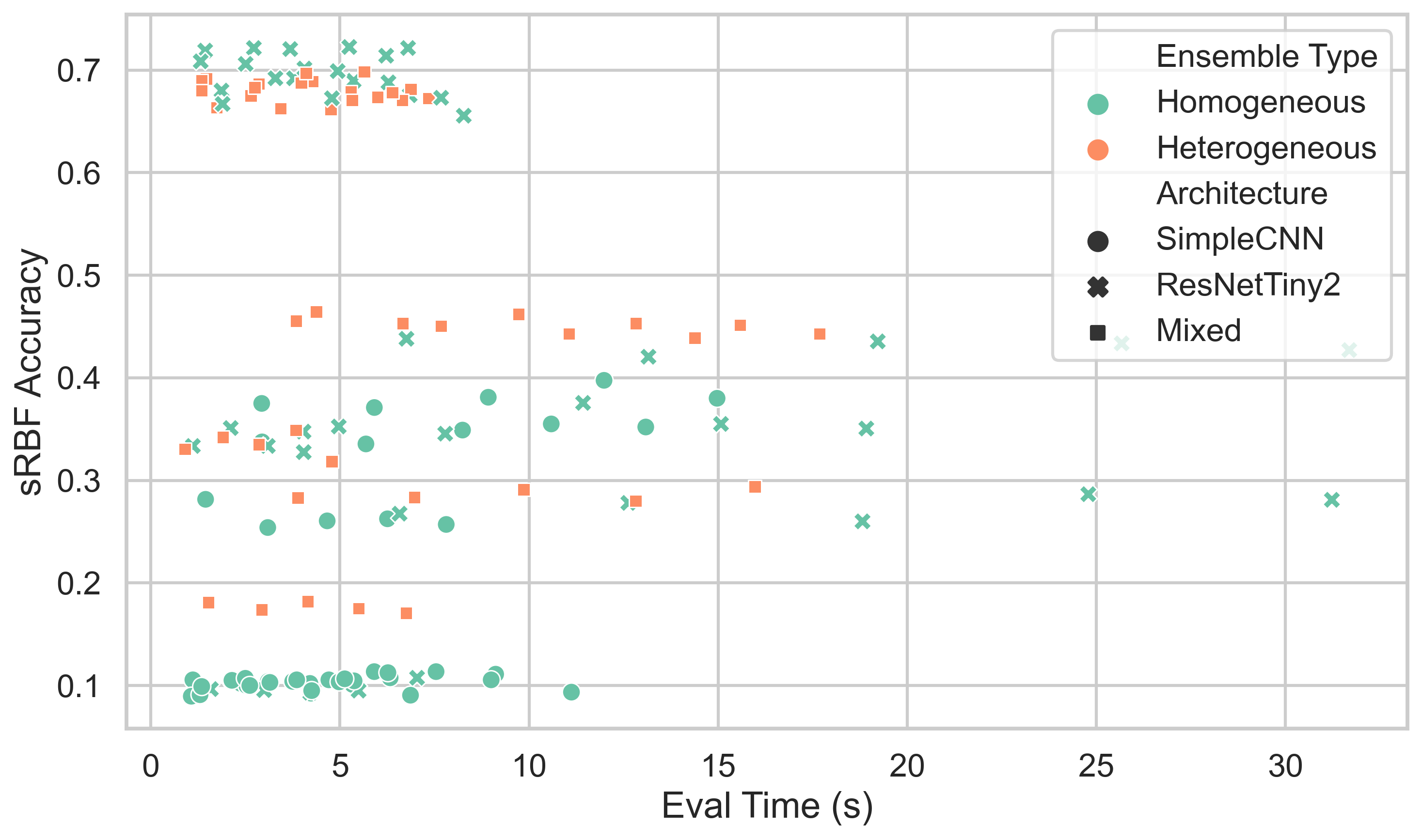}
        \caption{Evaluation time vs.\ accuracy (CIFAR-10)}
        \label{fig:eff_cifar10_2}
    \end{subfigure}

    \vspace{0.5em}

    \begin{subfigure}[t]{0.49\textwidth}
        \includegraphics[width=\linewidth]{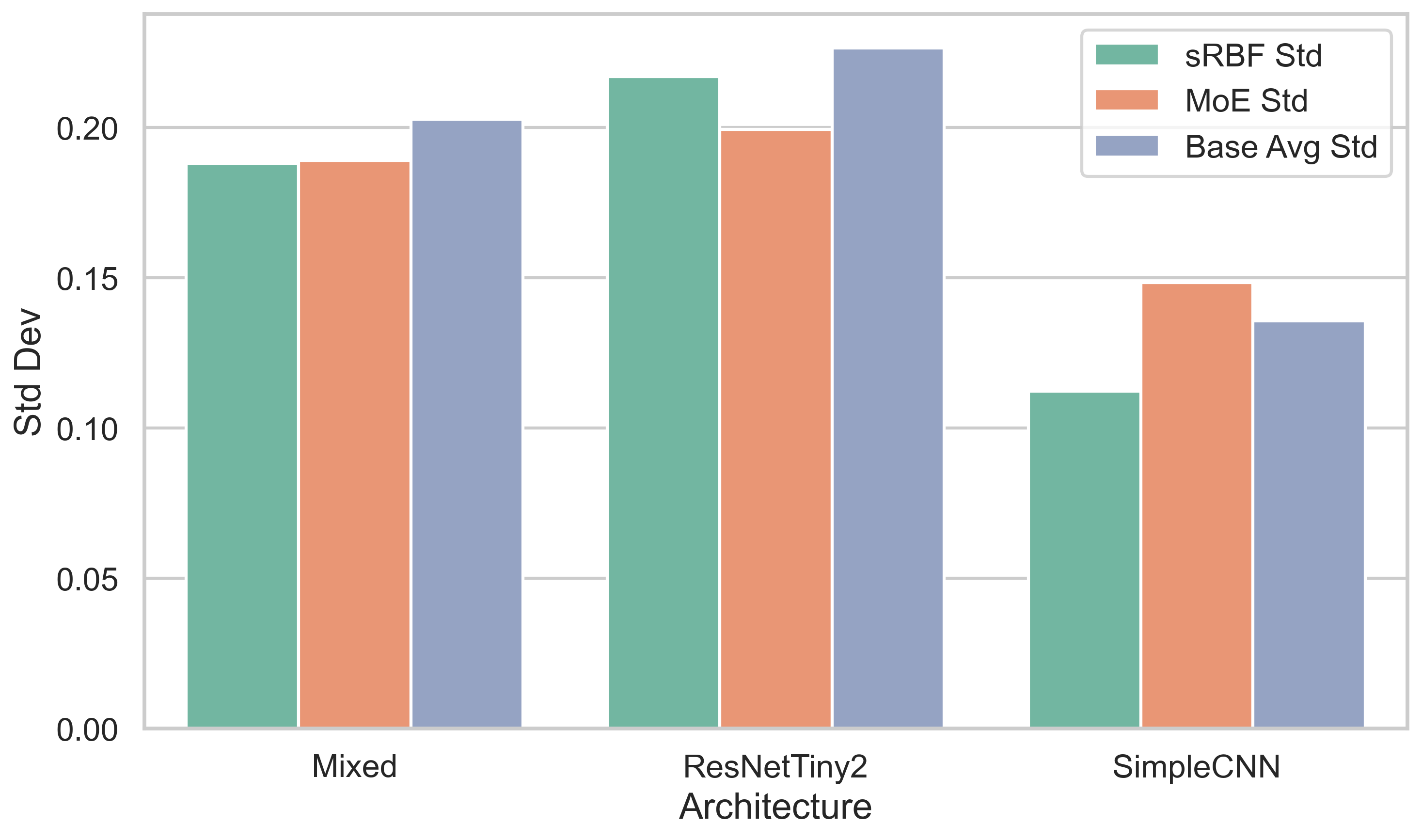}
        \caption{Stability of accuracy across runs (std.\ dev.)}
        \label{fig:eff_cifar10_2_1}
    \end{subfigure}
    \hfill
    \begin{subfigure}[t]{0.49\textwidth}
        \includegraphics[width=\linewidth]{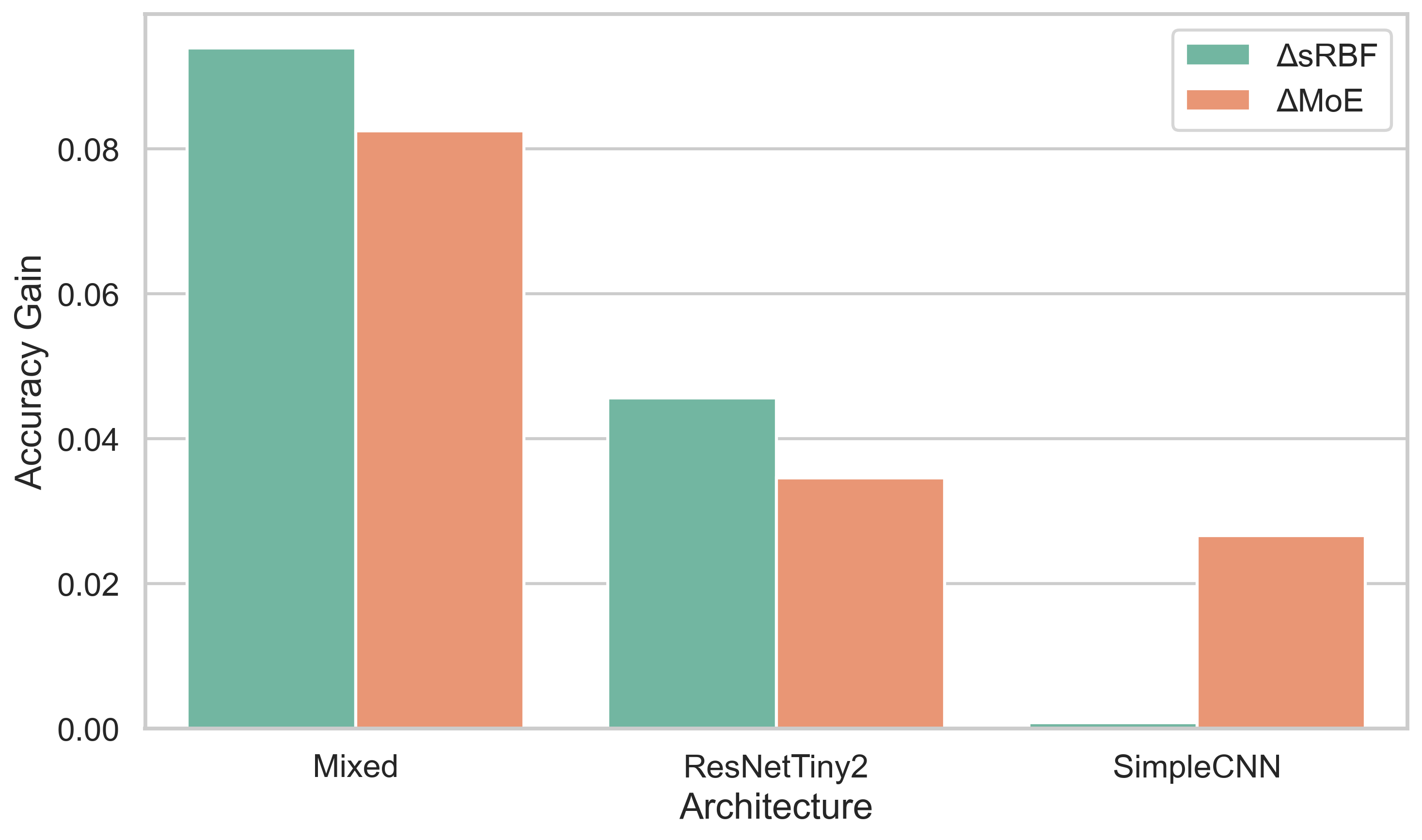}
        \caption{Accuracy gain over arithmetic average base learners (CIFAR-10)}
        \label{fig:eff_cifar10_4}
    \end{subfigure}

    \caption{Computational efficiency evaluation for different ensemble configurations on CIFAR-10. sRBF corresponds to the s-BFN combiner and Base Avg refers to the arithmetic mean of the individual base learners’ accuracies.}
    \label{fig:ComputationalEfficiency_CIFAR1}
\end{figure}

\begin{figure}[t!]
    \centering

    \begin{subfigure}[t]{0.49\textwidth}
        \centering
        \includegraphics[width=\linewidth]{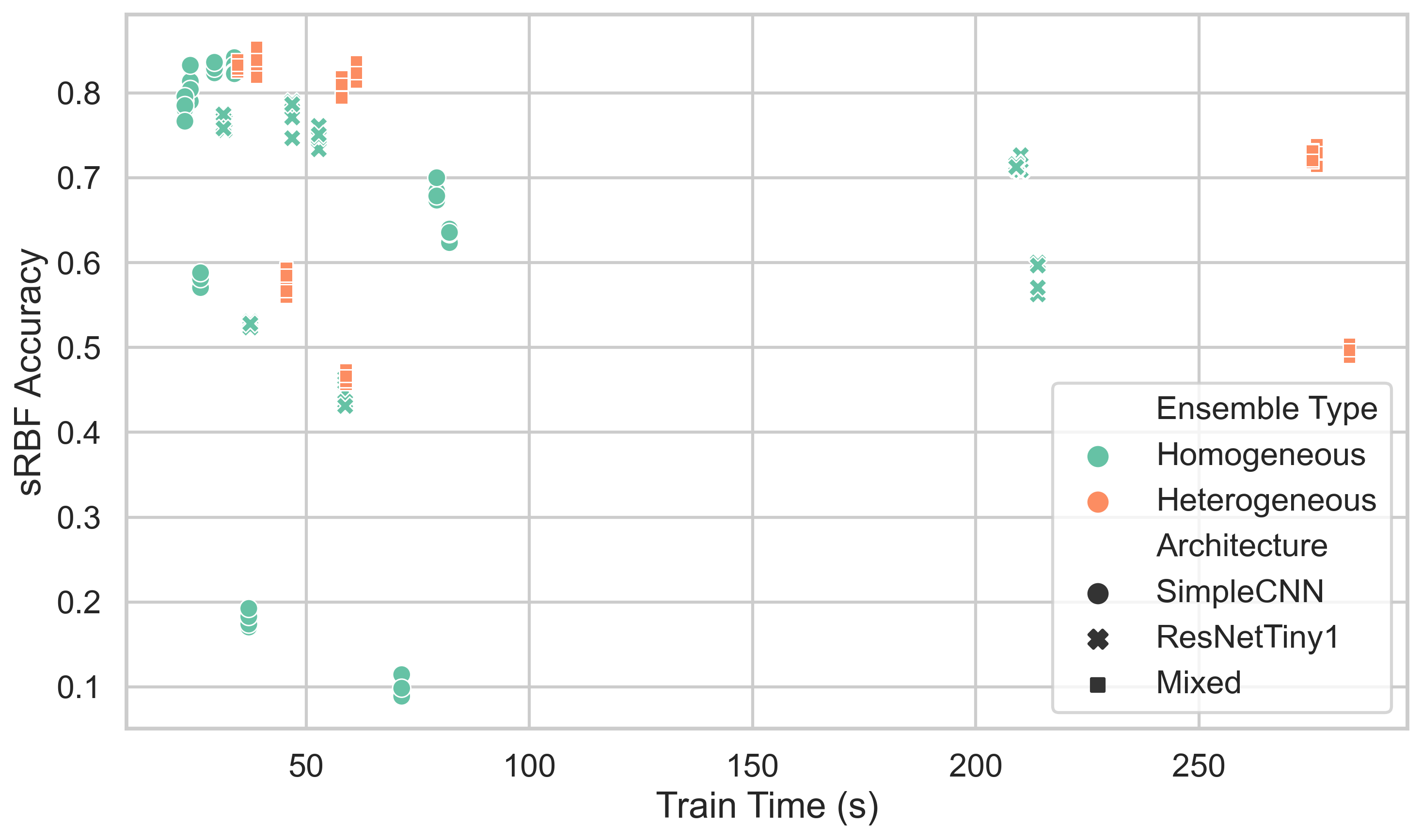}
        \caption{Structured dataset size vs.\ ensemble size}
        \label{fig:eff_fashionmnist_1_4}
    \end{subfigure}
    \hfill
    \begin{subfigure}[t]{0.49\textwidth}
        \centering
        \includegraphics[width=\linewidth]{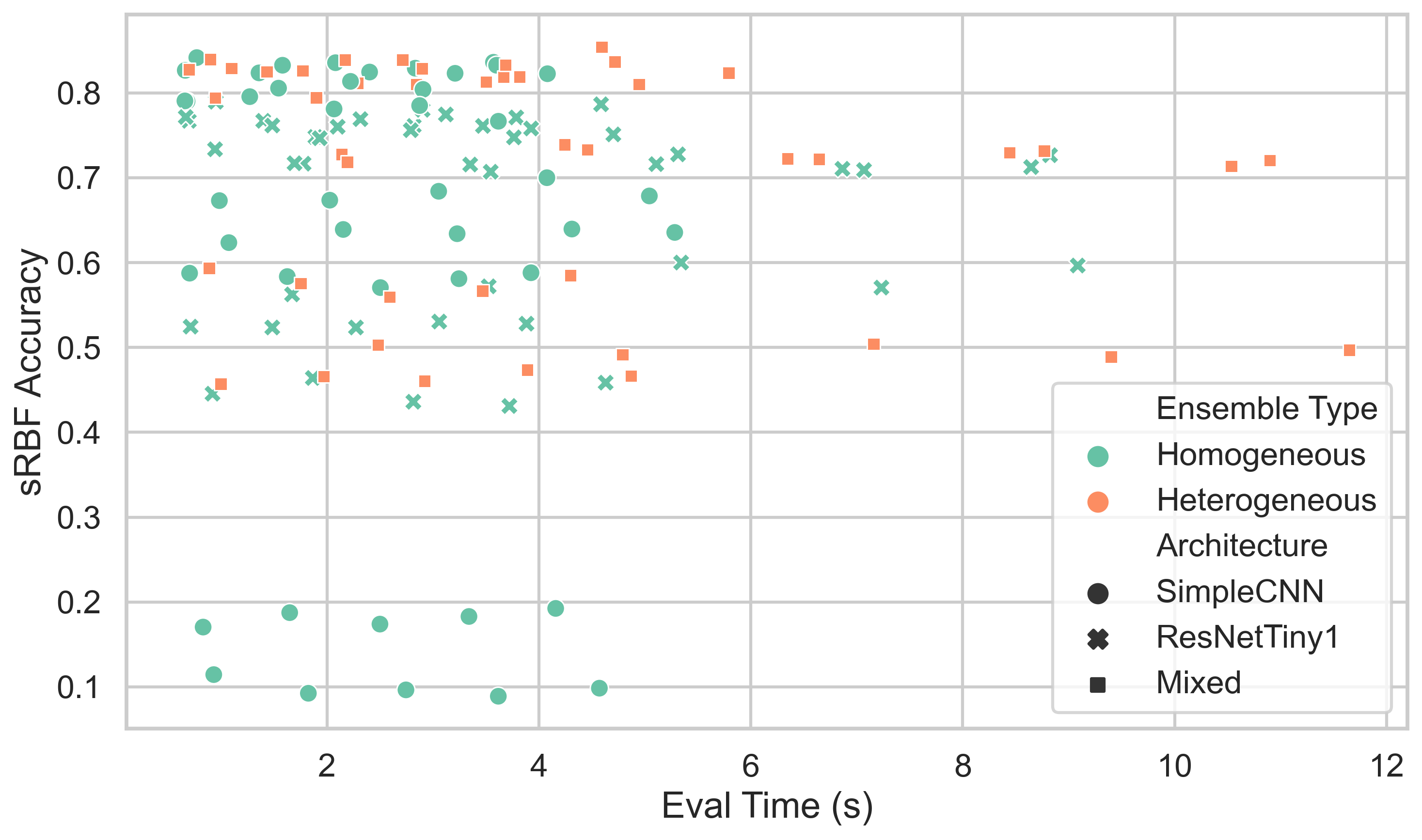}
        \caption{Combiner training time vs.\ \(\varepsilon\)}
        \label{fig:eff_fashionmnist_1_5}
    \end{subfigure}

    \vspace{1em}

    \begin{subfigure}[t]{0.49\textwidth}
        \centering
        \includegraphics[width=\linewidth]{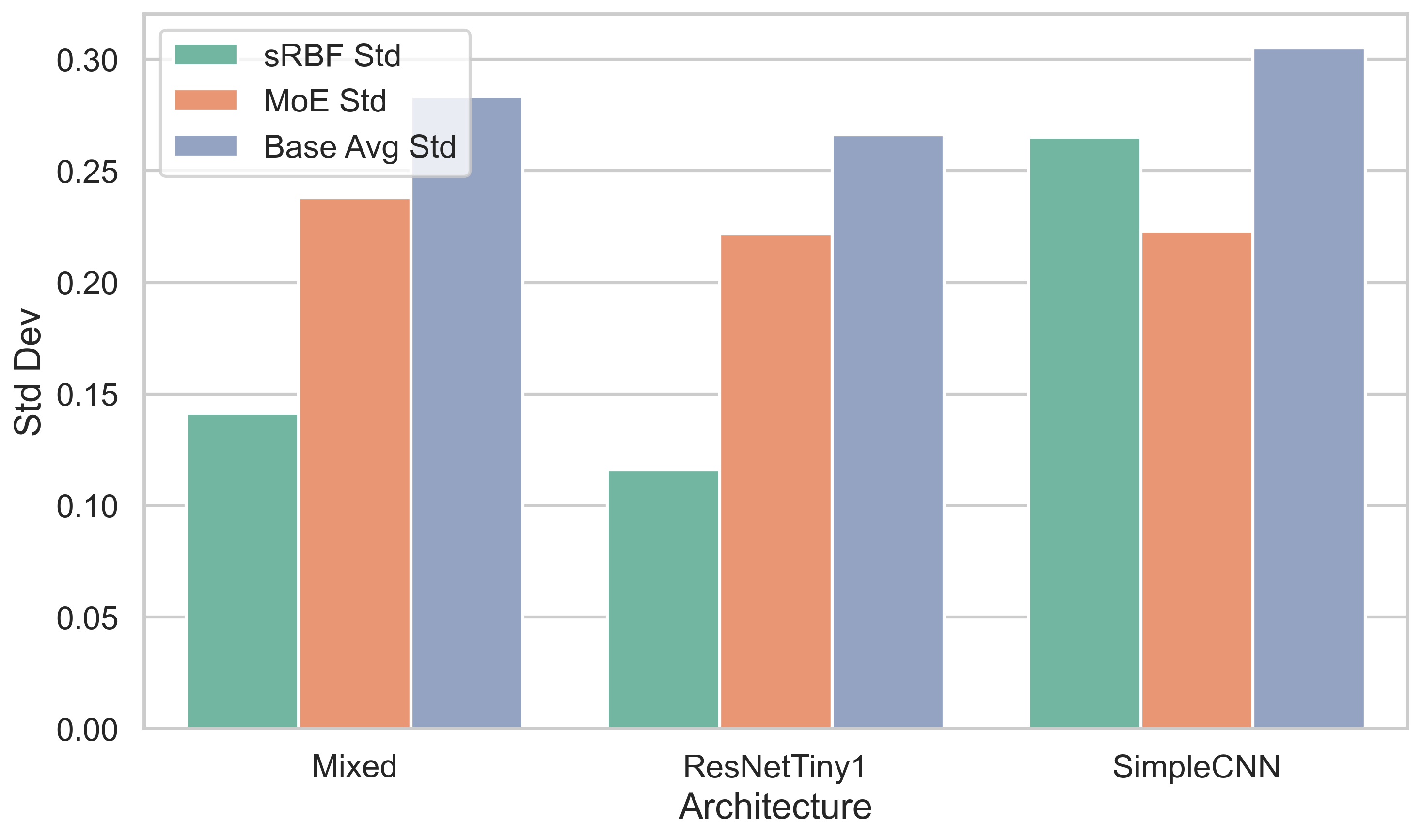}
        \caption{Inference time vs.\ ensemble size}
        \label{fig:eff_fashionmnist_2_1}
    \end{subfigure}
    \hfill
    \begin{subfigure}[t]{0.49\textwidth}
        \centering
        \includegraphics[width=\linewidth]{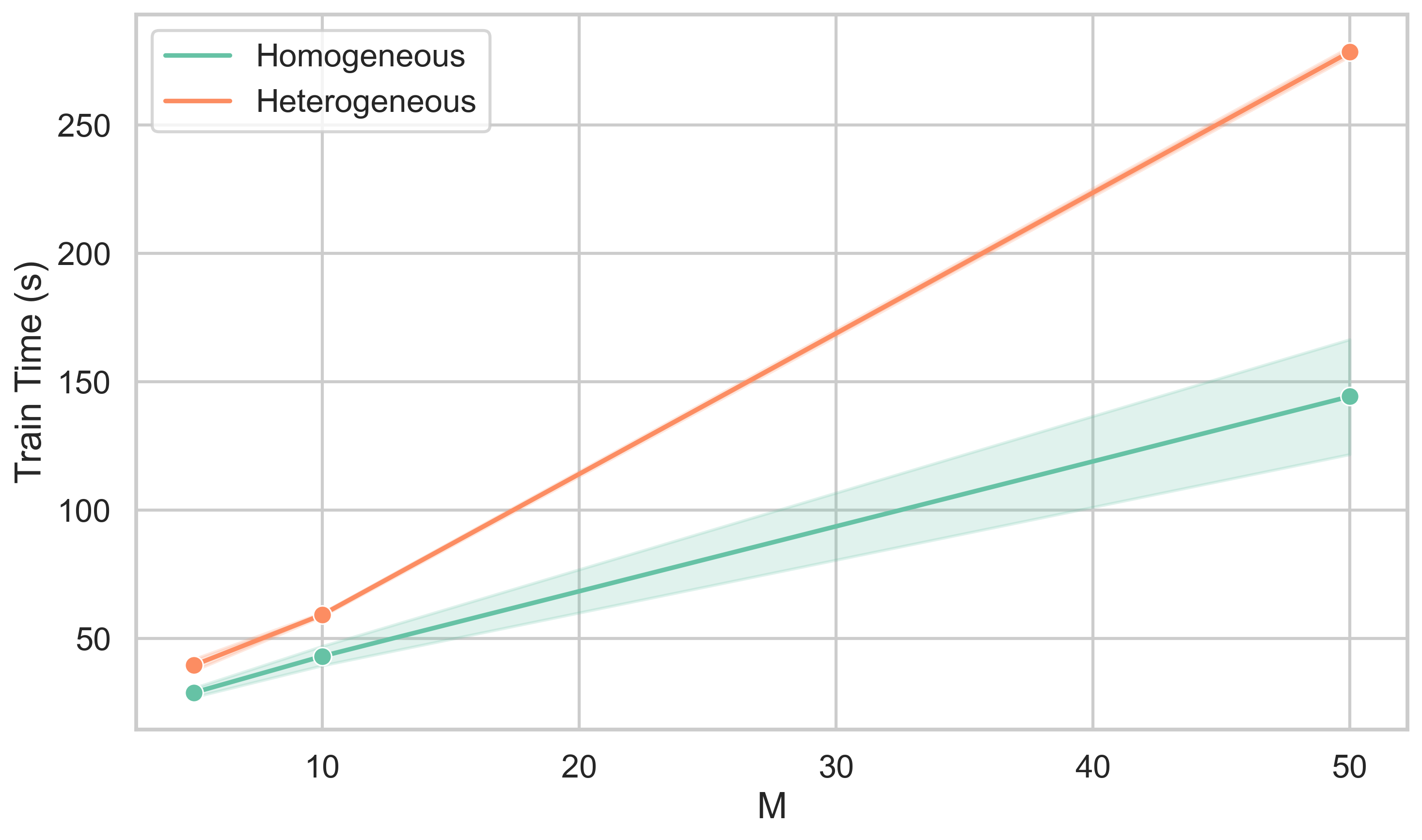}
        \caption{Structured dataset construction time}
        \label{fig:eff_fashionmnist_2_3}
    \end{subfigure}

    \caption{Computational efficiency results on Fashion-MNIST. sRBF corresponds to the s-BFN combiner and Base Avg refers to the arithmetic mean of the individual base learners’ accuracies.}
    \label{fig:ComputationalEfficiency_Fashion_MNIST}
\end{figure}

Across MNIST, Fashion-MNIST, and CIFAR-10, s-BFN ensembles generally improve over the average base learner, with the strongest gains typically observed in heterogeneous configurations. MoE remains competitive in some homogeneous settings, indicating that the relative advantage of s-BFN depends on the diversity and complementarity of the base learners. In Fashion-MNIST, s-BFN provides strong improvements that remain stable across a broad range of diversity values, with peak accuracy around \(\varepsilon \approx 0.5\). On CIFAR-10, s-BFN also delivers clear improvements, particularly with moderate ensemble sizes \(M\) and mid-range \(\varepsilon\). Notably, the most favorable accuracy--latency trade-offs are typically achieved by heterogeneous ensembles.

Tables~\ref{tab:overall_all}--\ref{tab:sensitivity_eps_all} summarize these results. Table~\ref{tab:overall_all} reports average accuracies and gains over the base arithmetic average, showing that s-BFN provides overall improvements across the evaluated datasets, particularly in heterogeneous ensembles. Table~\ref{tab:best_configs} lists the best-performing configurations, highlighting that moderate-to-high \(\varepsilon\) values frequently appear among the strongest settings. Timing values are reported where available and indicate reasonable training and evaluation costs relative to performance. Table~\ref{tab:efficiency_all} compares computational metrics for Fashion-MNIST and CIFAR-10, showing that heterogeneous ensembles often provide favorable accuracy--efficiency trade-offs. Finally, Table~\ref{tab:sensitivity_eps_all} shows that the strongest s-BFN accuracies generally occur in the mid-to-high diversity range, while also confirming that the relationship between \(\varepsilon\) and accuracy is not strictly monotonic.

\begin{table}[t!]
\centering
\caption{Overall accuracy (mean across settings) per dataset and ensemble. Values are reported in \%, with gains in percentage points (pp) relative to the average base-learner accuracy.}
\label{tab:overall_all}
\begin{tabular}{llrrrrr}
\toprule
Dataset & Ensemble & s-BFN\_Acc & MoE\_Acc & Base\_Acc & $\Delta$s-BFN & $\Delta$MoE \\
\midrule
MNIST        & Heterogeneous & 48.47 & 47.56 & 39.65 &  8.82 &  7.91 \\
MNIST        & Homogeneous   & 32.97 & 34.70 & 30.40 &  2.56 &  4.29 \\
FashionMNIST & Heterogeneous & 69.81 & 63.10 & 54.19 & 15.62 &  8.90 \\
FashionMNIST & Homogeneous   & 63.70 & 61.57 & 53.02 & 10.69 &  8.56 \\
CIFAR-10     & Heterogeneous & 49.10 & 47.96 & 39.71 &  9.39 &  8.25 \\
CIFAR-10     & Homogeneous   & 32.70 & 33.45 & 30.38 &  2.32 &  3.06 \\
\bottomrule
\end{tabular}
\end{table}

\begin{table}[t!]
\centering
\caption{Best s-BFN configuration per dataset and ensemble. All runs used temperature \(T=3\) and learning rate \(\eta_{\alpha}=0.1\); \(\varepsilon\) values follow the original configurations and are omitted for brevity. Accuracies in \%.}
\label{tab:best_configs}
\resizebox{\textwidth}{!}{%
\begin{tabular}{lllcrrrrr}
\toprule
Dataset & Ensemble & Architecture & \(M\) & s-BFN (\%) & MoE (\%) & Base Avg (\%) & Train (s) & Eval (s) \\
\midrule
MNIST        & Het.  & Mixed       & 5  & 70.45 & 66.48 & 59.34 & --     & --   \\
MNIST        & Homo. & ResNetTiny2 & 5  & 71.66 & 70.03 & 68.20 & --     & --   \\
Fashion-MNIST     & Het.  & Mixed       & 5  & 83.54 & 82.38 & 75.99 & 38.90  & 2.73 \\
Fashion-MNIST     & Homo. & CNN         & 10 & 83.01 & 81.02 & 80.25 & 29.50  & 2.10 \\
CIFAR-10     & Het.  & Mixed       & 10 & 68.94 & 66.99 & 60.28 & 313.10 & 4.22 \\
CIFAR-10     & Homo. & ResNetTiny2 & 5  & 71.72 & 66.27 & 64.33 & 258.30 & 4.05 \\
\bottomrule
\end{tabular}%
}
\end{table}

\begin{table}[t!]
\centering
\caption{Efficiency metrics for FashionMNIST and CIFAR-10: mean train/eval time, average FLOPs per base model, and total base-model parameters.}
\label{tab:efficiency_all}
\begin{tabular}{llrrrr}
\toprule
Dataset & Ensemble & Train (s) & Eval (s) & FLOPs & Params \\
\midrule
FashionMNIST & Heterogeneous &  125.8 & 4.09 &  5.8M & 1.24M \\
FashionMNIST & Homogeneous   &   72.2 & 2.98 &  1.7M & 0.80M \\
CIFAR-10     & Heterogeneous &  741.0 & 6.06 &  7.8M & 1.25M \\
CIFAR-10     & Homogeneous   & 2712.4 & 6.84 & 14.4M & 7.32M \\
\bottomrule
\end{tabular}
\end{table}

\begin{table}[t!]
\centering
\caption{Sensitivity of s-BFN accuracy (\%) to \(\varepsilon\) across datasets for homogeneous and heterogeneous s-BFN configurations.}
\label{tab:sensitivity_eps_all}
\begin{tabular}{rlrr}
\toprule
$\varepsilon$ & Dataset & Heterogeneous & Homogeneous \\
\midrule
0.0 & CIFAR-10     & 26.59 & 16.86 \\
0.5 & CIFAR-10     & 59.84 & 42.89 \\
0.8 & CIFAR-10     & 60.87 & 38.36 \\
0.0 & FashionMNIST & 51.25 & 40.25 \\
0.5 & FashionMNIST & 79.48 & 75.00 \\
0.8 & FashionMNIST & 78.71 & 75.86 \\
0.0 & MNIST        & 26.63 & 18.04 \\
0.5 & MNIST        & 60.11 & 41.55 \\
0.8 & MNIST        & 58.67 & 39.32 \\
\bottomrule
\end{tabular}
\end{table}

\subsection{Ablation Study}

This section analyzes the contribution of specific architectural and training choices to overall performance. Both regression and classification tasks are considered, with comparisons between structured basis function networks (s\mbox{-}BFN) trained via least squares or gradient descent and alternative ensemble strategies. Evaluation metrics include root mean square error (RMSE) for regression, classification accuracy for image recognition, and dispersion across hyperparameter configurations and folds.

Table~\ref{tab:rmse_stats} reports RMSE statistics for absolute humidity and energy appliance prediction across 80 configurations and 10-fold validation. For each configuration, RMSE values are averaged across the 10 folds, and the table reports the best configuration-level mean RMSE together with dispersion statistics across configurations. For absolute humidity, the least-squares s\mbox{-}BFN achieves the lowest mean RMSE over the evaluated configurations (22.46), although its quartile statistics indicate higher variability across configurations. Gradient-based variants ($G_1$--$G_3$) perform competitively, with $G_3$ exhibiting the lowest variance. In the energy appliance task, both $G_3$ and least-squares s\mbox{-}BFN attain the same best configuration-level mean RMSE (101.12), while $G_3$ demonstrates superior consistency, reflected in its narrower interquartile range and lowest standard deviation (0.69). The comparison between gradient-based strategies $G_1$--$G_3$ is specific to the multivariate regression setting considered here. In this low-dimensional continuous prediction space, $G_1$ and $G_2$ exhibit higher variability, particularly at smaller ensemble sizes, whereas $G_3$ provides a more stable trade-off between locality and smoothness. These observations are not intended to generalize to image and video experiments, which rely on different basis-function constructions (see Appendix~A).

For image recognition, ablation is conducted along three axes: ensemble size (\(M\)), \textcolor{black}{diversity-control parameter} (\(\varepsilon\)), and combiner strategy. Figures~\ref{M_MNIST}, \ref{M_FashionMNIST}, and \ref{M_CIFAR10} show that homogeneous ensembles quickly saturate with increasing \(M\), particularly in CIFAR-10, whereas heterogeneous ensembles continue to benefit, indicating a higher representational ceiling. Diversity effects are evident in Figures~\ref{diversityMNIST}, \ref{diversityfashionMNIST}, and \ref{hetecompCIFAR_diver2}, where accuracy generally improves with larger \(\varepsilon\), especially on CIFAR-10 where disagreement acts as a regularizer. Violin plots in Figures~\ref{hetecompfashMNIST_diver1} and \ref{hetecompCIFAR_diver1} further illustrate that higher \(\varepsilon\) values often yield tighter and higher-performing accuracy distributions.

Computational trade-offs are highlighted in Figures~\ref{fig:ComputationalEfficiency_CIFAR1} and \ref{fig:ComputationalEfficiency_Fashion_MNIST}. On CIFAR-10, heterogeneous ensembles achieve higher accuracy at lower training cost than homogeneous CNN ensembles (Figure~\ref{fig:eff_cifar10_1}). Evaluation time increases with accuracy (Figure~\ref{fig:eff_cifar10_2}), yet s\mbox{-}BFN maintains a favorable trade-off. Stability, measured by variance across runs (Figure~\ref{fig:eff_cifar10_2_1}), is generally highest in s\mbox{-}BFN ensembles. Accuracy gains relative to base learners are also largest for s\mbox{-}BFN in most reported settings (Figure~\ref{fig:eff_cifar10_4}). Fashion-MNIST exhibits similar trends: structured dataset size scales linearly with \(M\) (Figure~\ref{fig:eff_fashionmnist_1_4}); combiner training time grows moderately with \(\varepsilon\) (Figure~\ref{fig:eff_fashionmnist_1_5}); inference costs remain manageable (Figure~\ref{fig:eff_fashionmnist_2_1}); and dataset construction overhead stays within practical limits (Figure~\ref{fig:eff_fashionmnist_2_3}).

\begin{table}[t!]
\centering
\caption{Statistics of 10-fold validation RMSE across all hyperparameter combinations. For each configuration, RMSE values are averaged across the 10 folds; the table reports the best configuration-level mean RMSE and dispersion statistics across configurations (Q: quartiles).}
\label{tab:rmse_stats}

\vspace{0.8em}
\textit{(a) Absolute humidity prediction.} RMSE statistics across 80 hyperparameter configurations for different s-BFN variants. The best configuration-level mean RMSE, standard deviation, and first and third quartiles across configurations are reported. s-BFN LS is the closed-form variant.

\vspace{0.3em}
\begin{tabular}{lcccc}
\toprule
\textbf{Model} & \textbf{Best mean RMSE} & \textbf{Std. dev.} & \textbf{1st Q} & \textbf{3rd Q} \\
\midrule
\textbf{s-BFN LS}        & \textbf{22.46} & 9.14          & 38.98          & 54.71 \\
$\boldsymbol{G_1}$       & 37.32          & 2.92          & \textbf{37.94} & \textbf{39.66} \\
$\boldsymbol{G_2}$       & 37.28          & 3.03          & 38.72          & 40.87 \\
$\boldsymbol{G_3}$       & 37.35          & \textbf{2.51} & 39.02          & 39.67 \\
\bottomrule
\end{tabular}

\vspace{1.5em}
\textit{(b) Energy appliance prediction.} RMSE statistics across 80 hyperparameter configurations for different s-BFN variants. Both s-BFN LS and $G_3$ reach the lowest best configuration-level mean RMSE; quartile values indicate consistency across configurations.

\vspace{0.3em}
\begin{tabular}{lcccc}
\toprule
\textbf{Model} & \textbf{Best mean RMSE} & \textbf{Std. dev.} & \textbf{1st Q} & \textbf{3rd Q} \\
\midrule
\textbf{s-BFN LS}        & \textbf{101.12} & 8.26          & 102.25          & 103.64 \\
$\boldsymbol{G_1}$       & 101.83          & 8.33          & 102.35          & 106.95 \\
$\boldsymbol{G_2}$       & 102.00          & 5.69          & 102.33          & 109.24 \\
$\boldsymbol{G_3}$       & \textbf{101.12} & \textbf{0.69} & \textbf{102.00} & \textbf{102.91} \\
\bottomrule
\end{tabular}
\end{table}

A comparative analysis of combiners clarifies the distinction between a learned probability-space aggregator (s\mbox{-}BFN) and logit averaging. Logit averaging computes the arithmetic mean of pre-softmax scores followed by a single normalisation; it is compatible with the cross-entropy geometry and is theoretically motivated in distillation~\citep{Hinton2014,DBLP:conf/ijcai/TassiGFT22}. Its effectiveness, however, is sensitive to the calibration of base predictors~\citep{255237b86b164eb3ab81ccc8564d4596}. In contrast, s\mbox{-}BFN operates on per-model class-probability vectors, maps them through a radial-basis feature representation, and learns a linear readout prior to temperature-controlled normalisation, \textcolor{black}{thereby implementing a supervised probability-space aggregation architecture motivated by the geometry associated with the task loss}. Empirically, on CIFAR-10 ensembles with \(M=10\), s\mbox{-}BFN attains higher accuracy than logit averaging. These observations are consistent with prior results on deep ensembles and uncertainty estimation~\citep{lakshminarayanan2017simple}, and support probability-space aggregation via learned, geometry-aware combiners.

In summary, the ablation indicates that robust generalization is obtained when (i) ensemble capacity is matched to task complexity and paired with architectural heterogeneity; (ii) prediction diversity is maintained at moderate--high levels within practical computational budgets; and (iii) outputs are aggregated in probability space using a learned, geometry-aware combiner such as s\mbox{-}BFN. Across datasets and tasks, the structured ensembles generally achieve higher accuracy than standard alternatives while preserving stability and computational efficiency, with the strongest gains observed in heterogeneous and higher-complexity settings.

\textcolor{black}{
Taken together, these analyses provide a theoretical interpretation of the empirically observed diversity--accuracy trade-off. The parameter $\varepsilon$ acts directly on the optimization dynamics by redistributing sample-wise updates among predictors, whereas functional disagreement and predictive uncertainty are resulting ensemble properties. Their empirical relationship with $\varepsilon$ depends on dataset complexity, ensemble size, and aggregation capacity, consistent with the capacity--complexity--diversity trade-off observed throughout the experiments. In particular, expected disagreement generally decreases with $\varepsilon$ in the image experiments, whereas predictive entropy increases with $\varepsilon$ in UCF101, illustrating that different ensemble properties can respond differently to the same diversity-control mechanism. Loss-landscape flatness is not directly measured in the present experiments; therefore, although the altered update dynamics may affect the parameter-space regions reached at convergence, the present results do not establish a relationship between $\varepsilon$, flatness, and generalization.}

\section{Conclusion and Future Work}
\label{conc}

This study introduced the Structured Basis Function Network (s\mbox{-}BFN), a loss-aware, geometry-consistent framework that unifies multi-hypothesis prediction with structured ensemble learning. The formulation constructs a structured dataset of base predictions and \textcolor{black}{learns a supervised structured combiner whose architecture is motivated by the geometry associated with the training loss}; a \textcolor{black}{diversity-control mechanism} modulates specialisation among predictors. The framework supports both closed-form least-squares and first-order training, and applies to regression, image classification, and temporally structured prediction tasks.

Across tabular regression and image classification benchmarks, \textcolor{black}{s\mbox{-}BFN delivered strong overall performance, with its relative advantage depending on the dataset and aggregation baseline.} In image classification, three trends were observed: (i) generalization improved with increasing diversity within an intermediate range of $\varepsilon$; (ii) architectural heterogeneity became more beneficial as task complexity increased; and (iii) gains from larger ensembles emerged only when sufficient diversity was induced. \textcolor{black}{In the controlled image-classification comparison, s\mbox{-}BFN significantly outperformed the fixed-averaging and several ensemble baselines, while no significant difference was observed against MLPStack-Matched and LinearStack remained competitive or superior depending on the dataset (Table~\ref{tab:strong-aggregation-baselines}).}

These findings extend to the temporal domain through the UCF101 video experiments, which support the framework under substantially higher-dimensional and temporally structured prediction regimes. In this setting, controlled diversity and increased aggregation capacity play an important role, and s\mbox{-}BFN achieves improvements over the competing aggregation strategies considered here, suggesting that the advantages of geometry-aware aggregation can become more pronounced as prediction-space complexity increases.

Together, these results contribute by: (a) operationalising a loss-centric geometric view of ensemble aggregation into a scalable and task-agnostic framework; (b) empirically characterizing how the complexity--capacity--diversity trade-off evolves across data regimes; and (c) demonstrating that probability-space aggregation provides a robust alternative to conventional ensemble methods across static and temporal tasks. The inclusion of temporally structured video data further supports the claim that aggregation capacity and structured diversity are essential when prediction spaces become richer and more structured. Future work includes extensions to higher-capacity backbones, longer-horizon temporal tasks, and low-label regimes, as well as the development of a formal characterization of how data complexity modulates the complexity--capacity--diversity trade-off.

\clearpage

\appendix  

%
\section{Comparison of Basis Function Strategies}
\label{app:bf_compare}

The predictive accuracy of structured basis function networks (s-BFN) is examined using two basis strategies: Gaussian basis functions (GBF) and radial basis functions with K-Means centers (RBF-KMeans). Both methods facilitate structured interpolation across base model outputs but differ in how they position basis centers across the input space.

These strategies are representative of the $G_1$--$G_3$ basis-function variants considered in the multivariate tabular regression ablation study (Table \ref{tab:sbfn_variants}), differing in the degree of locality and smoothing they induce in prediction space.

In this Appendix, the task involves approximating the function \( y = \sin(x_1) + \cos(x_2) + \epsilon \), where
\( \epsilon \sim \mathcal{N}(0, 0.1) \), using inputs \( x_1, x_2 \sim \mathcal{U}[-3, 3] \). Models are trained on 300 samples and evaluated on 1000 test points, with the number of basis functions varied as \( M \in \{2, 5, 10, 20, 50\} \). Figure~\ref{fig:sbfn_rmse} reports the average RMSE across five independent runs. GBF yields lower and more consistent RMSE values at small \( M \), while RBF-KMeans improves with increasing basis size but exhibits higher variance.

In this low-dimensional continuous prediction setting, performance is driven by the locality--smoothness trade-off induced by the basis functions. As illustrated in Figure~\ref{fig:sbfn_centers}, GBF centers are uniformly assigned, whereas RBF-KMeans centers adapt to data density.

From the perspective of the $G_1$--$G_3$ comparison, GBF corresponds to smoother, globally regular mappings, RBF-KMeans emphasizes highly local responses, and $G_3$ balances these behaviours by adapting both center placement and scale, yielding improved stability in this setting.

These conclusions are specific to the multivariate regression setting and do not directly transfer to the image and video experiments, which rely on different basis-function parameterizations.

\begin{figure}[h!]
    \centering
    \includegraphics[width=0.7\linewidth]{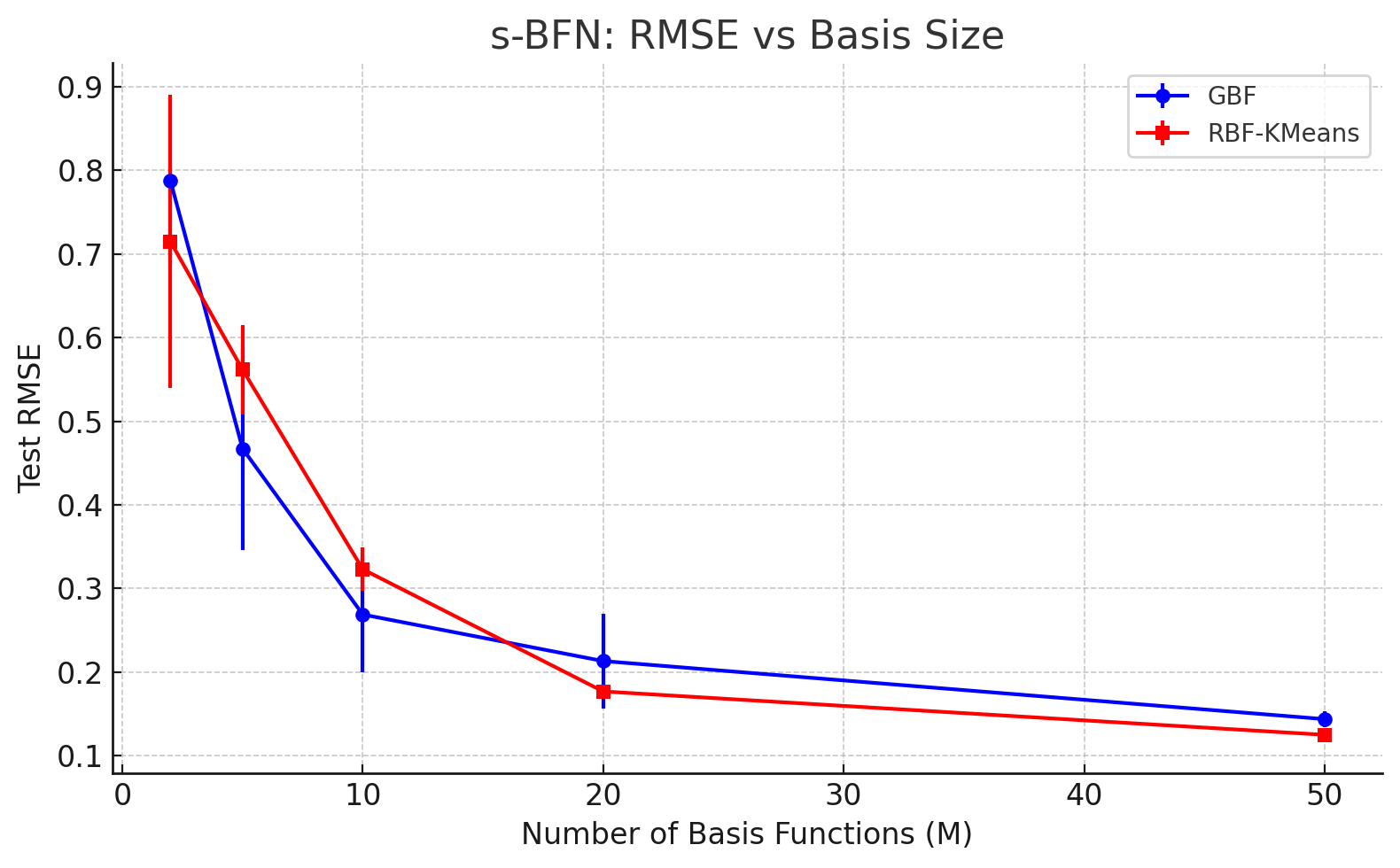}
    \caption{Average RMSE on test set for s-BFN using Gaussian basis functions (GBF, blue) and RBF-KMeans (red) across increasing basis sizes \( M \). Error bars denote standard deviation across 5 runs.}
    \label{fig:sbfn_rmse}
\end{figure}

\begin{figure}[h!]
    \centering
    \includegraphics[width=0.5\linewidth]{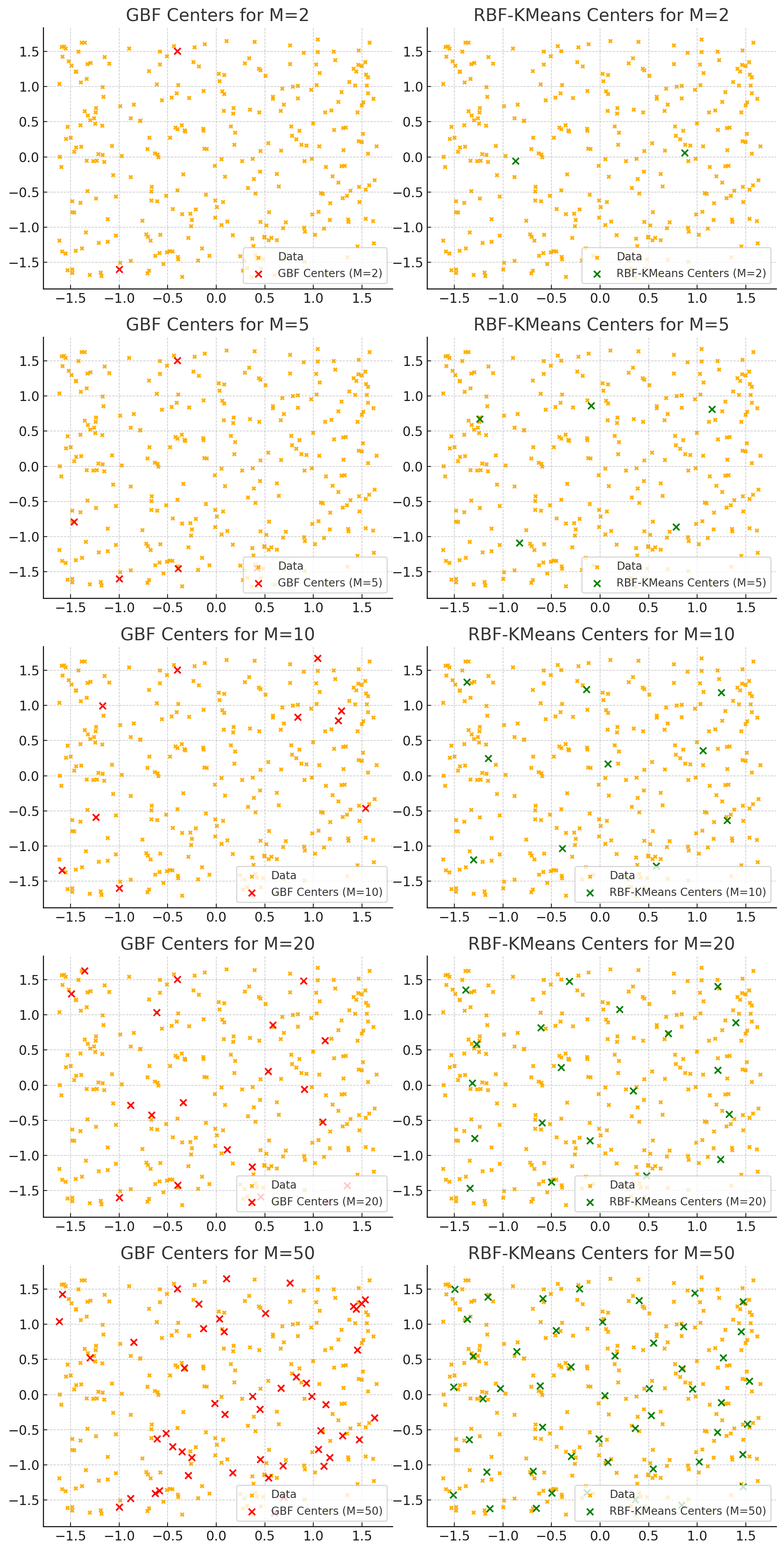}
    \caption{Basis function center locations for GBF (left, red crosses) and RBF-KMeans (right, green crosses), with \( M \in \{2, 5, 10, 20, 50\} \). RBF-KMeans adapts to data structure, GBF maintains even spread.}
    \label{fig:sbfn_centers}
\end{figure}

\clearpage

\section{Additional Analysis on s-BFN Behavior for Video Classification}
\label{app:ucf101}

This appendix provides complementary analyses for the UCF101 experiments. Specifically, we examine (i) the role of aggregation capacity, (ii) predictive entropy and diversity behavior, and (iii) the interaction between the \textcolor{black}{diversity-control parameter} and aggregation scale. These analyses clarify how s-BFN behaves in a high-dimensional temporal prediction regime and support the interpretation of the method as a stable, \textcolor{black}{loss-aware} aggregation mechanism in prediction space.

\subsection{Aggregation Capacity and the Role of $K$}

While the main subsection reports results for a fixed proportionality between ensemble size $M$ and aggregation capacity $K$, Figure~\ref{fig:ucf101-heatmap} allows a broader interpretation of how aggregation capacity interacts with diversity. As ensemble size increases, maintaining sufficient aggregation capacity is necessary to preserve performance gains. This behavior is consistent with the interpretation of s-BFN as operating in prediction space rather than expert space: increasing $K$ refines the resolution with which expert outputs are combined, rather than increasing the number of hypotheses.

Importantly, performance degrades smoothly when aggregation capacity is insufficient, rather than collapsing abruptly. This indicates that s-BFN does not rely on brittle routing decisions, but instead performs stable, geometry-aware aggregation. These trends empirically support the claim that aggregation capacity and ensemble size play complementary roles, while also showing that increasing the number of hypotheses alone is not sufficient to guarantee monotonic performance gains.

\textcolor{black}{
For substantially larger class spaces, e.g., $C\geq1000$, each expert output remains constrained to the probability simplex, such that $\|\boldsymbol{p}_{ij}\|_2\leq 1$ and consequently the concatenated structured vector satisfies $\|\boldsymbol{D}_i\|_2\leq\sqrt{M}$. Thus, increasing $C$ enlarges the ambient dimension $d_D=MC$ without causing the input norm to grow proportionally with the number of classes. This bounded input scale, however, does not eliminate the general challenges associated with distance-based representations in high-dimensional spaces. In particular, Euclidean distance can become less discriminative as dimensionality increases \citep{aggarwal2001surprising}. Since the radial responses in s\mbox{-}BFN are functions of these full-space Euclidean distances, a loss of distance contrast may in turn reduce the differentiation among RBF activations, making the choice of radial scales and representation capacity increasingly important. Accordingly, the UCF101 experiments provide empirical evidence of successful operation in the evaluated $d_D=2020$ setting, whereas behavior for substantially larger class spaces is a scaling consideration that is not empirically established in the present study.
}

\subsection{Predictive Entropy and Diversity Control}

To better understand how the \textcolor{black}{diversity-control parameter} $\varepsilon$ influences ensemble behavior, we analyze the predictive entropy of the aggregated output across configurations. Table~\ref{tab:ucf101-entropy} reports average entropy values alongside validation and test accuracy.

The entropy results should be interpreted as \textcolor{black}{a diagnostic of predictive uncertainty rather than as a direct measure of inter-predictor diversity or} a monotonic predictor of accuracy. Across the reported configurations,
increasing $\varepsilon$ increases predictive entropy, \textcolor{black}{indicating that the aggregated predictive distribution becomes less concentrated and more uncertain.}
However, the associated accuracy gains depend jointly on ensemble size, aggregation capacity, calibration, and optimization stability. Thus, a higher entropy value does not by itself guarantee a proportional improvement in accuracy.

For example, when $\varepsilon$ increases from $0.1$ to $0.6$, entropy increases for all reported ensemble sizes. The corresponding test accuracy also improves in each case, but the magnitude of the improvement is not uniform across $M$. This behavior supports the capacity--complexity--diversity interpretation adopted in the paper: controlled diversity is beneficial when the learned combiner has sufficient capacity to exploit complementary expert predictions, but the relationship between entropy and accuracy is not expected to be strictly linear or monotonic.

Notably, entropy grows smoothly with $\varepsilon$ and does not exhibit abrupt transitions. \textcolor{black}{This is consistent with gradual changes in the aggregated predictive distribution rather than abrupt predictive collapse or instability.}

\begin{table}[t]
\centering
\caption{Predictive entropy and performance of s-BFN on UCF101 across ensemble sizes and values of the \textcolor{black}{diversity-control parameter}.}
\label{tab:ucf101-entropy}
\begin{tabular}{c c c c c}
\toprule
$M$ & $\varepsilon$ & Entropy & Val Acc. & Test Acc. \\
\midrule
5  & 0.1 & 3.887 & 0.795 & 0.792 \\
5  & 0.6 & 4.014 & 0.843 & 0.843 \\
10 & 0.1 & 4.036 & 0.821 & 0.823 \\
10 & 0.6 & 4.145 & 0.841 & 0.843 \\
20 & 0.1 & 4.121 & 0.815 & 0.822 \\
20 & 0.6 & 4.147 & \textbf{0.858} & \textbf{0.864} \\
\bottomrule
\end{tabular}
\end{table}

\subsection{Joint Effect of Diversity Control and Aggregation Scale}

Figure~\ref{fig:ucf101-pareto} summarizes the interaction between the \textcolor{black}{diversity-control parameter} and aggregation scale using a two-dimensional performance map. For each aggregation capacity, intermediate-to-high diversity yields the strongest results. Importantly, the optimal diversity level shifts gradually as aggregation capacity increases, rather than exhibiting sharp or dataset-specific tuning behavior.

This joint behavior empirically supports the paper's central interpretation: s-BFN enables a smooth capacity--complexity--diversity trade-off that can be navigated using interpretable hyperparameters. The absence of sharp optima or unstable regions suggests that the proposed framework is robust and practical for larger ensemble configurations. At the same time, these results should not be read as implying that increasing either diversity or aggregation capacity independently will always improve performance. Instead, the observed gains arise from their interaction.

\begin{figure}[t]
\centering
\includegraphics[width=0.6\linewidth]{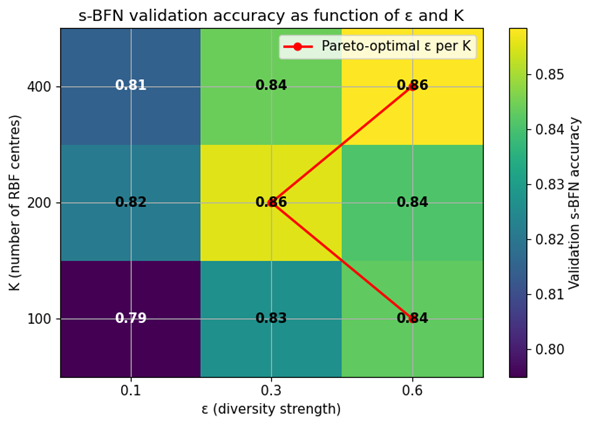}
\caption{Validation accuracy of s-BFN on UCF101 as a function of the \textcolor{black}{diversity-control parameter} and aggregation capacity. Higher values indicate better performance.}
\label{fig:ucf101-pareto}
\end{figure}

Overall, the analyses presented in this appendix reinforce the main experimental findings by demonstrating that s\mbox{-}BFN exhibits stable aggregation behavior, interpretable diversity control, and stable performance behavior across the reported UCF101 configurations. These results support the hypothesis that structured aggregation in prediction space provides a principled and scalable alternative to conventional ensemble combination strategies, while also clarifying that the empirical trends reflect an interaction among diversity, ensemble size, and aggregation capacity rather than a simple monotonic law.

\section{\textcolor{black}{Theoretical Role of the Bregman-Centroid Reference in s\mbox{-}BFN}}
\label{app:centroid}

\textcolor{black}{This appendix clarifies the theoretical role of the loss-induced Bregman centroid as a geometric reference for the s\mbox{-}BFN architecture and distinguishes it from the supervised aggregation mapping actually learned from the observed labels. The goal is not to claim that s\mbox{-}BFN explicitly solves or is trained to recover the exact centroid, but to establish that its function class is, in principle, expressive enough to represent continuous aggregation mappings, including the centroid mapping under the corresponding regularity conditions.}

\subsection{\textcolor{black}{Bregman Centroid as a Geometric Aggregation Reference}}

Let $\phi$ be a strictly convex and differentiable generator defining the Bregman divergence
\[
B_\phi(u \,\|\, v)
= \phi(u) - \phi(v) - \langle \nabla \phi(v), u - v \rangle .
\]
Given $M$ base predictors $f_1,\dots,f_M$ producing probabilistic outputs in $\Delta_C$, define the structured prediction vector
\[
\mathbf{p}(x) := \big(f_1(x),\dots,f_M(x)\big) \in \mathbb{R}^{MC}.
\]
\textcolor{black}{
The corresponding Bregman-centroid reference is given by
}
\begin{equation}
\textcolor{black}{
c_\phi\!\big(\mathbf{p}(x)\big)
:=
\arg\min_{q\in\Delta_C}
\frac{1}{M}
\sum_{m=1}^{M}
B_\phi\!\big(q\,\|\,f_m(x)\big),
\qquad
\nabla\phi\!\left(c_\phi\!\big(\mathbf{p}(x)\big)\right)
=
\frac{1}{M}
\sum_{m=1}^{M}
\nabla\phi\!\big(f_m(x)\big).
}
\label{eq:bregman_centroid}
\end{equation}

\textcolor{black}{
Under the regularity conditions associated with the selected Bregman geometry, the objective in \eqref{eq:bregman_centroid} admits a unique minimizer. The centroid therefore defines a deterministic geometric aggregation rule mapping the structured prediction vector $\mathbf{p}(x)$ to a single point in the prediction space \citep{banerjee2005clustering,wood2023unified}.
}

\textcolor{black}{
Under additional regularity assumptions, such as $\phi\in C^2$ with a positive-definite Hessian on the relevant region of $\Delta_C$, the mapping $c_\phi(\mathbf{p})$ depends continuously on $\mathbf{p}$. Consequently, on any compact subset $\mathcal{P}\subset\mathbb{R}^{MC}$, the centroid defines a continuous mapping
}
\[
\textcolor{black}{
c_\phi:\mathcal{P}\to\Delta_C.
}
\]

\subsection{\textcolor{black}{s\mbox{-}BFN as a Supervised Parametric Combiner}}

\textcolor{black}{The s\mbox{-}BFN architecture is not intended to solve \eqref{eq:bregman_centroid} analytically or to use $c_\phi(\mathbf{p})$ as its training target. Instead, it is trained supervised on the observed labels, while the centroid supplies the loss-geometric reference motivating the aggregation architecture.} Specifically, s-BFN applies a radial basis function (RBF) feature map to the structured prediction vector,
\[
\Phi_K(\mathbf{p})
=
\big(
\exp(-\|\mathbf{p}-c_1\|^2/(2\gamma_1^2)),
\dots,
\exp(-\|\mathbf{p}-c_K\|^2/(2\gamma_K^2))
\big),
\]
followed by a linear transformation and a softmax to produce a valid probability vector. Let $\mathcal{F}_K$ denote the class of functions realizable by this architecture with $K$ RBF units. A classical result in approximation theory states that RBF networks with linear readout are universal approximators of continuous functions on compact subsets of Euclidean space. Consequently, for any continuous function $f \in C(\mathcal{P};\Delta_C)$ and any $\varepsilon>0$, there exist $K$ and parameters such that
\[
\sup_{\mathbf{p} \in \mathcal{P}}
\| f_K(\mathbf{p}) - f(\mathbf{p}) \| < \varepsilon .
\]
 
Applying this result to \textcolor{black}{$f=c_\phi$} shows that the s-BFN hypothesis class $\mathcal{F}_K$ is, in principle, sufficient to \textcolor{black}{represent the loss-induced} Bregman centroid mapping arbitrarily well on the region of the structured prediction space encountered in practice. \textcolor{black}{This is a statement about the representational capacity of the hypothesis class; it does not imply that supervised training recovers the centroid mapping. The population target of the actual supervised training procedure is instead the conditional-risk-minimizing combiner}
\begin{equation}
\textcolor{black}{
G^\star(\mathbf{p})
\in
\arg\min_{q\in\Delta_C}
\mathbb{E}
\!\left[
\mathcal{L}(q,Y)
\mid
\mathbf{p}
\right].
}
\label{eq:supervised_population_combiner}
\end{equation}

\textcolor{black}{In general, $G^\star(\mathbf{p})$ and $c_\phi(\mathbf{p})$ need not coincide. Their equality requires additional conditions relating the base predictions to the conditional distribution of the target, and is not assumed as a universal property of s\mbox{-}BFN.}

\subsection{\textcolor{black}{Representation Capacity and the Role of the Number of RBF Units}}

\textcolor{black}{
For the purpose of characterizing the representational capacity of the architecture relative to the centroid reference, define
}
\[
e_K
:=
\inf_{f\in\mathcal{F}_K}
\sup_{\mathbf{p}\in\mathcal{P}}
\left\|
f(\mathbf{p})
-
\textcolor{black}{c_\phi(\mathbf{p})}
\right\|.
\]

\textcolor{black}{
When the function classes are nested, $\mathcal{F}_K\subseteq\mathcal{F}_{K+1}$, $e_K$ is non-increasing in $K$, reflecting the increase in expressive capacity of the RBF feature representation.
}

When the centroid mapping \textcolor{black}{$c_\phi$} possesses additional smoothness, classical approximation theory yields quantitative decay of $e_K$ as $K$ increases, with rates depending on the effective dimension $MC$ of the structured prediction space and on the smoothness properties of
\textcolor{black}{$c_\phi$}.

Deriving explicit approximation rates or constants in the present setting would require additional assumptions on the distribution and geometry of the base predictors' outputs, as well as on the selection or learning of RBF centers and bandwidths. For this reason, the paper treats $K$ as an aggregation-capacity parameter and studies its effect empirically.

\subsection{\textcolor{black}{Supervised Excess Risk and Relation to the Centroid Reference}}

\textcolor{black}{
Let $f_K\in\mathcal{F}_K$ denote the s\mbox{-}BFN predictor obtained by empirical risk minimization, and let $G^\star$ denote the population-optimal supervised combiner in \eqref{eq:supervised_population_combiner}. The excess population risk admits the standard decomposition
}
\[
\textcolor{black}{
\mathcal{R}(f_K)-\mathcal{R}(G^\star)
=
\underbrace{
\left(
\mathcal{R}(f_K)
-
\inf_{f\in\mathcal{F}_K}\mathcal{R}(f)
\right)
}_{\text{estimation and optimization error}}
+
\underbrace{
\left(
\inf_{f\in\mathcal{F}_K}\mathcal{R}(f)
-
\mathcal{R}(G^\star)
\right)
}_{\text{supervised approximation error}}.
}
\]

\textcolor{black}{
When the loss is Lipschitz in its prediction argument on $\Delta_C$, the second term can be controlled by the ability of $\mathcal{F}_K$ to approximate $G^\star$. The centroid representation error $e_K$ controls this term directly only when $G^\star=c_\phi$, or indirectly when the two mappings are sufficiently close under additional problem-specific conditions.
}

\textcolor{black}{
The theoretical role of s\mbox{-}BFN in this work is therefore that of a flexible supervised parametric combiner whose architecture is motivated by the loss-induced Bregman-centroid geometry. The analysis establishes the existence and continuity of the centroid reference under suitable regularity conditions and the capacity of RBF networks to represent continuous aggregation mappings. It does not claim that the supervised training procedure explicitly targets or exactly recovers the Bregman centroid.
}

\section{\textcolor{black}{Full Controlled Aggregation Benchmark}}
\label{app:full-aggregation-grid}

\textcolor{black}{
This appendix reports the complete configuration-level results underlying Table~\ref{tab:strong-aggregation-baselines}. For each $(\mathrm{dataset},M,\varepsilon,\mathrm{seed})$ configuration, all aggregation methods use the same frozen expert predictions. The \textcolor{black}{diversity-control parameter} $\varepsilon$ is applied only during upstream expert training and is not supplied to the aggregation methods. Accuracy is reported as mean test accuracy $\pm$ 95\% $t$-confidence-interval half-width over the five common seeds. For readability, the complete grid is split by dataset.
}

\textcolor{black}{
SB: SingleBest;
MP: MeanProb~\citep{lakshminarayanan2017simple};
LA: LogitAverage~\citep{DBLP:conf/ijcai/TassiGFT22};
LS: LinearStack~\citep{WOLPERT1992241};
MLP: MLPStack-Matched~\citep{WOLPERT1992241};
MoE: Mixture-of-Experts~\citep{Jacobs1991Adaptive};
NES: NE-Stack~\citep{pmlr-v293-arango25a};
NEMA: NE-MA~\citep{pmlr-v293-arango25a};
GES: Greedy Ensemble Selection~\citep{Caruana2004Ensemble}.
}


\begin{table*}[p]
\color{black}
\centering
\caption{
Complete controlled aggregation grid for MNIST. Accuracy is reported as mean $\pm$ 95\% confidence-interval half-width (\%) over five seeds.
}
\label{tab:full-grid-mnist}

\scriptsize
\setlength{\tabcolsep}{1.4pt}
\renewcommand{\arraystretch}{1.02}

\resizebox{\textwidth}{!}{%
\begin{tabular}{cc*{10}{c}}
\toprule
$M$ & $\varepsilon$
& SB & MP & LA & LS & MLP & MoE & NES & NEMA & GES & s-BFN \\
\midrule

5 & 0.0
& 28.91$\pm$9.51
& 49.58$\pm$17.60
& 51.98$\pm$5.27
& 69.78$\pm$6.60
& 66.69$\pm$7.86
& 75.48$\pm$7.28
& 46.36$\pm$11.03
& 43.03$\pm$11.82
& 49.17$\pm$15.99
& 72.27$\pm$7.99 \\

5 & 0.3
& 95.63$\pm$1.62
& 96.85$\pm$0.41
& 97.10$\pm$0.45
& 97.77$\pm$0.14
& 97.66$\pm$0.14
& 97.12$\pm$0.42
& 96.71$\pm$0.75
& 95.85$\pm$1.83
& 96.43$\pm$0.87
& 97.69$\pm$0.14 \\

5 & 0.5
& 96.98$\pm$0.77
& 97.48$\pm$0.13
& 97.51$\pm$0.17
& 98.08$\pm$0.17
& 98.04$\pm$0.17
& 97.95$\pm$0.20
& 97.35$\pm$0.67
& 97.14$\pm$0.83
& 97.75$\pm$0.41
& 98.07$\pm$0.23 \\

5 & 0.8
& 97.22$\pm$0.43
& 97.01$\pm$0.70
& 96.91$\pm$0.77
& 97.96$\pm$0.15
& 97.92$\pm$0.23
& 97.71$\pm$0.39
& 97.23$\pm$0.63
& 97.28$\pm$0.78
& 97.56$\pm$0.46
& 97.91$\pm$0.14 \\

10 & 0.0
& 9.12$\pm$1.19
& 16.07$\pm$6.10
& 19.49$\pm$6.77
& 44.75$\pm$6.89
& 41.14$\pm$5.12
& 45.85$\pm$7.07
& 22.32$\pm$9.51
& 19.86$\pm$7.35
& 22.47$\pm$4.84
& 44.32$\pm$5.65 \\

10 & 0.3
& 94.73$\pm$0.65
& 96.38$\pm$0.52
& 96.49$\pm$0.41
& 97.63$\pm$0.19
& 97.54$\pm$0.20
& 97.14$\pm$0.41
& 96.11$\pm$1.52
& 95.85$\pm$0.80
& 95.74$\pm$0.31
& 97.61$\pm$0.13 \\

10 & 0.5
& 96.29$\pm$0.66
& 96.79$\pm$0.62
& 96.78$\pm$0.65
& 97.88$\pm$0.28
& 97.84$\pm$0.18
& 97.45$\pm$0.28
& 96.76$\pm$1.08
& 96.82$\pm$0.81
& 97.12$\pm$0.30
& 97.82$\pm$0.29 \\

10 & 0.8
& 96.62$\pm$0.51
& 97.09$\pm$0.74
& 97.03$\pm$0.73
& 98.00$\pm$0.12
& 97.94$\pm$0.22
& 97.56$\pm$0.53
& 97.32$\pm$0.66
& 97.33$\pm$0.71
& 97.46$\pm$0.70
& 98.00$\pm$0.19 \\

20 & 0.0
& 11.46$\pm$4.27
& 11.59$\pm$2.24
& 16.38$\pm$4.07
& 55.18$\pm$3.77
& 47.48$\pm$4.41
& 49.86$\pm$5.42
& 23.80$\pm$7.24
& 19.80$\pm$3.26
& 14.48$\pm$4.25
& 49.94$\pm$3.06 \\

20 & 0.3
& 91.13$\pm$0.88
& 94.73$\pm$0.61
& 94.82$\pm$0.44
& 97.23$\pm$0.24
& 97.06$\pm$0.26
& 95.51$\pm$0.48
& 95.84$\pm$0.43
& 94.30$\pm$1.28
& 93.78$\pm$1.41
& 97.04$\pm$0.30 \\

20 & 0.5
& 94.61$\pm$0.73
& 96.07$\pm$0.56
& 96.06$\pm$0.51
& 97.57$\pm$0.14
& 97.47$\pm$0.21
& 96.71$\pm$0.20
& 96.72$\pm$0.32
& 96.28$\pm$0.79
& 96.31$\pm$0.62
& 97.28$\pm$0.22 \\

20 & 0.8
& 95.92$\pm$0.35
& 96.55$\pm$0.57
& 96.38$\pm$0.72
& 97.71$\pm$0.19
& 97.66$\pm$0.14
& 96.99$\pm$0.28
& 97.17$\pm$0.18
& 96.71$\pm$0.51
& 96.95$\pm$0.27
& 97.39$\pm$0.23 \\

50 & 0.0
& 11.04$\pm$3.81
& 12.51$\pm$3.04
& 13.53$\pm$3.52
& 76.13$\pm$2.32
& 74.23$\pm$2.12
& 69.80$\pm$3.32
& 25.01$\pm$7.29
& 24.44$\pm$1.92
& 17.53$\pm$4.38
& 61.40$\pm$6.50 \\

50 & 0.3
& 75.88$\pm$6.51
& 85.08$\pm$1.54
& 86.67$\pm$1.55
& 96.74$\pm$0.17
& 96.68$\pm$0.17
& 89.59$\pm$4.46
& 90.68$\pm$6.22
& 81.89$\pm$2.66
& 85.29$\pm$1.80
& 96.30$\pm$0.34 \\

50 & 0.5
& 88.15$\pm$1.40
& 92.00$\pm$0.90
& 91.89$\pm$1.09
& 97.28$\pm$0.23
& 97.30$\pm$0.25
& 92.87$\pm$3.41
& 94.36$\pm$2.64
& 91.35$\pm$0.96
& 92.35$\pm$1.58
& 96.47$\pm$0.27 \\

50 & 0.8
& 92.71$\pm$0.73
& 94.38$\pm$0.76
& 93.94$\pm$0.86
& 97.49$\pm$0.11
& 97.45$\pm$0.11
& 95.84$\pm$0.69
& 95.86$\pm$0.96
& 94.99$\pm$0.43
& 95.21$\pm$0.58
& 96.41$\pm$0.24 \\

\bottomrule
\end{tabular}%
}
\end{table*}


\begin{table*}[p]
\color{black}
\centering
\caption{
Complete controlled aggregation grid for Fashion-MNIST. Accuracy is reported as mean $\pm$ 95\% confidence-interval half-width (\%) over five seeds.
}
\label{tab:full-grid-fashion}

\scriptsize
\setlength{\tabcolsep}{1.4pt}
\renewcommand{\arraystretch}{1.02}

\resizebox{\textwidth}{!}{%
\begin{tabular}{cc*{10}{c}}
\toprule
$M$ & $\varepsilon$
& SB & MP & LA & LS & MLP & MoE & NES & NEMA & GES & s-BFN \\
\midrule

5 & 0.0
& 26.37$\pm$8.38
& 32.29$\pm$6.95
& 33.86$\pm$3.42
& 57.68$\pm$1.81
& 55.02$\pm$3.93
& 58.76$\pm$2.42
& 36.86$\pm$5.56
& 36.26$\pm$7.28
& 34.24$\pm$6.95
& 58.58$\pm$2.62 \\

5 & 0.3
& 66.84$\pm$3.46
& 73.67$\pm$3.16
& 74.20$\pm$2.70
& 79.06$\pm$0.54
& 77.37$\pm$0.21
& 77.94$\pm$1.48
& 74.29$\pm$1.09
& 71.77$\pm$5.32
& 71.30$\pm$2.23
& 79.05$\pm$0.17 \\

5 & 0.5
& 74.85$\pm$2.22
& 76.53$\pm$0.91
& 76.51$\pm$0.75
& 79.86$\pm$0.56
& 79.01$\pm$0.73
& 78.81$\pm$0.40
& 76.41$\pm$2.00
& 75.89$\pm$2.75
& 77.18$\pm$1.17
& 79.81$\pm$0.34 \\

5 & 0.8
& 76.62$\pm$1.48
& 76.78$\pm$1.13
& 76.72$\pm$1.10
& 79.88$\pm$0.68
& 78.97$\pm$0.78
& 78.34$\pm$1.17
& 77.32$\pm$1.55
& 76.65$\pm$2.44
& 77.40$\pm$1.21
& 80.18$\pm$0.88 \\

10 & 0.0
& 8.40$\pm$3.07
& 19.54$\pm$7.41
& 27.99$\pm$4.40
& 55.21$\pm$3.20
& 50.25$\pm$2.56
& 55.33$\pm$2.50
& 27.14$\pm$3.66
& 21.48$\pm$6.68
& 19.37$\pm$3.64
& 54.45$\pm$2.96 \\

10 & 0.3
& 69.16$\pm$2.02
& 73.60$\pm$1.37
& 74.08$\pm$1.05
& 79.19$\pm$0.75
& 77.64$\pm$0.66
& 78.05$\pm$1.39
& 72.64$\pm$3.24
& 71.42$\pm$5.29
& 73.44$\pm$1.00
& 79.08$\pm$0.83 \\

10 & 0.5
& 71.66$\pm$2.11
& 75.12$\pm$1.48
& 75.06$\pm$1.56
& 80.61$\pm$0.46
& 79.48$\pm$0.61
& 79.50$\pm$0.93
& 74.71$\pm$4.11
& 74.41$\pm$2.53
& 76.31$\pm$1.43
& 80.58$\pm$0.76 \\

10 & 0.8
& 76.41$\pm$0.57
& 76.86$\pm$0.67
& 76.83$\pm$0.69
& 81.11$\pm$0.30
& 80.15$\pm$0.49
& 78.96$\pm$0.51
& 76.58$\pm$4.20
& 77.56$\pm$0.63
& 78.22$\pm$0.51
& 80.96$\pm$0.20 \\

20 & 0.0
& 8.68$\pm$2.79
& 14.86$\pm$2.74
& 20.30$\pm$3.84
& 67.68$\pm$0.73
& 64.00$\pm$0.50
& 65.62$\pm$0.78
& 31.98$\pm$5.74
& 22.08$\pm$9.29
& 19.04$\pm$3.71
& 65.86$\pm$1.41 \\

20 & 0.3
& 67.21$\pm$1.65
& 71.16$\pm$1.17
& 71.73$\pm$1.11
& 79.57$\pm$1.00
& 78.44$\pm$0.87
& 75.77$\pm$0.77
& 73.43$\pm$1.09
& 71.12$\pm$2.62
& 71.73$\pm$1.05
& 79.11$\pm$0.96 \\

20 & 0.5
& 71.54$\pm$0.65
& 73.35$\pm$0.73
& 73.24$\pm$1.15
& 80.38$\pm$0.77
& 79.50$\pm$0.87
& 77.47$\pm$0.70
& 75.30$\pm$0.50
& 74.28$\pm$0.39
& 74.36$\pm$0.48
& 80.04$\pm$0.79 \\

20 & 0.8
& 73.52$\pm$1.25
& 74.94$\pm$0.67
& 74.76$\pm$0.78
& 81.16$\pm$0.38
& 80.68$\pm$0.58
& 78.33$\pm$0.82
& 77.04$\pm$0.69
& 76.12$\pm$0.55
& 76.12$\pm$0.58
& 80.52$\pm$0.56 \\

50 & 0.0
& 11.66$\pm$2.11
& 12.87$\pm$2.76
& 15.99$\pm$3.05
& 75.50$\pm$0.84
& 74.63$\pm$0.74
& 72.77$\pm$1.54
& 34.08$\pm$5.61
& 27.10$\pm$4.03
& 19.33$\pm$1.60
& 70.93$\pm$1.41 \\

50 & 0.3
& 60.93$\pm$1.37
& 64.44$\pm$2.46
& 64.27$\pm$2.83
& 80.96$\pm$0.48
& 80.72$\pm$0.44
& 71.24$\pm$2.02
& 70.76$\pm$2.04
& 66.56$\pm$0.43
& 65.63$\pm$2.07
& 79.68$\pm$0.41 \\

50 & 0.5
& 67.06$\pm$1.45
& 69.26$\pm$1.63
& 68.88$\pm$1.86
& 81.48$\pm$0.35
& 81.34$\pm$0.62
& 76.48$\pm$2.39
& 72.80$\pm$1.99
& 70.86$\pm$1.33
& 70.85$\pm$1.03
& 80.07$\pm$0.38 \\

50 & 0.8
& 70.09$\pm$1.34
& 71.74$\pm$0.95
& 71.35$\pm$1.12
& 81.79$\pm$0.32
& 81.81$\pm$0.33
& 77.10$\pm$2.11
& 74.45$\pm$2.32
& 73.07$\pm$1.03
& 72.71$\pm$0.87
& 80.30$\pm$0.51 \\

\bottomrule
\end{tabular}%
}
\end{table*}


\begin{table*}[p]
\color{black}
\centering
\caption{
Complete controlled aggregation grid for CIFAR-10. Accuracy is reported as mean $\pm$ 95\% confidence-interval half-width (\%) over five seeds.
}
\label{tab:full-grid-cifar}

\scriptsize
\setlength{\tabcolsep}{1.4pt}
\renewcommand{\arraystretch}{1.02}

\resizebox{\textwidth}{!}{%
\begin{tabular}{cc*{10}{c}}
\toprule
$M$ & $\varepsilon$
& SB & MP & LA & LS & MLP & MoE & NES & NEMA & GES & s-BFN \\
\midrule

5 & 0.0
& 19.81$\pm$5.62
& 33.78$\pm$5.21
& 28.36$\pm$2.07
& 39.64$\pm$2.23
& 38.71$\pm$2.49
& 40.49$\pm$2.55
& 29.96$\pm$3.79
& 26.68$\pm$5.65
& 29.76$\pm$6.24
& 40.39$\pm$2.33 \\

5 & 0.5
& 65.36$\pm$2.89
& 68.68$\pm$1.55
& 69.51$\pm$1.66
& 71.91$\pm$1.10
& 71.75$\pm$1.05
& 71.00$\pm$0.94
& 68.20$\pm$3.43
& 67.65$\pm$2.67
& 69.67$\pm$1.46
& 71.80$\pm$0.91 \\

5 & 0.8
& 65.82$\pm$4.00
& 68.07$\pm$1.43
& 68.25$\pm$1.31
& 71.36$\pm$1.27
& 71.32$\pm$1.16
& 70.49$\pm$1.22
& 68.43$\pm$2.76
& 68.19$\pm$1.82
& 69.78$\pm$1.43
& 71.58$\pm$1.23 \\

10 & 0.0
& 10.00$\pm$0.00
& 14.34$\pm$3.05
& 15.17$\pm$0.65
& 22.01$\pm$1.56
& 16.75$\pm$5.41
& 16.48$\pm$5.44
& 12.39$\pm$3.87
& 10.21$\pm$0.58
& 14.34$\pm$3.05
& 16.73$\pm$6.18 \\

10 & 0.5
& 58.27$\pm$1.15
& 65.12$\pm$0.61
& 65.65$\pm$0.77
& 68.90$\pm$0.46
& 68.96$\pm$0.47
& 67.63$\pm$0.43
& 64.78$\pm$4.23
& 65.55$\pm$1.63
& 66.85$\pm$0.76
& 68.89$\pm$0.49 \\

10 & 0.8
& 65.46$\pm$1.44
& 67.69$\pm$0.95
& 67.88$\pm$1.02
& 71.65$\pm$0.40
& 71.91$\pm$0.38
& 70.26$\pm$0.51
& 68.60$\pm$3.69
& 69.21$\pm$1.25
& 69.86$\pm$0.36
& 71.91$\pm$0.40 \\

20 & 0.0
& 10.28$\pm$2.47
& 10.47$\pm$4.05
& 15.87$\pm$6.16
& 33.84$\pm$1.31
& 31.65$\pm$1.42
& 32.28$\pm$1.54
& 20.11$\pm$1.50
& 17.54$\pm$3.21
& 14.85$\pm$1.69
& 31.79$\pm$1.48 \\

20 & 0.5
& 53.53$\pm$2.22
& 58.45$\pm$0.50
& 59.15$\pm$0.48
& 64.58$\pm$0.96
& 64.91$\pm$0.97
& 63.53$\pm$1.22
& 61.57$\pm$1.02
& 60.11$\pm$1.01
& 60.92$\pm$1.07
& 64.51$\pm$1.16 \\

20 & 0.8
& 59.52$\pm$2.06
& 63.72$\pm$0.90
& 63.88$\pm$0.95
& 68.85$\pm$0.48
& 69.01$\pm$0.66
& 67.05$\pm$0.62
& 67.15$\pm$0.83
& 66.61$\pm$1.05
& 66.77$\pm$0.98
& 68.76$\pm$0.61 \\

50 & 0.0
& 9.46$\pm$1.30
& 12.83$\pm$3.66
& 15.13$\pm$2.81
& 40.96$\pm$0.59
& 39.88$\pm$0.29
& 38.38$\pm$0.59
& 22.37$\pm$2.42
& 19.59$\pm$2.65
& 16.89$\pm$1.30
& 36.04$\pm$1.34 \\

50 & 0.5
& 49.84$\pm$1.00
& 50.82$\pm$0.28
& 51.04$\pm$0.61
& 59.31$\pm$0.29
& 59.45$\pm$0.59
& 57.26$\pm$1.59
& 55.21$\pm$0.57
& 53.41$\pm$0.30
& 53.47$\pm$0.18
& 56.82$\pm$0.72 \\

50 & 0.8
& 52.56$\pm$0.75
& 55.54$\pm$0.66
& 55.55$\pm$0.81
& 63.15$\pm$1.00
& 63.39$\pm$1.06
& 61.20$\pm$1.70
& 60.19$\pm$0.59
& 58.85$\pm$0.81
& 59.15$\pm$0.94
& 61.33$\pm$1.45 \\

\bottomrule
\end{tabular}%
}
\end{table*}

\clearpage










\bibliographystyle{apalike}
\bibliography{cas-refs}
\end{document}